\documentclass[12pt]{article}

\usepackage{amsmath}
\usepackage{amssymb}
\usepackage{mathtools}
\usepackage{amsthm}

\usepackage{microtype}
\usepackage{lmodern}
\usepackage{graphicx}
\usepackage{adjustbox}
\usepackage{booktabs}
\usepackage{arydshln}
\usepackage{multirow}
\usepackage{subcaption}

\usepackage{tikz}
\usetikzlibrary{positioning}
\usetikzlibrary{arrows.meta}

\tikzset{
    block/.style={
        draw, 
        rectangle, 
        minimum width=3cm, 
        minimum height=1cm, 
        align=center
    }
}

\usepackage{hyperref}
\hypersetup{
    colorlinks=true,
    linkcolor=blue,
    citecolor=blue,
    urlcolor=blue
}

\usepackage{paralist}

\usepackage[textsize=tiny]{todonotes}

\usepackage[capitalize,noabbrev]{cleveref}

\usepackage{algorithm}
\usepackage{algpseudocode}

\theoremstyle{plain}
\newtheorem{theorem}{Theorem}[section]

\theoremstyle{definition}

\theoremstyle{remark}

\usepackage[numbers]{natbib}
\usepackage[margin=1in]{geometry}
\usepackage[textsize=tiny]{todonotes}

\begin{document}

\title{Boosting Statistic Learning with Synthetic Data from Pretrained Large Models}
\author{
    Jialong Jiang\textsuperscript{1}, 
    Wenkang Hu\textsuperscript{1}, 
    Jian Huang\textsuperscript{2}, 
    Yuling Jiao\textsuperscript{3}, 
    Xu Liu\textsuperscript{1}\thanks{Corresponding author: \texttt{liu.xu@mail.shufe.edu.cn}} \\
    \textsuperscript{1}Shanghai University of Finance and Economics, Shanghai, China \\
    \textsuperscript{2}The Hong Kong Polytechnic University, Hong Kong SAR, China \\
    \textsuperscript{3}School of Mathematics and Statistics, Wuhan University, Wuhan, China
}
\date{}
\maketitle


\begin{abstract}

The rapid advancement of generative models, such as Stable Diffusion, raises a key question: how can synthetic data from these models enhance predictive modeling? While they can generate vast amounts of datasets, only a subset meaningfully improves performance. We propose a novel end-to-end framework that generates and systematically filters synthetic data through domain-specific statistical methods, selectively integrating high-quality samples for effective augmentation. Our experiments demonstrate consistent improvements in predictive performance across various settings, highlighting the potential of our framework while underscoring the inherent limitations of generative models for data augmentation. Despite the ability to produce large volumes of synthetic data, the proportion that effectively improves model performance is limited. 
\end{abstract}

\section{Introduction}
\label{intr}

Data lies at the core of modern artificial intelligence (AI) and machine learning (ML) systems, serving as the foundation for their performance, robustness, and generalization capabilities. Despite its critical role, the availability of high-quality, representative datasets remains a pervasive challenge, particularly in domains such as healthcare and finance. These fields often face constraints related to privacy, regulatory compliance, and high data acquisition costs, leading to a scarcity of training data that hampers the development of reliable ML models\cite{albahri2023systematic,assefa2020generating}. This limitation is especially detrimental in high-stakes applications where model predictions directly influence critical decision-making processes.

Traditional approaches to data generation have predominantly relied on statistical methodologies such as bootstrapping, Monte Carlo simulations, and parametric sampling techniques \cite{efron1994introduction}. These methods operate under the assumption that the underlying data distribution can be either explicitly known or accurately approximated. However, such approaches encounter significant limitations when applied to complex, high-dimensional data distributions that are typical in many real-world applications \cite{donoho2000high}. Moreover, real-world data distributions often exhibit intricate dependencies and non-linearities that elude conventional statistical modeling techniques.

The emergence of generative modeling frameworks has attempted to address these limitations by learning data distributions directly from observations. Pioneering architectures such as Generative Adversarial Networks (GANs) \cite{goodfellow2014generative} and Variational Autoencoders (VAEs) \cite{kingma2014auto} have demonstrated promising results in modeling complex distributions. Nevertheless, these methods require substantial computational resources, careful hyperparameter tuning, and often fail to fully capture the nuanced properties of the target data distribution \cite{Bao_2017_ICCV, salimans2016improved}, posing challenges for comprehensive evaluation \cite{AssessingGenerativeModelsPRRecall2018}.

In recent years, Large Models (LMs) have revolutionized generative modeling by leveraging extensive pre-training on massive datasets to synthesize and encode knowledge across diverse domains \cite{HAN2021225}. Among these, \textit{Stable Diffusion} \cite{rombach2022high} has emerged as a particularly effective model for generating semantically rich and diverse outputs. Unlike traditional methods, \textit{Stable Diffusion} employs latent space representations to produce high-quality synthetic data that reflects the inherent complexity of the underlying distribution. This capability makes it a promising candidate for addressing data scarcity in ML and statistical modeling.

A key challenge, however, lies in the selection of high-quality synthetic data from the vast quantities generated by these models \cite{bolon2013review}. While \textit{Stable Diffusion} can theoretically produce an unlimited amount of data, not all generated samples contribute meaningfully to model performance. To address this, we propose domain-specific metrics for data evaluation and selection. For statistical datasets, we utilize \(p\)-value-based hypothesis testing to measure the relevance of generated samples, while for image data, the Wasserstein distance \cite{villani2009wasserstein} is employed to assess fidelity to the original distribution. These metrics enable the systematic filtering of low-quality information, ensuring that only high-quality data is incorporated into downstream applications.

This paper introduces a novel framework that leverages \textit{Stable Diffusion} for data augmentation through tabular data to image data. Building upon recent work in using generative models for data synthesis across various domains \cite{Trabucco2024DAFusion, ShenBoostingDataAnalytics, CinquiniBoostingSyntheticCogMI2021, Hemmat2023FeedbackGuided, Khurana2023Fillup}, our approach presents a unique perspective of generation and filtering, especially in numerical data. Unlike traditional or recent tabular data augmentation methods such as SMOTE \cite{Chawla2002SMOTE},TVAE and CTGAN \cite{Xu2019CTGAN}, our method transforms numerical datasets into grayscale images, generates synthetic data via diffusion processes (leveraging capabilities seen in controllable image generation \cite{Gal2022TextualInversion}), and maps the augmented data back into the original numerical space. By rigorously filtering the generated data using statistical methods, we effectively enhance predictive modeling and statistical inference. Importantly, our findings reveal that while large generative models can produce vast quantities of data, the fraction of data that meaningfully improves estimation and prediction is inherently limited. This underscores both the potential and the constraints of leveraging large models for data augmentation. Our findings highlight the practical utility of large generative models in resource-constrained scenarios and provide a pathway for further refinements in leveraging generative frameworks for data augmentation.

{\bf Roadmap.}
The rest of this section is organized as follows. In Section ~\ref{method}, we present a novel data augmentation framework based on Stable Diffusion XL refiner that enhances synthetic data generation.In Section~\ref{Boost}, we present a robust data filtering mechanism to curate the synthetic samples and empirically validate its effectiveness through comprehensive simulation experiments. Experiments on real world data are presented in Section~\ref{experiments}.

\section{Synthetic Data via Stable Diffusion}\label{method}

\begin{figure}[htbp]
    \centering
    \includegraphics[width=0.45\textwidth]{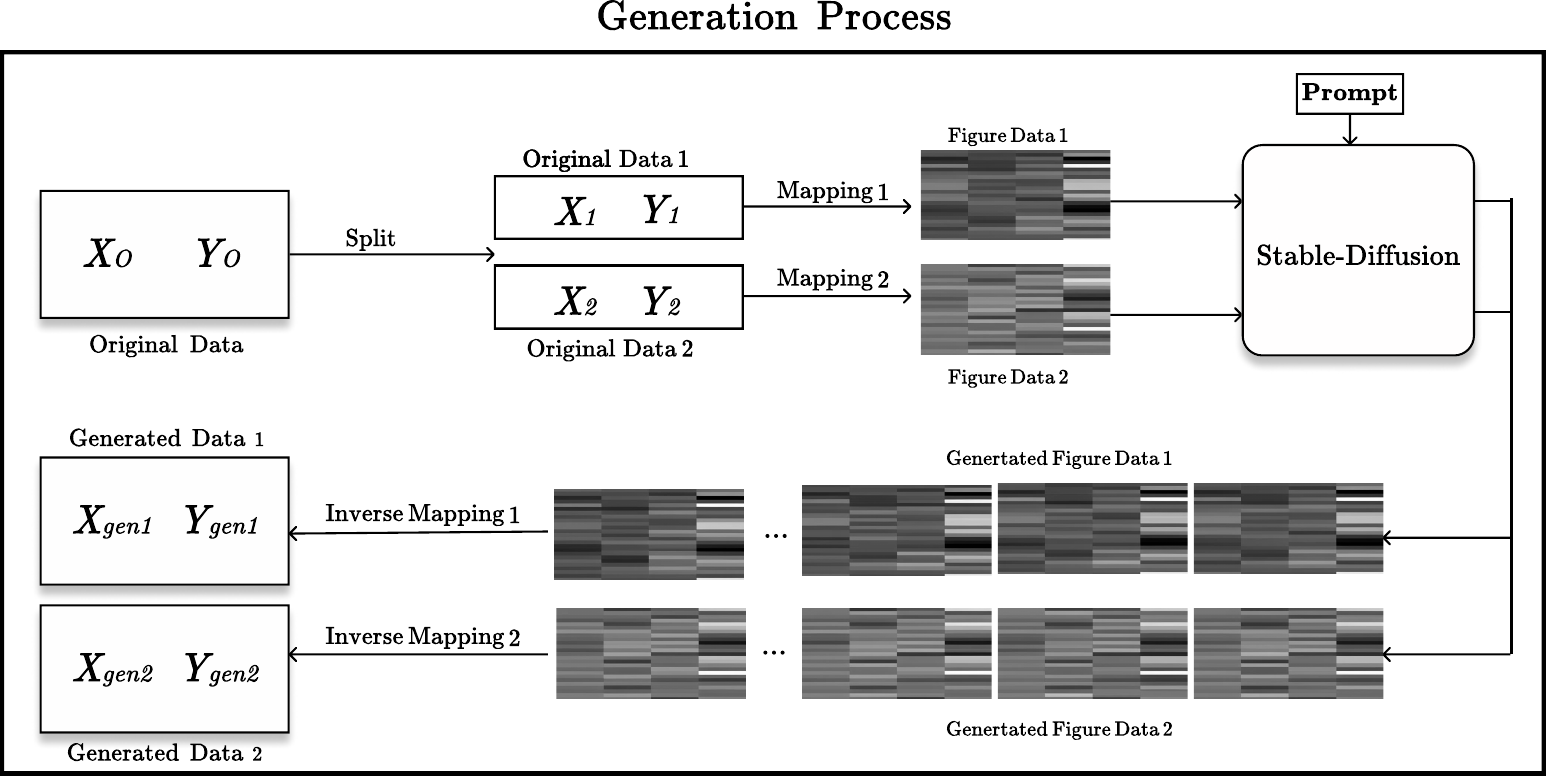}
    \caption{Tabular Data Generation Framework}
    \label{generation}
\end{figure}

To effectively expand the original dataset, we propose a novel data augmentation framework (Figure \ref{generation}) utilizing the Stable Diffusion XL refiner (SD-XL) model \cite{podell2023sdxl}, which addresses critical geometric distortions in synthetic image generation. Specifically, the refiner stage enhances structural integrity by recursively rectifying edge alignment and suppressing irregular pixel clusters. This ensures geometric consistency in line segments and regional boundaries, mitigating variable misidentification risks caused by skewed features. 

The original dataset $[X_o,Y_o]$ is partitioned into subsets. For clarity and as demonstrated in Equation (1), we primarily consider a strict horizontal bisection (the dashline) into two mutually exclusive subsets, $V_1$ and $V_2$:
\begin{align}
\begin{bmatrix}
    X_o,& Y_o
\end{bmatrix} =
\begin{bmatrix}
    V_1 \\
    \hdashline
    V_2
\end{bmatrix} =
\begin{bmatrix}
    X_{1},& Y_{1} \\
    \hdashline
    X_{2},& Y_{2}
\end{bmatrix},
\end{align}
where $[X_1, Y_1] = V_1 \in \mathbb{R}^{m \times (d+1)}$ and $[X_2, Y_2] = V_2 \in \mathbb{R}^{(n-m) \times (d+1)}$. Here $n$ is the total sample size, $d$ is the number of predictor dimensions, and $m = \lfloor n/2 \rfloor$. For our framework, it is crucial that these subsets are statistically independent and drawn identically and independently from the same underlying distribution $P_{XY}$. The subsets $V_1$ and $V_2$ play interchangeable roles throughout the data augmentation and filtering process: one is used to generate synthetic data, and the other acts as a hold-out set to evaluate and select the generated samples. This strategy ensures that evaluation is performed on data statistically independent from the subset used for generation, promoting robustness in the augmented dataset. Notably, for generation of image data, the diffusion process can be directly applied without additional preprocessing steps.

We apply a reversible mapping \(\mathcal{M}_i: V_i \rightarrow \mathcal{F}_i\) (\(i=1,2\)) to the data subsets \(V_1\) and \(V_2\), transforming them into a new representation space \(\mathcal{F}_i\). The mapping \(\mathcal{M}_i\) is specifically designed to ensure non-negativity and invertibility, which enhances the numerical sensitivity and interpretability of the data. After transformation, the data is normalized to create grayscale representations \(\mathcal{F}_i\). These representations are subsequently used as inputs for downstream tasks, forming the basis for data augmentation. This iterative process selects synthetic data based on the independent contributions of \(V_1\) and \(V_2\), fostering a robust and reliable augmentation framework. The alternating and reversible nature of \(\mathcal{M}_i\) preserves the independence between \(V_1\) and \(V_2\), which is critical for maintaining model generalization and performance.

The \texttt{SD-XL} model is employed to process the transformed data \(\mathcal{F}_i\) using a carefully designed prompt within an image-to-image diffusion framework. By varying the diffusion strength parameter \(k \in [0.001, 1]\), a diverse set of synthetic images is generated:
\begin{align}
\mathcal{G}_{i}^{(k)} = \text{\texttt{SD-XL}}(\mathcal{F}_i, \text{prompt}, \text{strength}=k).
\end{align}
To reconstruct numerical data from the generated images \(\mathcal{G}_i^{(k)}\), the inverse mapping \(\mathcal{M}^{-1}\) is applied to each pixel value \(p_i^{(k)}\) in \(\mathcal{G}_{i}^{(k)}\):
\begin{align}
[X_{\text{gen}}, Y_{\text{gen}}]^{(k)}_i = \mathcal{M}^{-1}(p_i^{(k)}),
\end{align}
where \(\mathcal{M}^{-1}\) denotes the inverse transformation that maps the synthetic results back to the original data space. This process establishes a bijective correspondence between the synthetic images \(\mathcal{G}_{i}^{(k)}\) and the original data \([X_o, Y_o]\), ensuring dimensional consistency and preserving the structural integrity of the augmented data. 

During the generation process, our framework operates without explicit knowledge of the underlying data distribution. The model requires only two fundamental specifications: (i) the cardinality of independent and dependent variables, corresponding to the column dimensionality in the graphical representation, and (ii) the positional indices of these variables within the representation space. This minimalistic specification ensures both computational efficiency and flexibility in handling diverse data structures.

A key advantage of this framework is the dimensional stability of the Stable Diffusion process, which ensures that each transformed feature vector maintains its correspondence with the target variable. This stability prevents ambiguities in the mapping process, thereby preserving the predictive power of the augmented dataset. By iteratively generating and reconstructing synthetic data, our framework leverages the inherent properties of reversible mappings and diffusion models to enhance both the quality and diversity of the data, providing a scalable solution to augment small datasets.

\section{Boostability Identification}\label{Boost}
{Roadmap.} 
This section is organized as follows. 
We first introduce a dual-source transfer learning framework for boostability quantification. 
Next, we propose a Wasserstein distance-based method for boostability verification. 
Finally, we validate our approach through simulation studies in low- and high-dimensional settings.
\subsection{Boostability Quantification via Transfer Learning}

In this work, we introduce a dual-source transfer learning framework (Algorithm \ref{alg:dual_transfer}), which aims to enhance the performance of models on target domains by effectively leveraging two source domains. The central idea is to adapt source models using statistical techniques, thereby reducing the domain shift between the source and target domains.

Algorithm \ref{alg:dual_transfer} takes as inputs the two source domains ($\mathcal{S}_1, \mathcal{S}_2$), two target domains ($\mathcal{T}_1, \mathcal{T}_2$), an independent test set ($\mathcal{D}_{\text{test}}$), a set of sampling ratios ($\mathcal{P}$), and a batch size ($b$). The process begins by establishing a baseline error $\varepsilon_0$ using a Lasso regression model. For each sampling ratio $\rho \in \mathcal{P}$, the algorithm samples subsets from source domains, adapts models ($f_{1|2}, f_{2|1}$) using batches of size $b$, and evaluates performance and adaptability metrics to find the optimal sampling ratio $\rho^*$ that minimizes the average prediction error. A complete and formal definition of all notations, variables, and functions used in this algorithm can be found in Appendix \ref{sec:alg_dual_transfer_notations}. 

\begin{algorithm}
\caption{Dual-Source Transfer Learning with Statistical Adaptation for Tabular Data}
\label{alg:dual_transfer}
\begin{algorithmic}[1]
   \State \textbf{Input:} Source domains $\mathcal{S}_1, \mathcal{S}_2$, target domains $\mathcal{T}_1, \mathcal{T}_2$, test set $\mathcal{D}_{\text{test}}$, ratio set $\mathcal{P} = \{\rho_i\}_{i=1}^n$, batch size $b$
   \State Initialize $\varepsilon_0 \gets \text{Lasso}(\mathcal{T}_1 \cup \mathcal{T}_2, \mathcal{D}_{\text{test}})$
   \For{$\rho \in \mathcal{P}$}
       \State $\mathcal{E}_{\text{comb}} \gets \emptyset$, $\mathcal{E}_{\text{adapt}} \gets \emptyset$
       \For{$k = 1$ \textbf{to} $K$} \Comment{Number of iterations $K$ defined in text}
           \State Sample $\tilde{\mathcal{S}}_1$ from $\mathcal{S}_1$ with ratio $\rho$, $\tilde{\mathcal{S}}_2$ from $\mathcal{S}_2$ with ratio $\rho$
           \State $\mathcal{B}_1, \mathcal{B}_2 \gets \text{BatchSplit}(\tilde{\mathcal{S}}_1, \tilde{\mathcal{S}}_2, b)$
           \State $f_{1|2} \gets \text{Adapt}(\mathcal{T}_2, \mathcal{B}_1)$, $f_{2|1} \gets \text{Adapt}(\mathcal{T}_1, \mathcal{B}_2)$
           \If{$f_{1|2}, f_{2|1}$ pass validation criteria} \Comment{Validation criteria described in text}
               \State Calculate combined prediction $\hat{y} = \frac{1}{2}(f_{1|2}(x) + f_{2|1}(x))$ for $x \in \mathcal{D}_{\text{test}}$
               \State $\mathcal{E}_{\text{comb}} \gets \mathcal{E}_{\text{comb}} \cup \{\text{MSE}(\hat{y}, y_{\text{test}})\}$
               \State $\mathcal{E}_{\text{adapt}} \gets \mathcal{E}_{\text{adapt}} \cup \{d(f_{1|2}, f_{2|1})\}$
           \EndIf
       \EndFor
       \State Record $\bar{\varepsilon}_\rho \gets \text{mean}(\mathcal{E}_{\text{comb}})$, $\bar{\Delta}_\rho \gets \text{mean}(\mathcal{E}_{\text{adapt}})$
   \EndFor
   \State \textbf{Return:} $\rho^* = \arg\min_{\rho \in \mathcal{P}} \bar{\varepsilon}_\rho$
\end{algorithmic}
\end{algorithm}

In lower-dimensional settings, we utilize the \texttt{glmtrans} package to identify transferable sources and estimate the parameter vector ${\beta}$ in generalized linear models. The methodology proposed by \cite{tian2023transfer} in the \texttt{glmtrans} package offers a computationally efficient implementation of a two-step multi-source transfer learning framework specifically designed for generalized linear models (GLMs) . A distinctive feature of this methodology is its transferable source detection algorithm, which mitigates the risk of negative transfer by selectively incorporating only those sources that are beneficial for parameter estimation. This enhances both model accuracy and interpretability.

In high-dimensional scenarios, we use statistical hypothesis test to identify transferable sources, which is implemented by the R package \href{ https://github.com/xliusufe/hdtrd}{\texttt{hdtrd}}. These selected sources are subsequently used as inputs for the \texttt{glmtrans} package to estimate ${\beta}$. The \texttt{hdtrd} package is specifically designed for high-dimensional contexts and offers robust tools for source selection. This ensures that only the most informative sources are used in the transfer learning process, thus enhancing the overall model performance by filtering out irrelevant or detrimental sources.


\subsection{Boostability Verification through Distributional Fidelity}

For the generation of synthetic images, we employ a rigorous screening methodology based on the \textit{Wasserstein distance} to ensure the quality and relevance of the generated samples. The Wasserstein distance is a metric quantifying the divergence between probability distributions. Formally, the Wasserstein-1 distance \( W_1(P_{\text{real}}, P_{\text{synth}}) \) between the distribution of real images \( P_{\text{real}} \) and synthetic images \( P_{\text{synth}} \) is defined as:
\begin{align}
W_1(P_{\text{real}}, P_{\text{synth}}) = \inf_{\gamma \in \Gamma(P_{\text{real}},
P_{\text{synth}})} \mathbb{E}_{(x, y) \sim \gamma} \left[ \|x - y\| \right],
\end{align}
where \( \Gamma(P_{\text{real}}, P_{\text{synth}}) \) is the set of joint distributions with marginals \( P_{\text{real}} \) and \( P_{\text{synth}} \), and \( \|x - y\| \) is the Euclidean distance.

While this definition is general, in the context of our framework, we compute the Wasserstein distance not directly in the high-dimensional image space, but in the latent space derived from images. This approach offers computational efficiency and aligns with the operational space of diffusion models like Stable Diffusion. Computing the distance between latent representations of real and synthetic images allows us to measure the discrepancy between their underlying distributions in a lower-dimensional and potentially more semantically meaningful space. This measure enables the identification and exclusion of low-quality or irrelevant synthetic samples that deviate significantly from the real data distribution captured in the latent space.

Beyond Wasserstein distance, various other metrics like KL divergence \cite{kullback1951information}, Maximum Mean Discrepancy (MMD) \cite{gretton2006kernel}, Total Variation distance (TV)  \cite{billingsley2017probability}, and Fréchet Inception Distance (FID) \cite{Heusel2017GANs} are also commonly used for synthetic data evaluation. FID is notable as it relates to the Wasserstein distance between Gaussian distributions fitted to image features \cite{DowsonLandau1982Frechet}, so we choose Wasserstein distance. We recognize that the empirical performance and suitability of different evaluation metrics can vary depending on the specific dataset characteristics and the aspects of distribution similarity they capture \cite{Betzalel2024EvaluationMetrics}. However, the necessity of employing a rigorous metric for filtering remains paramount in our framework, given the inherent variability in quality of data generated by large models despite their vast generation capabilities. We further investigate the use of alternative filtering metrics, including MMD and TV distance, presenting comparative results on various datasets.


The proposed image generation and filtering framework, detailed in Algorithm \ref{alg:image_generation}, proceeds as follows. Let $\mathcal{X}$ denote the set of original images, with each image associated with a corresponding label $\mathcal{L}_x$. Synthetic images are generated, forming the set $\mathcal{Y}$. A VAE model is employed to map both original and generated images into their respective latent representations, $\mathcal{Z}_x$ and $\mathcal{Z}_y$. The primary objective is to obtain a high-quality subset of generated images, $\mathcal{Y}_{\text{filtered}}$, by filtering $\mathcal{Y}$. This filtering process assesses the similarity between the latent representations of the generated images ($\mathcal{Z}_y$) and the original images ($\mathcal{Z}_x$), utilizing the Wasserstein distance metric evaluated against a predefined threshold. The augmented dataset $\mathcal{X}_{\text{augmented}}$ is subsequently constructed by combining the original images $\mathcal{X}$ and the filtered synthetic images $\mathcal{Y}_{\text{filtered}}$. Comprehensive definitions for all notations, variables, and functions used in this algorithm are provided in Appendix \ref{sec:alg_image_notations}.

Our framework's effectiveness is grounded in the following theoretical guarantee, which bounds the generalization error when using filtered synthetic data. Let \( W_1(P, Q) \) denote the Wasserstein-1 distance between distributions \( P \) and \( Q \), and \( \mathfrak{R}_n(\mathcal{H}) \) be the Rademacher complexity of hypothesis class \( \mathcal{H} \).

\begin{theorem}[Generalization Error Bound] \label{thm:bound}
Suppose that the loss function \( \ell: \mathcal{H} \times \mathcal{Z} \to [0, M] \) is \( L_\ell \)-Lipschitz continuous, and the synthetic distribution \( P_{\text{synth}} \) satisfies \( W_1(P_{\text{synth}}, P_{\text{real}}) \leq \epsilon \), where \( W_1 \) denotes the Wasserstein distance. Then, for any hypothesis \( h \in \mathcal{H} \), with probability at least \( 1 - \delta \), the generalization error satisfies:
\end{theorem}
\begin{equation}
\begin{split}
    \mathbb{E}_{P_{\text{real}}}[\ell(h,z)] &\leq \mathbb{E}_{P_{\text{synth}}}[\ell(h,z)] + L_\ell \epsilon 
   + 2\mathfrak{R}_n(\mathcal{H}) + M \sqrt{\frac{\log(1/\delta)}{2n}}.
\end{split}
\end{equation}

Theorem~\ref{thm:bound} implies that controlling \( \epsilon \) (via Wasserstein filtering) directly reduces the generalization gap. Full proof is deferred to Appendix~\ref{app:proof}.

   
   
   

\begin{algorithm}
\caption{Image Generation and Filtering with Wasserstein Distance}
\label{alg:image_generation}
\begin{algorithmic}[1]
   \State \textbf{Input:} Dataset $\mathcal{X}$, labels $\mathcal{L}_x$, generation prompt $\mathcal{P}$, VAE model, Wasserstein distance threshold
   \State Generate images: $\mathcal{Y} \gets \text{GenerateImages}(\mathcal{X}, \mathcal{P}, \mathcal{L}_x)$
   \State Encode images to latent space: $\mathcal{Z}_x \gets \text{VAE}(\mathcal{X})$, $\mathcal{Z}_y \gets \text{VAE}(\mathcal{Y})$
   \For{each generated image in $\mathcal{Y}$}
       \State Calculate Wasserstein distance: $d(\mathcal{Z}_x, \mathcal{Z}_y)$
   \EndFor
   \State Select images with minimal Wasserstein distance: $\mathcal{Y}_{\text{filtered}} \gets \text{FilterImages}(\mathcal{Y}, \mathcal{Z}_x, \mathcal{Z}_y, \mathcal{P})$
   \State Augment dataset: $\mathcal{X}_{\text{augmented}} \gets \mathcal{X} \cup \mathcal{Y}_{\text{filtered}}$
\end{algorithmic}
\end{algorithm}

\subsection{Simulation Studies}

\subsubsection{Low-dimensional Linear Regression}

\begin{figure}[htbp]
    \centering
    \includegraphics[width=0.25\textwidth]{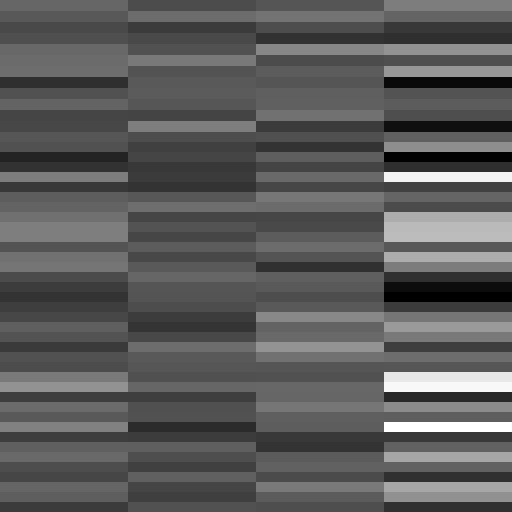}
    \caption{Grayscale representation \(\mathcal{F}_1\) of \(V_1\), generated using the transformation \(\mathcal{M}_i(v) = e^{0.05v}\). Each column corresponds to \((x_1, x_2, x_3, y)\) from right to left, satisfying \(y = 2X_1 - X_2 + 0.5X_3 + \varepsilon\).}
    \label{low_dimention_grayscale_representation}
\end{figure}

We evaluate our proposed methodology within the context of a linear regression framework. Specifically, we consider the model:
\begin{align}
y = X^T \beta + \varepsilon, \quad X \sim \mathcal{N}(0, I_p), \quad \varepsilon \sim \mathcal{N}(0, 1),
\end{align}
where \(p = 3\) and \(n = 100\). The true parameter vector is specified as \(\beta_0 = (2, -1, 0.5)^T\). To illustrate our framework, we partition the dataset into two subsets: \(V_1 = (X_{V_1}, Y_{V_1})\) containing 50 samples and \(V_2 = (X_{V_2}, Y_{V_2})\) comprising the remaining 50 samples. Subset \(V_1\) is used to generate the grayscale representation \(\mathcal{F}_1\) depicted in Figure~\ref{low_dimention_grayscale_representation}, while \(V_2\) serves as an independent reference for transfer learning.

The exponential function $\mathcal{M}_i(v) = e^{0.05v}$ is chosen as the mapping operator due to its monotonicity, smoothness, and ability to capture nonlinear patterns in synthetic data. Crucially, it satisfies Lipschitz continuity within bounded domains, ensuring numerical stability during optimization. For real-world data, where features often have heterogeneous scales, we apply column-wise max-min normalization to project values into $P[0,1]$. This preprocessing aligns with the Lipschitz properties of the exponential function and improves generalizability across varying measurement units.

The generation process utilizes the \texttt{StableDiffusion\allowbreak Img2ImgPipeline} with the \textit{stable-diffusion-xl-refiner-1.0} model. We set the diffusion strength parameter to range from 0.001 to 0.1 in increments of 0.001, with \texttt{guidance\_scale} fixed at 7.5. The prompt for generating images was specified as follows:

\begin{quote}
\textit{``Create a grayscale matrix image with four vertical columns, designed to visually represent complex data distributions. The image should feature a smooth gradient from left to right, mimicking statistical patterns.''}
\end{quote}

The Stable Diffusion process operates in a purely data-driven manner, without explicit knowledge of the underlying linear structure or distributional assumptions. This design ensures that the generated data augmentation remains unbiased, relying solely on the raw predictors \((X_1, X_2, X_3)\) and the response \(y\).

\begin{figure}[htbp]
    \centering
    \includegraphics[width=0.475\textwidth]{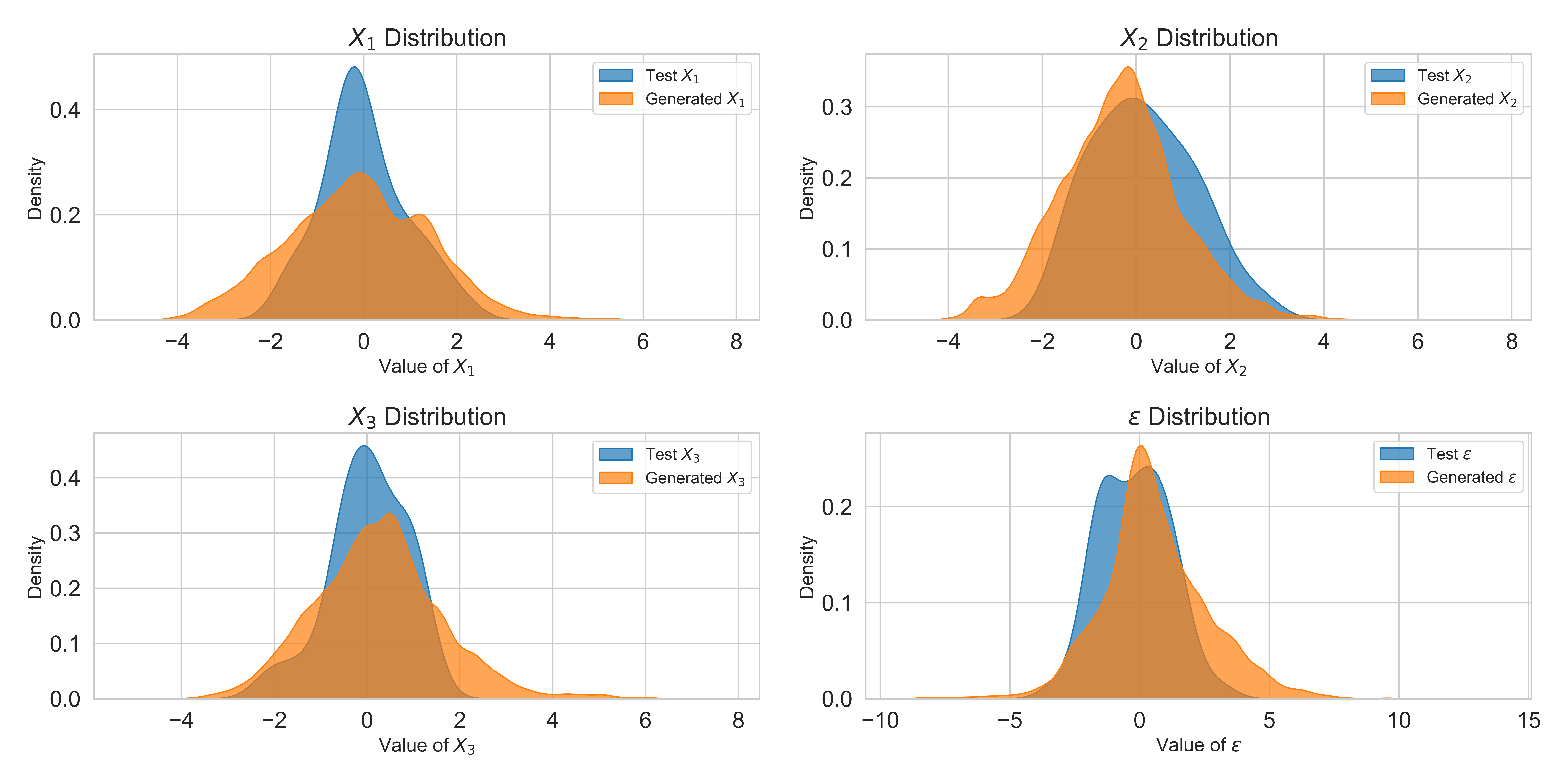}
    \caption{Density plots of the generated variables \([X_{\text{new}}, Y_{\text{new}}]_{1}^{(k)}\) and residuals derived from OLS estimation.}
    \label{lowdemensionkde}
\end{figure}

Applying the Stable Diffusion pipeline to \(V_1\), we generated a synthetic dataset \([X_{\text{new}}, Y_{\text{new}}]_{1}^{(k)}\), comprising 49,664 observations. A distributional analysis of the generated variables and residuals, derived through ordinary least squares (OLS) estimation, was conducted. Figure~\ref{lowdemensionkde} illustrates the density plots of the synthetic variables alongside residuals, highlighting the fidelity of the generated data to the original Gaussian distribution. 

To further validate the utility of the generated data, we employed the \texttt{glmtrans}, treating \(V_1\) as the target dataset and \(V_2\) as the source dataset. Using transferable detection, we filtered \(V_1\) to identify transferable data based on \(V_2\) as a reference, and vice versa. The combined dataset was then evaluated on a test set, leveraging prediction methods outlined in \texttt{glmtrans}. 

\begin{figure}[htbp]
    \centering
    \includegraphics[width=0.475\textwidth]{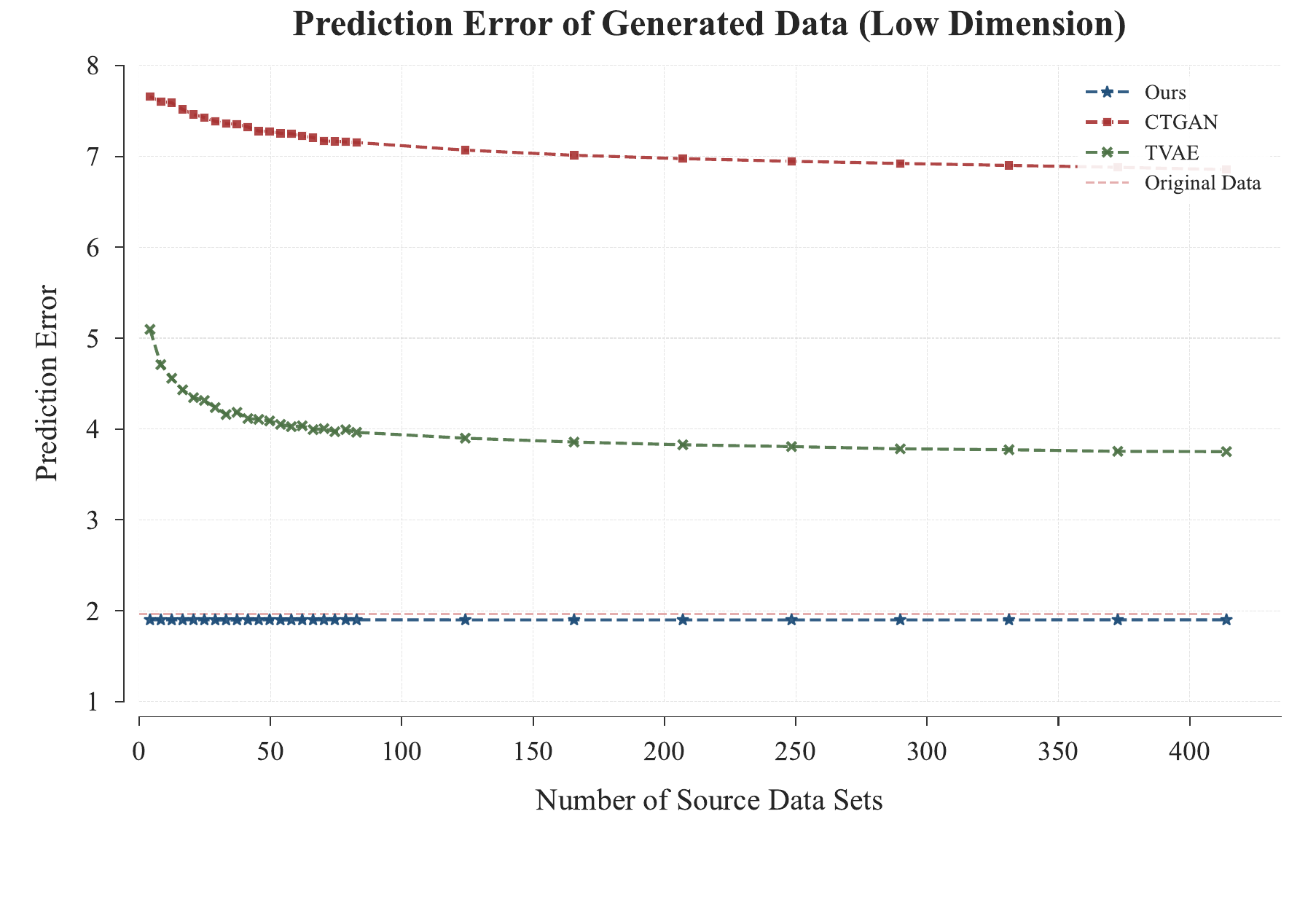}
    \caption{Prediction Error Comparison on Low-Dimensional Regression Simulation. "Ours" refers to data generated using SD-XL and filtered by Glmrtrans, CTGAN refers to data generated with CTGAN, and TVAE refers to data generated with TVAE. The red dashed line represents the prediction error of the original data. The meanings of CTGAN, TVAE, and Original Data remain consistent in subsequent figures.}
    \label{linear_pe}
\end{figure}

As a complementary evaluation, we combined multiple randomly sampled batches from the synthetic dataset with the original observations. The predictive performance was assessed using OLS, focusing on both the prediction error and squared error. As depicted in Figure~\ref{linear_pe}, the results demonstrate that the incorporation of synthetic data substantially reduces the prediction error, particularly in the context of low-dimensional linear models.

The improvements observed in prediction accuracy are primarily attributed to the prior information embedded within the generated data. As the sample size increases through repeated random sampling, the prediction error decreases, reflecting the effective integration of this prior knowledge. However, since the prior information available in the original dataset is inherently finite, the data that meaningfully reduce prediction error are limited. Consequently, the overall improvement eventually stabilizes as additional synthetic data contribute diminishing returns.

\subsubsection{High-dimensional Linear Regression}

We apply the proposed methodology to a high-dimensional linear regression framework as follows. Consider the model:
\begin{align}
y = X ^T \beta + \varepsilon, \quad X \sim \mathcal{N}(0, I_p), \quad \varepsilon \sim \mathcal{N}(0, 1),
\end{align}
where the sample size is \( n = 200 \) and the number of co-variates is \( p = 511 \). The true parameter vector \( \beta \) is specified such that the first three entries are \( [2, -1, 0.5] \), while all remaining entries are zero. Details are listed in Appendix \ref{High-dimensional Linear Regression}.

\begin{figure}[htbp]
    \centering
    \includegraphics[width=0.475\textwidth]{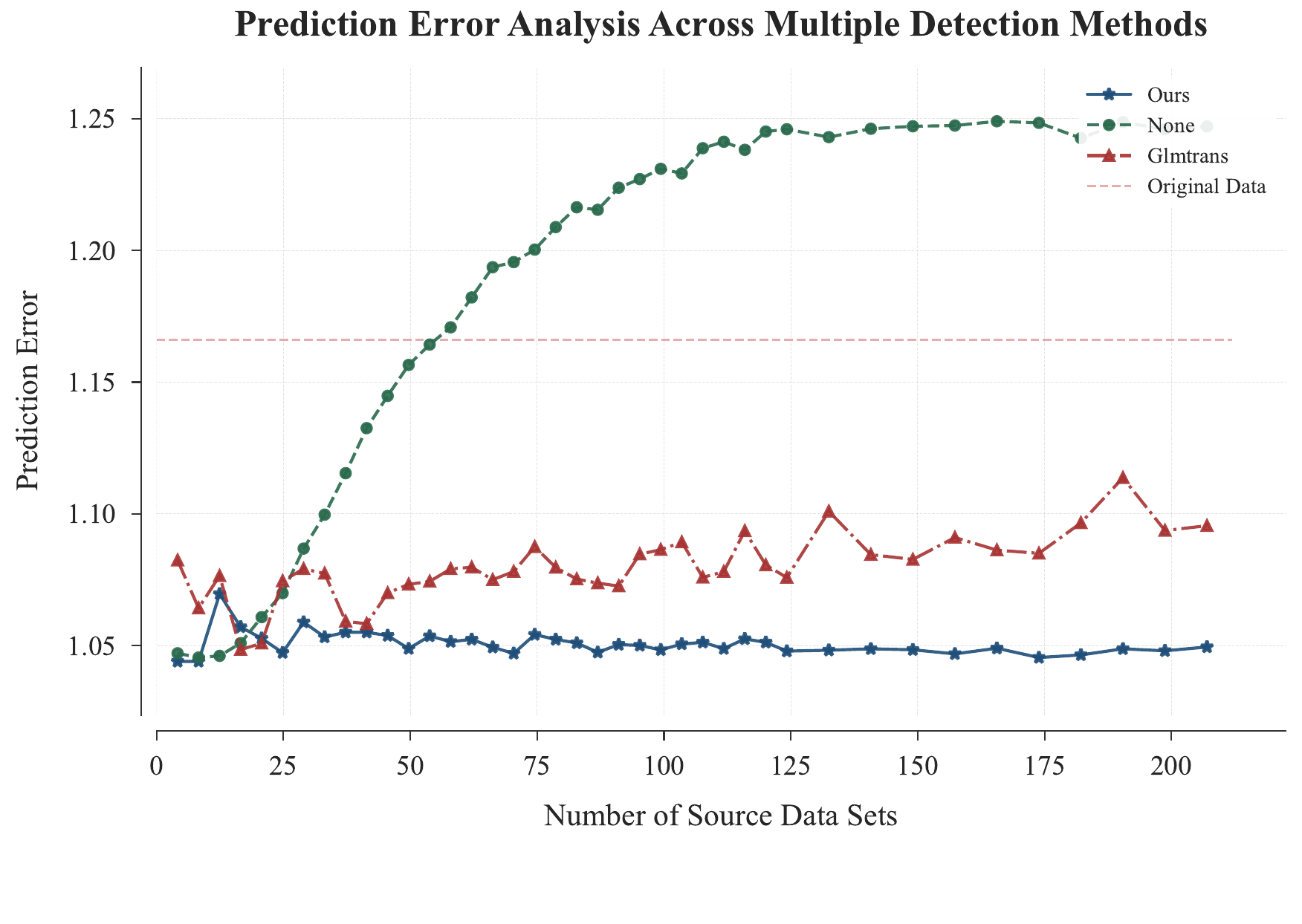}
    \caption{Prediction Error Comparison on High-Dimensional Linear Regression Simulation. "Ours" applies p-value-based filtering, "None" uses no filtering, and "Glmtrans" denotes the Glmtrans method. The results demonstrate the effectiveness and stability of our filtering method here.}
    \label{high_pe}
\end{figure}

To identify transferable sources, we employ the \texttt{hdtrd} package with a parameter set of \(\delta_0 = 2\). We then input the detected sources into \texttt{ glmtrans} to calculate \(\hat{\beta}\). For each iteration, we randomly select subsets of data to construct source datasets, each containing 100 samples. This process is repeated 100 times, and the results are averaged across all iterations. As demonstrated in Figure \ref{high_pe}, the proposed method shows superior performance compared to existing approaches. Specifically, in high-dimensional linear scenarios, our method consistently identifies transferable data samples while effectively mitigating negative transfer phenomena. Moreover, even in high-dimensional settings where the feature space vastly exceeds the number of samples, Stable Diffusion consistently generates data that reduce prediction error, leading to a marked improvement in prediction accuracy. This enhancement is driven by the effective incorporation of prior information embedded in the generated data. In particular, as the number of randomly selected samples increases, the prediction error decreases more noticeably, reflecting the influence of this prior knowledge. However, given the finite nature of the available prior information, the reduction in prediction error becomes asymptotically limited, and the improvement ultimately stabilizes as the sample size grows.

\subsubsection{High-dimensional Generalized Linear Regression}

We apply the proposed methodology to a high-dimensional generalized linear regression framework, specifically a logistic regression model. Consider the model:
\begin{align}
P(y = 1 \mid X) = \frac{1}{1 + e^{-X^T\beta}}, \quad X \sim \mathcal{N}(0, I_p),
\end{align}
where the sample size is \( n = 200 \) and the number of covariates is \( p = 511 \). The true parameter vector \( \beta \) is specified so that the first three entries are \( [2, -1, 0.5] \), while all remaining entries are zero.

We similarly extracted a subset \( V_1 \) of size \( n = 100 \) to generate the grayscale representation \( \mathcal{F} \), as shown in Figure \ref{high_dimention_grayscale_representation}. The remaining 100 samples are denoted as \( V_2 \). 

\begin{figure}[htbp]
    \centering
    \includegraphics[width=0.475\textwidth]{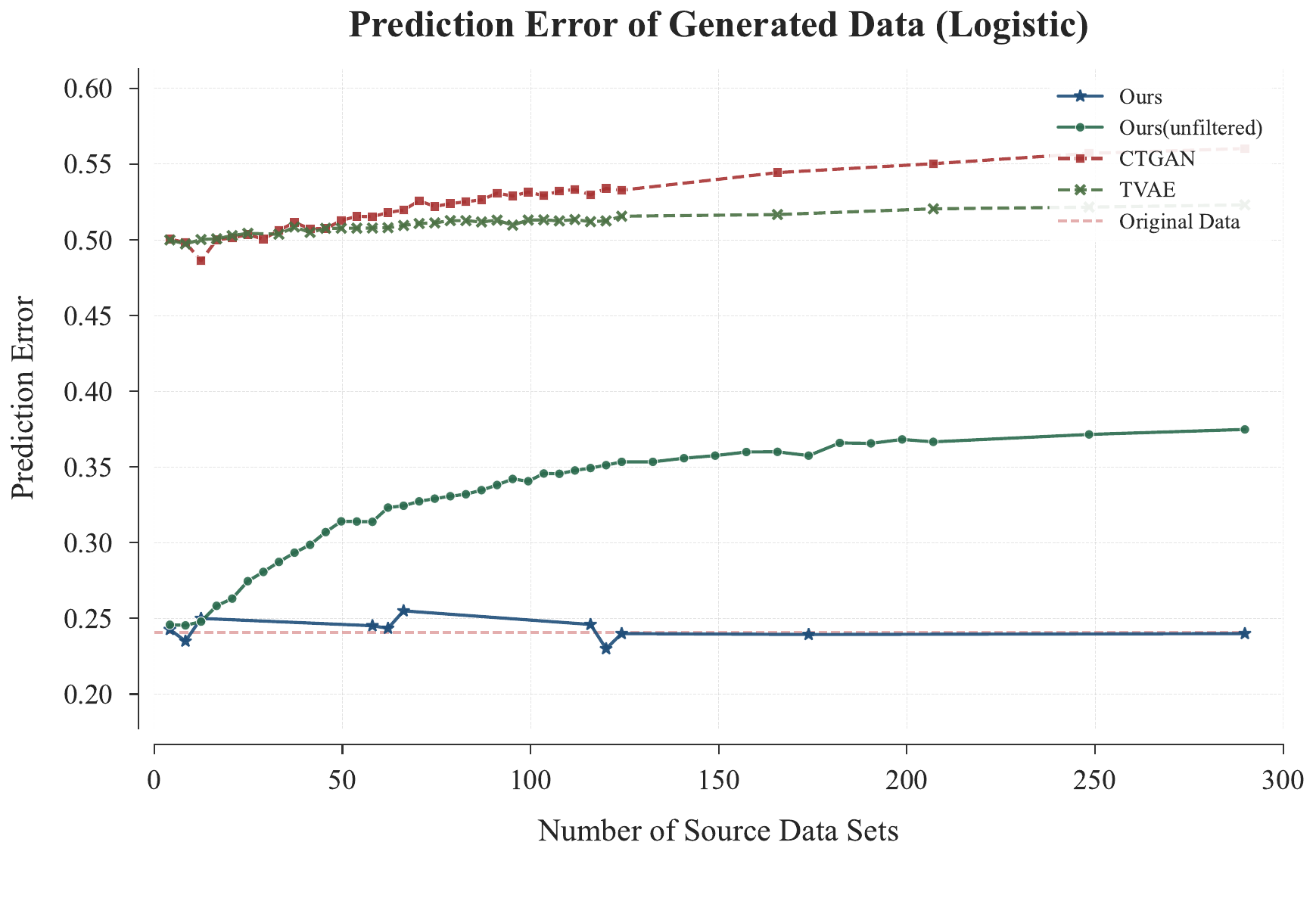}
    \caption{Prediction error of generated data in high-dimensional generalized linear models."Ours" refers to data generated using Glmtrans with filtering.This figure highlights the necessity of filtering.}
    \label{sig_pe}
\end{figure}

In the image reconstruction process, we adopt a thresholding strategy for the response variable \( y \). Specifically, the pixel value in the last column of the generated grayscale matrix is compared to a threshold of 0.5. If the value of the pixels exceeds 0.5, we classify \( y = 1 \); otherwise, \( y = 0 \). This binary classification is consistent with the logistic regression framework, where the model predicts the probability of \( y = 1 \).Details are listed in Appendix \ref{highglm}.

As illustrated in Figure \ref{sig_pe}, the results show , even in high-dimensional settings where the feature space vastly exceeds the number of samples, Stable Diffusion consistently generates data that reduces prediction error, leading to a marked improvement in prediction accuracy. This enhancement is primarily attributed to the effective incorporation of prior information embedded in the generated data. Notably, as the number of randomly selected samples increases, the prediction error decreases more noticeably, reflecting the contribution of the prior knowledge embedded in the generated data. However, due to the finite nature of the available prior information, the reduction in prediction error becomes increasingly limited as the sample size grows, and the improvement ultimately stabilizes.

\section{Real World Experiments} \label{experiments}

\subsection{Boston House Price Dataset}
We propose a symmetric reversible mapping framework for structured data augmentation, applied to the Boston Housing dataset. The framework partitions the data into sets ($V_1$, $V_2$) and each employs min-max normalization to generate grayscale representations. Details are in Appendix(\ref{boston}).

\begin{figure}[htbp]
    \centering
    \includegraphics[width=0.475\textwidth]{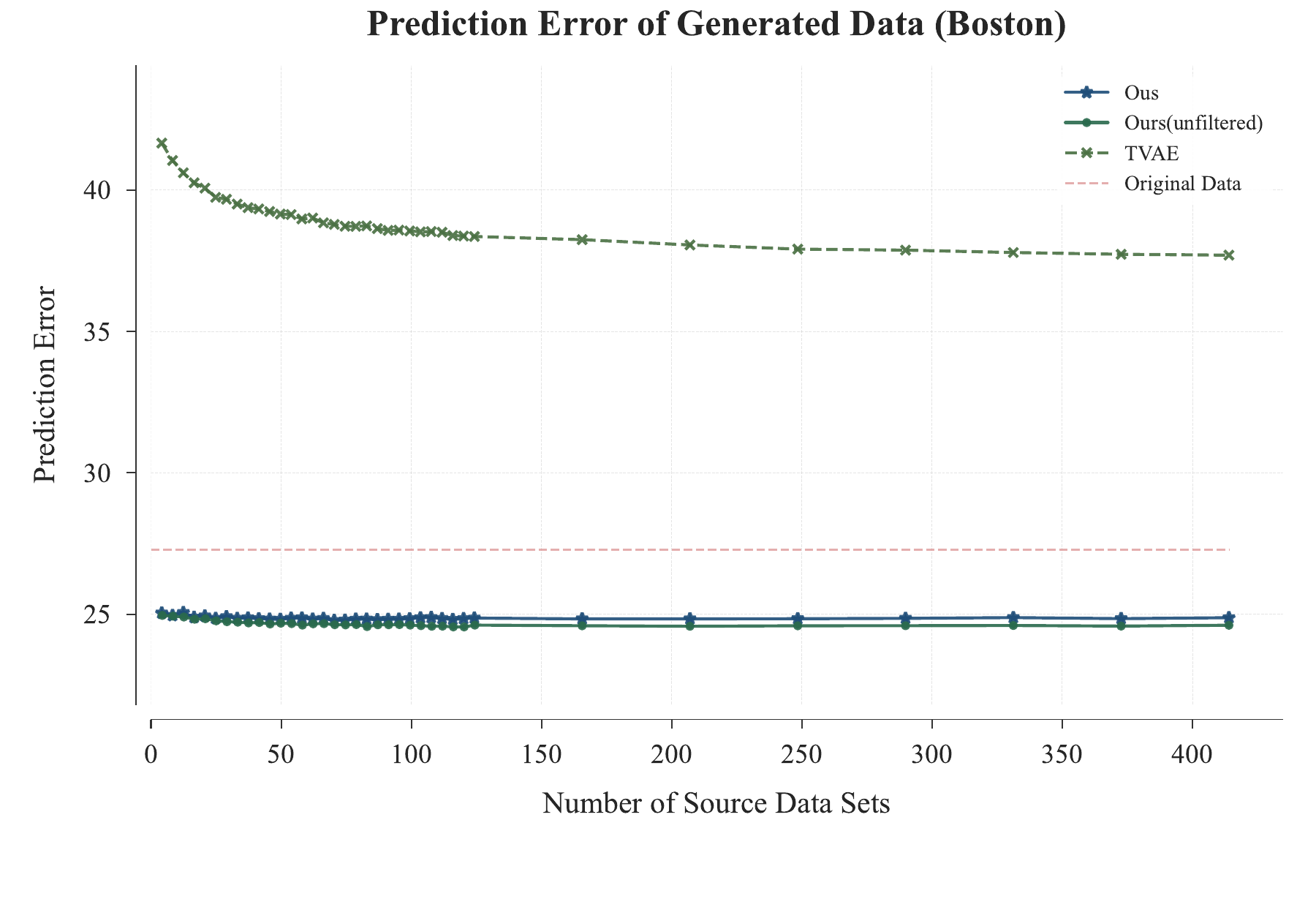}
    \caption{Prediction Error Comparison on Boston Dataset. Due to CTGAN's limitations in effectively modeling mixed discrete-continuous tabular data, its prediction error reaches approximately 70. We exclude it from the plot for clarity.}
    \label{boston_result}
\end{figure}

The results, illustrated in Figure~\ref{boston_result}, demonstrate that the generative model produces informative synthetic data, as evidenced by the reduction in prediction error with increasing data volume. This improvement is bounded, as the error asymptotically approaches a lower limit, reflecting the inherent limitations of synthetic data in fully capturing the underlying data distribution. 

\subsection{GTEx Data} 
We adapt our symmetric mapping framework to high-dimensional genomic regression using Alzheimer's disease-related gene expression data from 13 brain tissues \cite{carithers2015genotype}. The method encodes APOE (response variable) and 118 AD-associated predictors into min-max normalized grayscale matrices. Transfer learning is implemented via package \texttt{hdtrd} and \texttt{glmtrans} for comparison, with 100-sample source subsets validated through 100 repetitions. Details are in Appendix \ref{gtex}.

\begin{figure}[htbp]
    \centering
    \includegraphics[width=0.475\textwidth]{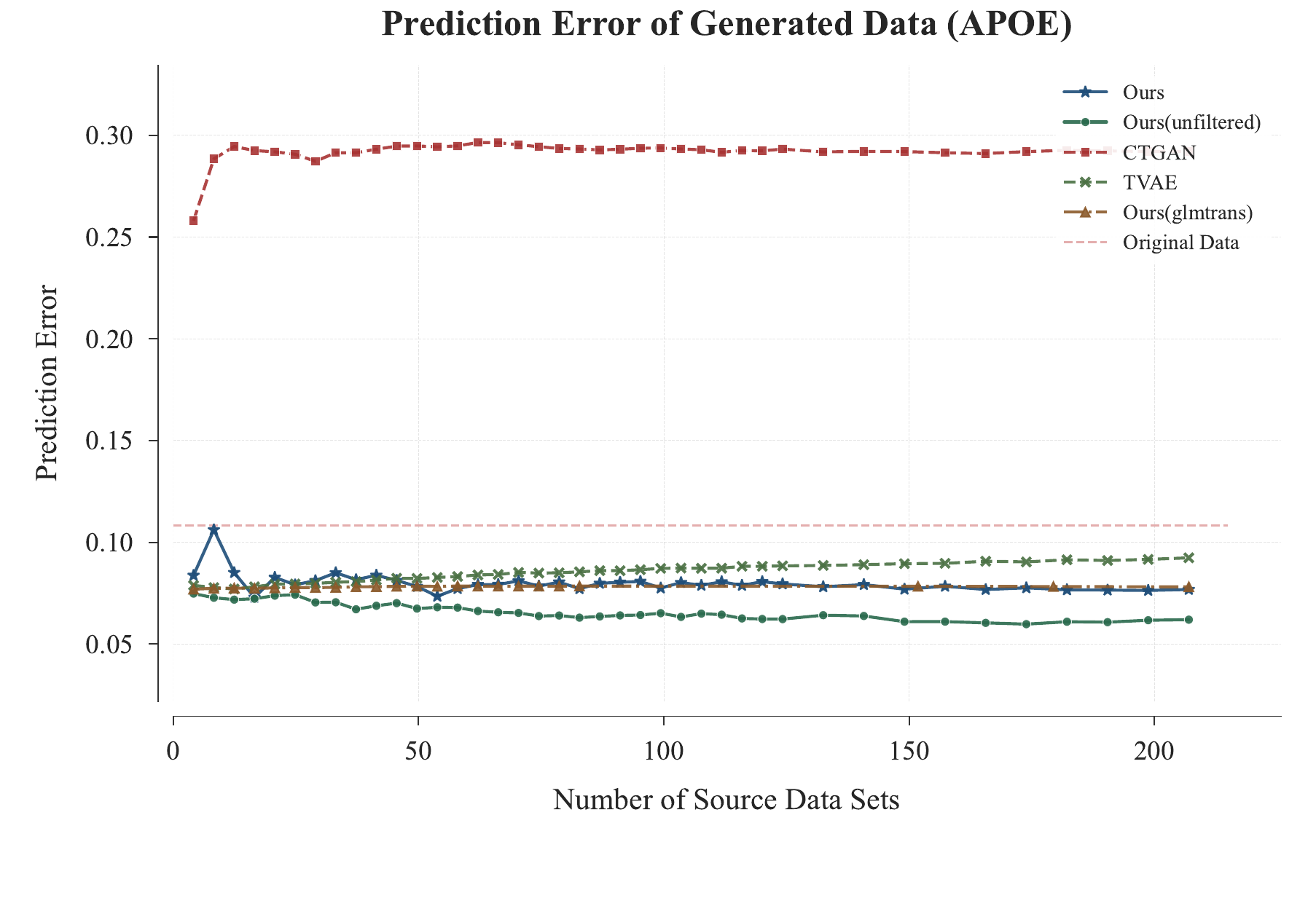}
    \caption{Prediction Error of Generated Data Based on GTex Data Set.On moderate dimension dataset, the difference between (unfiltered) and "Ours" is negligible. For consistency, we adopt the filtered results.}
    \label{apoe_result}
\end{figure}

Figure~\ref{apoe_result} reveals two distinct operational regimes: synthetic data initially reduces prediction error by 18.7\% compared to baseline, followed by asymptotic convergence. 

\subsection{German Credit Dataset}

We extend our symmetric reversible mapping framework to high-dimensional logistic regression using the German Credit dataset.The results in such high-dimensional settings highlight the advantages of our method in generating high-quality synthetic data compared to other approaches. Details are listed in Appendix \ref{german}.

\begin{figure}[htbp]
    \centering
    \includegraphics[width=0.475\textwidth]{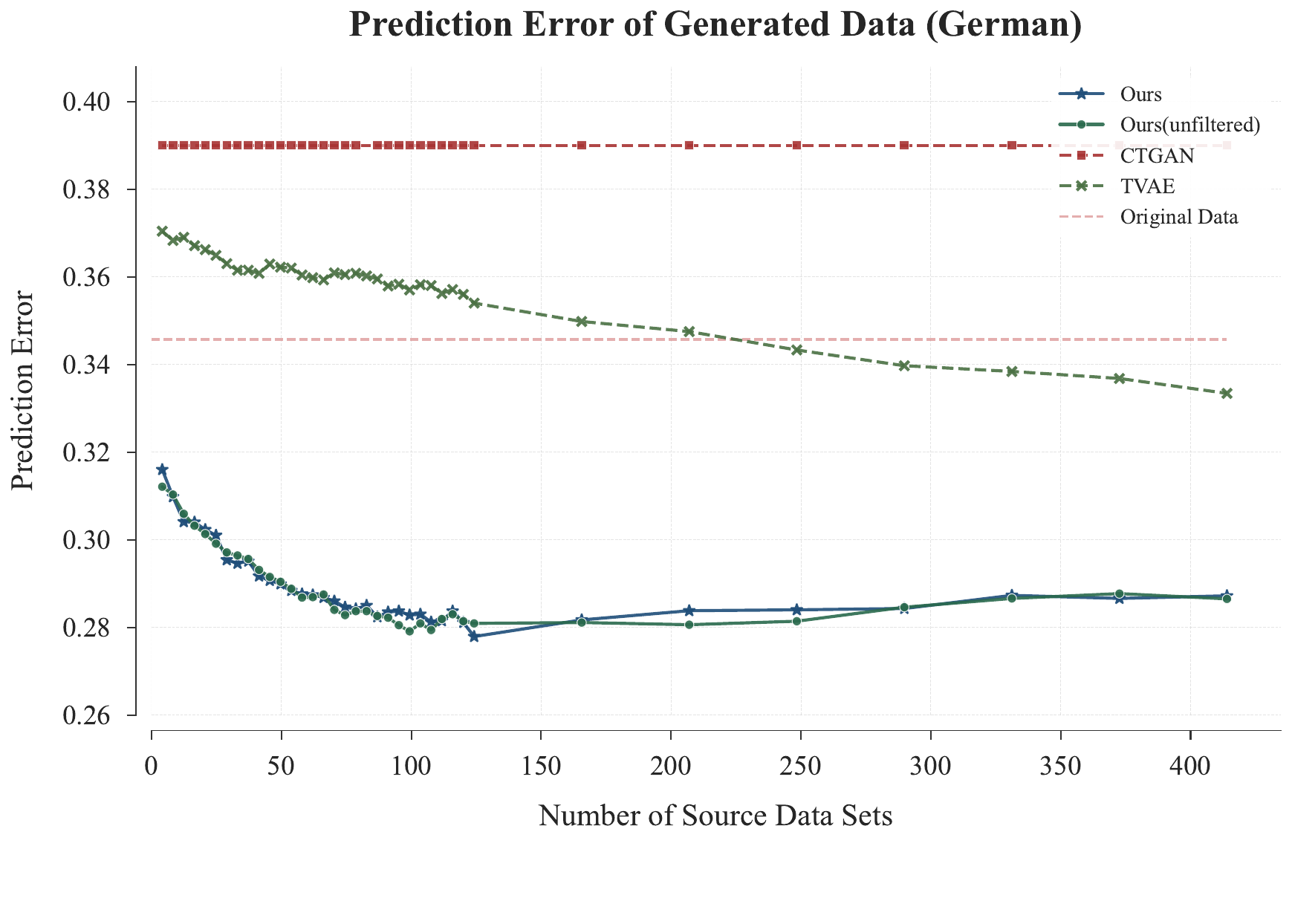}
    \caption{Prediction error of generated data based on German Credit Data Set.Ours means generated by our method and filtered by Glmtrans. "Ours" means generated by our method and filtered by Glmtrans. On moderate dimension dataset, the difference between (unfiltered) and "Ours" is negligible. For consistency, we adopt the filtered results. }
    \label{german_result}
\end{figure}


\subsection{MNIST Dataset}
The MNIST dataset was compiled by the National Institute of Standards and Technology (NIST) and consists of 60,000 training images and 10,000 testing images of handwritten digits. This dataset has become an important benchmark in the fields of machine learning and deep learning, widely used for algorithm testing and evaluation \cite{lecun1998gradient,srivastava2014dropout}.

We designed a simple Convolutional Neural Network (CNN) as a baseline for our experiments. With 600 training samples, the accuracy on the test set reached approximately 90\%. Building upon this baseline, we fixed the 600 training samples and applied stable diffusion to each image individually. After performing the diffusion, we selected the top 80\% of the diffused images based on the Wasserstein distance compared to the original image in latent space. These selected diffused images were then merged with the fixed dataset of 600 samples and fed into the same CNN architecture. After selection and merging, accuracy can be increased to around 95\%. Details and results can be seen in \ref{mnitst}.

\subsection{CIFAR-10 Dataset}
The CIFAR-10 dataset, compiled by the Canadian Institute for Advanced Research, consists of 60,000 color images in 10 different classes, with 50,000 training images and 10,000 testing images. Each image is of size 32x32 pixels and comes in RGB color format. This dataset is widely used as a benchmark in machine learning and computer vision, particularly for evaluating image classification algorithms \cite{krizhevsky2012imagenet,ioffe2015batch,he2016deep}.

For our experiments, we utilized ResNet-20 \cite{he2016deep} as a reference architecture for our baseline model. Using a fixed set of 1,000 training samples (100 per selected class).Building upon this baseline, we applied stable diffusion to each image individually in the training set. After performing the diffusion, we selected the top 60\% of the diffused images based on the Wasserstein distance compared to the original image set in latent space. These selected diffused images were then merged with the fixed dataset of 2,500 samples and fed into the same architecture. Table \ref{tab:cifar10-key} show that augmentation improves over the baseline, with Wass, MMD, and TV showing comparable
performance, showing consistent trends. Full results and details available in Appendix \ref{tab:cifar10}.

\begin{table}[t]
\centering
\caption{Key Performance Comparison on CIFAR10 (Selected Generations)}
\label{tab:cifar10-key}
\begin{tabular}{@{}clrrrr@{}}
\toprule
\textbf{Gen} & \textbf{Model} & \textbf{Acc} & \textbf{Prec} & \textbf{Rec} & \textbf{F1} \\
\midrule
\multirow{4}{*}{\textbf{6}} 
& Baseline & 38.86 & 38.96 & 38.86 & 38.76 \\
& Wass & 41.97 & 42.12 & 41.97 & 41.79 \\
& MMD & \textbf{44.65} & \textbf{44.52} & \textbf{44.65} & \textbf{44.32} \\
& TV & 42.75 & 43.29 & 42.75 & 42.73 \\
\midrule
\multirow{4}{*}{\textbf{12}} 
& Baseline & 38.18 & 38.44 & 38.18 & 38.02 \\
& Wass & \textbf{43.88} & \textbf{43.67} & \textbf{43.88} & \textbf{43.55} \\
& MMD & 43.43 & 42.95 & 43.43 & 43.02 \\
& TV & 42.70 & 42.24 & 42.70 & 42.27 \\
\midrule
\multirow{4}{*}{\textbf{20}} 
& Baseline & 38.92 & 38.93 & 38.92 & 38.67 \\
& Wass & 43.52 & 43.15 & 43.52 & 43.15 \\
& MMD & \textbf{43.97} & \textbf{43.45} & \textbf{43.97} & \textbf{43.27} \\
& TV & 42.59 & 42.41 & 42.59 & 42.30 \\
\bottomrule
\end{tabular}
\end{table}

\subsection{CIFAR-100 Dataset}
Our experiments employ a frozen ResNet-18 (ImageNet weights) with only \texttt{layer4} and classifier trained (Adam, lr=5e-5, dropout=0.5). The baseline achieves 68.2\% accuracy with 1,000 samples (50 per class). Augmenting via Stable Diffusion XL - generating 10 variants per image (5 at \texttt{strength=0.15} for fidelity, 5 at \texttt{strength=0.8} for diversity). Filtering through different methods yields performance nearly identical to unfiltered augmentation, with minimal differences nearly 1\%, as CIFAR-100 is well-represented in Stable Diffusion's pretraining, reducing generation anomalies. However, for fine-grained classification tasks, we recommend filtering to enhance robustness. Full resutls and details available in Appendix~\ref{cifar100}).

\subsection{ISIC Dataset}
The ISIC Dataset~\cite{brinker2018skin} consists of skin cancer images across 7 classes, with 10,015 original training samples. We employed a ResNet-20 architecture, training 1,257 images on only the final layers using the Adam optimizer (lr=0.001, dropout=0.5). The baseline model achieved an average accuracy of 52.32\%. For augmentation, we generated varying numbers of images (Gen) per original sample using Stable Diffusion XL at \texttt{strength=0.15} and \texttt{strength=0.8}. The Wasserstein filtering method was applied to retain the top 60\% of generated images based on their similarity to the original images in latent space.

Performance metrics (\%) are reported in Table~\ref{table:key-isic2018}, showing that Wasserstein filtering consistently improves accuracy, precision, recall, and F1-score over the baseline and unfiltered augmentation. This enhancement is crucial for fine-grained classification tasks like skin cancer diagnosis, where filtering reduces generation anomalies and improves model robustness. Full results and details are in Appendix \ref{ISIC}

\begin{table}[htbp]
\centering
\caption{Performance on ISIC Dataset at Selected Generations (Gen=6, 18, 24), with Baseline Average.}

\begin{tabular}{clrrrr}
\toprule
\textbf{Gen} & \textbf{Model} & \textbf{Acc} & \textbf{Prec} & \textbf{Rec} & \textbf{F1} \\
\midrule
\multirow{2}{*}{\centering\textbf{6}} & Augmented      & 45.71 & 45.69 & 45.71 & 44.48 \\
                                      & Wass           & \textbf{57.14} & \textbf{63.81} & \textbf{57.14} & \textbf{56.76} \\
\midrule
\multirow{2}{*}{\centering\textbf{18}} & Augmented     & 48.57 & 61.77 & 48.57 & 47.52 \\
                                       & Wass          & \textbf{58.57} & \textbf{57.51} & \textbf{58.57} & \textbf{57.38} \\
\midrule
\multirow{2}{*}{\centering\textbf{24}} & Augmented     & 55.71 & 52.91 & 55.71 & 51.67 \\
                                       & Wass          & \textbf{64.29} & \textbf{65.37} & \textbf{64.29} & \textbf{63.91} \\
\midrule
\multicolumn{2}{c}{\textbf{Baseline (Avg.)}} & 52.32 & 56.64 & 52.32 & 51.88 \\
\bottomrule
\end{tabular}
\label{table:key-isic2018}
\end{table}

\subsection{Cassava Leaf Disease Dataset}
We evaluate the effectiveness of filtered data augmentation on the \href{https://www.kaggle.com/competitions/cassava-leaf-disease-classification/data}{\textit{Cassava Leaf Disease Classification Dataset}}, which consists of varying training sizes (Size) of 5 classes. For each original image, 10 augmented images are generated using Stable Diffusion XL, with 5 images at a strength of 0.2 to preserve fidelity and 5 at a strength of 0.6 to enhance diversity. The models employ a pretrained EfficientNet-B0 architecture with ImageNet weights, where feature extraction layers are frozen, and the classifier is fine-tuned using the Adam optimizer (learning rate $10^{-4}$, batch size 32, dropout 0.5). This experimental setup allows us to assess the impact of augmentation strategies under controlled conditions. Tab \ref{key-cassava} shows that filtering is effective in such fine-grained task categorization. Full results with different filtering methods and various sizes and details are in \ref{Cassava}.

\begin{table}[htbp]
\centering
\caption{Evaluation of Wasserstein-Filtered Data Augmentation on Cassava Leaf Disease Dataset (Sizes 250 and 500). Baseline: original samples; None: unfiltered augmentation (mean of 100\% tolerance); Wass: filtered data at 20\%, 60\%, 80\% tolerance.} 
\begin{tabular}{clrrrr}
\toprule
\textbf{Size} & \textbf{Model} & \textbf{Acc} & \textbf{Prec} & \textbf{Rec} & \textbf{F1} \\
\midrule
\multirow{2}{*}{\centering\textbf{250}} 
  & Baseline & 0.350 & 0.351 & 0.350 & 0.337 \\
  & None & \textbf{0.387} & 0.398 & 0.387 & 0.367 \\
\cmidrule(lr){2-6}
  & Wass-20 & \textbf{0.396} & 0.392 & 0.396 & 0.384 \\
  & Wass-60 & 0.384 & 0.396 & 0.384 & 0.364 \\
  & Wass-80 & 0.388 & 0.396 & 0.388 & 0.381 \\
\midrule
\multirow{2}{*}{\centering\textbf{500}} 
  & Baseline & 0.414 & 0.415 & 0.414 & 0.411 \\
  & None & \textbf{0.465} & 0.465 & 0.465 & 0.460 \\
\cmidrule(lr){2-6}
  & Wass-20 & \textbf{0.476} & 0.477 & 0.476 & 0.472 \\
  & Wass-60 & 0.452 & 0.461 & 0.452 & 0.436 \\
  & Wass-80 & 0.440 & 0.439 & 0.440 & 0.428 \\
\bottomrule
\end{tabular}
\label{key-cassava}
\end{table}

\section{Conclusion} 

In this study, we propose a novel data augmentation framework that begins by transforming numerical datasets into grayscale images. These images are then processed using the \textit{Stable Diffusion} model to generate synthetic data, which is subsequently reverted back into the original numerical space. The effectiveness of this framework is demonstrated by rigorous algorithmic evaluations, which confirm that the synthetic data generated often contains instances that significantly improve prediction accuracy in a variety of tasks.

To evaluate the quality of the synthetic data, we apply \(p\)-value-based hypothesis testing. This statistical method allows us to filter out low-quality synthetic data and retain only those instances that meaningfully contribute to improving prediction error. By integrating the selected synthetic data with the original dataset, we are able to enhance model performance and improve statistical estimations.

Our results underscore the utility of large generative models for data augmentation, as they can generate useful synthetic data that enhances model predictions. However, we also identify a fundamental limitation: while generative models like \textit{Stable Diffusion} can produce vast quantities of synthetic data, the improvement in model performance diminishes as more data is generated. This is due to the finite amount of information contained within both the original dataset and the generative model, which ultimately constrains the creation of novel and informative data.

Moreover, in the context of image generation, we demonstrate that replacing the traditional \(p\)-value-based evaluation with the Wasserstein distance yields similar improvements in performance. This suggests that the approach is adaptable and effective even when the evaluation metric is changed, highlighting the versatility of the method across different domains and types of data.

Overall, this work provides valuable insights into the potential and limitations of using large generative models for data augmentation in predictive modeling. Future research could explore further refinements in the generation process and consider complementary methods for enhancing data in high-dimensional and complex domains.

\section*{Limitations} 
\label{sec:limitations}

Our framework significantly advances synthetic data augmentation for predictive modeling, yet certain limitations warrant consideration for future enhancements. First, the methodology generates a large volume of synthetic data, which is subsequently filtered using metrics such as Wasserstein distance or p-value-based criteria to ensure quality. While effective, this post-generation filtering incurs substantial computational costs. Future research could investigate mechanisms to embed quality assurance within the generation process, potentially through refined generative models or adaptive prompting strategies, thereby improving computational efficiency without compromising the framework's robust performance.

Additionally, our empirical validation primarily focuses on generalized linear models and select image datasets, offering controlled settings to demonstrate efficacy across low- and high-dimensional scenarios. However, these contexts may not fully encapsulate the complexities of non-linear models or advanced deep learning architectures. Extending the framework to a broader spectrum of statistical and machine learning paradigms represents a valuable direction for further exploration, leveraging the strong theoretical and empirical foundation established herein.

Lastly, for tabular data, our reliance on cross-validation-based metrics provides robust evaluation without requiring additional validation sets. Nevertheless, the lack of a standardized metric selection protocol introduces variability across applications. Developing a unified evaluation framework could enhance the generalizability of our approach. These limitations underscore opportunities for refinement while affirming the substantial contributions of our methodology to data augmentation research.



\newpage
\appendix
\onecolumn
\section{Appendix}
\subsection{Detailed Notation Definitions for Algorithm \ref{alg:dual_transfer}} 
\label{sec:alg_dual_transfer_notations} 

To provide a complete and formal description of Algorithm \ref{alg:dual_transfer}, this section details all the notations, variables, and functions used.

\begin{itemize}
    \item \textbf{Inputs:}
    \begin{itemize}
        \item $\mathcal{S}_1, \mathcal{S}_2$: These denote the two source domains providing data for transfer learning. In our framework, these are typically generated or derived from subsets of the original dataset (e.g., from $V_1$ and $V_2$ mentioned previously). Each $\mathcal{S}_i$ is a dataset, potentially represented as a matrix or collection of data points.
        \item $\mathcal{T}_1, \mathcal{T}_2$: These represent the corresponding target domains used for model adaptation. Like the source domains, they are also derived from subsets of the original data. $\mathcal{T}_i$ provides the target distribution characteristics for adapting from the source domains.
        \item $\mathcal{D}_{\text{test}}$: This is an independent test set used exclusively for evaluating the prediction performance of the adapted models. It is a dataset $[X_{\text{test}}, Y_{\text{test}}]$.
        \item $\mathcal{P} = \{\rho_i\}_{i=1}^n$: This set contains predefined sampling ratios. Each $\rho_i \in [0, 1]$ controls the proportion of data sampled from the source domains $\mathcal{S}_1$ and $\mathcal{S}_2$ used in each adaptation iteration. $n$ is the number of distinct ratios considered.
        \item $b$: This specifies the batch size used when splitting the sampled source data into smaller batches for the adaptation process. $b$ is a positive integer.
    \end{itemize}
    \item \textbf{Variables:}
    \begin{itemize}
        \item $\varepsilon_0$: Represents the initial baseline error. It is computed as the Mean Squared Error (MSE) of a Lasso regression model (see Functions below) trained on the combined target domains ($\mathcal{T}_1 \cup \mathcal{T}_2$) and evaluated on $\mathcal{D}_{\text{test}}$. $\varepsilon_0$ is a scalar value.
        \item $\tilde{\mathcal{S}}_1, \tilde{\mathcal{S}}_2$: These are temporary subsets of data sampled from $\mathcal{S}_1$ and $\mathcal{S}_2$, respectively, using the current sampling ratio $\rho$ within each iteration $k$. Their size is proportional to $\rho$ and the size of $\mathcal{S}_1, \mathcal{S}_2$.
        \item $\mathcal{B}_1, \mathcal{B}_2$: These denote mini-batches of size $b$ created by splitting the sampled subsets $\tilde{\mathcal{S}}_1$ and $\tilde{\mathcal{S}}_2$. Each $\mathcal{B}_i$ is a dataset of $b$ data points.
        \item $f_{1|2}, f_{2|1}$: These represent the adapted models learned during the process. $f_{1|2}$ is a model adapted to map data from batches of $\tilde{\mathcal{S}}_1$ to the domain characteristics of $\mathcal{T}_2$, and $f_{2|1}$ maps data from batches of $\tilde{\mathcal{S}}_2$ to the domain of $\mathcal{T}_1$. These are typically regression models (e.g., linear models).
        \item $\hat{y}$: Represents the combined prediction for an input $x \in \mathcal{D}_{\text{test}}$. It is calculated as the average of the predictions from the two adapted models for input $x$: $\hat{y}(x) = \frac{1}{2}(f_{1|2}(x) + f_{2|1}(x))$.
        \item $\mathcal{E}_{\text{comb}}$: A set used to collect the Mean Squared Error (MSE) values obtained from the combined predictions $\hat{y}$ on the test set $\mathcal{D}_{\text{test}}$ across multiple adaptation iterations $k$ for a fixed $\rho$. $\mathcal{E}_{\text{comb}}$ is a set of scalar MSE values.
        \item $\mathcal{E}_{\text{adapt}}$: A set collecting values of the adaptability metric $d(f_{1|2}, f_{2|1})$ (see Functions below) for each iteration $k$. It quantifies the similarity or agreement between the two adapted models. $\mathcal{E}_{\text{adapt}}$ is a set of scalar values.
        \item $\bar{\varepsilon}_\rho, \bar{\Delta}_\rho$: These are the average prediction error (mean of $\mathcal{E}_{\text{comb}}$) and the average adaptability metric (mean of $\mathcal{E}_{\text{adapt}}$) respectively, computed for a specific sampling ratio $\rho$ over all iterations $k$. These are scalar values.
        \item $\rho^*$: The optimal sampling ratio determined from the set $\mathcal{P}$ that minimizes the average prediction error $\bar{\varepsilon}_\rho$. It is the element $\rho \in \mathcal{P}$ that yields the smallest $\bar{\varepsilon}_\rho$.
    \end{itemize}
    \item \textbf{Functions:}
    \begin{itemize}
        \item $\text{Lasso}(\mathcal{D}_{\text{train}}, \mathcal{D}_{\text{eval}})$: This function fits a Lasso regression model (L1-regularized linear regression) on the training dataset $\mathcal{D}_{\text{train}}$ and returns its Mean Squared Error when evaluated on the dataset $\mathcal{D}_{\text{eval}}$. Input $\mathcal{D}_{\text{train}}$ is a dataset $[X_{\text{train}}, Y_{\text{train}}]$.
        \item $\text{BatchSplit}(\mathcal{D}, b)$: This function divides a given dataset $\mathcal{D}$ into smaller batches, each of size $b$. It typically returns a list or collection of data batches.
        \item $\text{Adapt}(\mathcal{D}_{\text{target}}, \mathcal{D}_{\text{source\_batch}})$: This function trains a transfer learning model. It takes data from a source batch $\mathcal{D}_{\text{source\_batch}}$ and adapts it to align with the characteristics of the target domain $\mathcal{D}_{\text{target}}$. The specific adaptation method depends on the implementation. It returns the resulting adapted model.
        \item $\text{MSE}(y_{\text{pred}}, y_{\text{true}})$: This standard function computes the Mean Squared Error between a set of predicted values $y_{\text{pred}}$ and the corresponding true values $y_{\text{true}}$. Both inputs are vectors of the same dimension.
        \item $d(m_1, m_2)$: This function measures the adaptability or statistical similarity between two adapted models, $m_1$ and $m_2$. The specific implementation of this metric is crucial and should be described in detail elsewhere (e.g., in the experimental setup). It returns a scalar value.
    \end{itemize}
\end{itemize}


\subsection{Detailed Notation Definitions for Algorithm \ref{alg:image_generation}} 
\label{sec:alg_image_notations} 

To provide a complete and formal description of the Image Generation and Filtering algorithm (Algorithm \ref{alg:image_generation}), this section details all the notations, variables, and functions used.

\begin{itemize}
    \item \textbf{Inputs:}
    \begin{itemize}
        \item $\mathcal{X}$: The original dataset containing feature-response pairs or images. This is the source data from which synthetic data is generated.
        \item $\mathcal{L}_x$: Labels, metadata, or conditions associated with the original data $\mathcal{X}$. These can be used to guide or condition the generation process. $\mathcal{L}_x$ could be a set of labels corresponding to images in $\mathcal{X}$.
        \item $\mathcal{P}$: A textual prompt or set of prompts used to guide the image generation process, particularly relevant if leveraging text-to-image diffusion models. $\mathcal{P}$ is typically a string or a collection of strings.
        \item \textbf{VAE model}: A pre-trained or trained Variational Autoencoder model used for encoding images into and potentially decoding from a latent space. This is a function $\text{VAE}(\cdot)$.
        \item \textbf{Wasserstein distance threshold}: A predefined scalar value representing the criterion used for filtering generated images based on the Wasserstein distance computed in the latent space. Images with a distance below this threshold might be selected.
    \end{itemize}
    \item \textbf{Variables:}
    \begin{itemize}
        \item $\mathcal{Y}$: The set of synthetic images generated from the original data $\mathcal{X}$ and/or guided by the prompt $\mathcal{P}$. This is a set of image data points.
        \item $\mathcal{Z}_x$: The set of latent representations for the original data $\mathcal{X}$, computed by encoding $\mathcal{X}$ using the VAE model, formally $\mathcal{Z}_x = \text{VAE}(\mathcal{X})$. $\mathcal{Z}_x$ is a set of vectors in the latent space.
        \item $\mathcal{Z}_y$: The set of latent representations for the generated images $\mathcal{Y}$, computed by encoding $\mathcal{Y}$ using the same VAE model, formally $\mathcal{Z}_y = \text{VAE}(\mathcal{Y})$. $\mathcal{Z}_y$ is also a set of vectors in the latent space.
        \item $d(\mathcal{Z}_x, \mathcal{Z}_y)$: The Wasserstein distance measuring the similarity between the distribution of latent representations of the original data ($\mathcal{Z}_x$) and the distribution of latent representations of the generated data ($\mathcal{Z}_y$). This is a scalar value.
        \item $\mathcal{Y}_{\text{filtered}}$: The subset of generated images from $\mathcal{Y}$ that satisfy the filtering criterion (e.g., having a Wasserstein distance to $\mathcal{Z}_x$ below the predefined threshold). These are the high-quality selected images, a subset of $\mathcal{Y}$.
        \item $\mathcal{X}_{\text{augmented}}$: The final augmented dataset, defined as the union of the original dataset $\mathcal{X}$ and the filtered set of synthetic images $\mathcal{Y}_{\text{filtered}}$, i.e., $\mathcal{X}_{\text{augmented}} = \mathcal{X} \cup \mathcal{Y}_{\text{filtered}}$.
    \end{itemize}
    \item \textbf{Functions:}
    \begin{itemize}
        \item $\text{GenerateImages}(\mathcal{X}, \mathcal{P}, \mathcal{L}_x)$: This function encapsulates the process of producing synthetic images $\mathcal{Y}$. It takes the original data $\mathcal{X}$, textual prompt $\mathcal{P}$, and original labels $\mathcal{L}_x$ as input to guide the generation, typically using a diffusion model. It outputs the set of generated images $\mathcal{Y}$.
        \item $\text{VAE}(\cdot)$: This function represents the encoding part of the Variational Autoencoder model, mapping an image or a set of images to their corresponding latent space representation. $\text{VAE}: \mathbb{R}^{H \times W \times C} \to \mathbb{R}^{D_L}$ (for a single image to a latent vector of dimension $D_L$).
        \item $\text{FilterImages}(\mathcal{Y}, \mathcal{Z}_x, \mathcal{Z}_y, \text{threshold})$: This function implements the selection process. It takes the generated images $\mathcal{Y}$, their latent representations $\mathcal{Z}_y$, the original latent representations $\mathcal{Z}_x$, and a filtering threshold as input. It selects a subset of images from $\mathcal{Y}$ based on a criterion related to the similarity (e.g., Wasserstein distance $d(\mathcal{Z}_x, \mathcal{Z}_y)$ compared to the threshold) and returns the filtered set $\mathcal{Y}_{\text{filtered}}$. Note: The prompt $\mathcal{P}$ was listed as an input in your rebuttal for this function, but might be used by `GenerateImages` or for validation, not strictly for filtering based on latent distance. The core filtering seems to use the latent spaces and threshold. I've kept it based on your rebuttal but note its less direct role in latent-based filtering.
    \end{itemize}
\end{itemize}

\subsection{Proof of Theorem~\ref{thm:bound}}
\label{app:proof} 

\begin{proof}[Proof Sketch]
Let $\ell$ be an $L_\ell$-Lipschitz loss function bounded by $M$, and $\mathcal{H}$ be the hypothesis class. Let $P_{\text{real}}$ be the real data distribution and $P_{\text{synth}}$ be the synthetic data distribution, assumed to satisfy $W_1(P_{\text{real}}, P_{\text{synth}}) \leq \epsilon$. Let $\hat{P}_n$ be the empirical distribution constructed from $n$ i.i.d. samples $\{z_1, \ldots, z_n\}$ drawn from $P_{\text{synth}}$. Formally, $\hat{P}_n = \frac{1}{n} \sum_{i=1}^n \delta_{z_i}$, where $\delta_{z_i}$ is the Dirac measure centered at sample $z_i$. 

The theorem establishes a bound on the expected loss under the real distribution, $\mathbb{E}_{P_{\text{real}}}\ell$, in terms of the empirical loss under the synthetic distribution, $\mathbb{E}_{\hat{P}_n}\ell$, plus terms related to the Wasserstein distance between the distributions and the complexity of the hypothesis class. The proof sketch involves bounding the gaps between the real, synthetic, and empirical synthetic distributions:

\textbf{1. Bounding the Gap between Real and Synthetic Expectations:}
Since $\ell$ is $L_\ell$-Lipschitz, by the Kantorovich-Rubinstein duality \cite{villani2009optimal}, the difference between expected losses under $P_{\text{real}}$ and $P_{\text{synth}}$ is bounded by:
\begin{align}
|\mathbb{E}_{P_{\text{real}}}\ell - \mathbb{E}_{P_{\text{synth}}}\ell| \leq L_\ell W_1(P_{\text{real}}, P_{\text{synth}}).
\end{align}
Given the assumption $W_1(P_{\text{real}}, P_{\text{synth}}) \leq \epsilon$, this implies $\mathbb{E}_{P_{\text{real}}}\ell - \mathbb{E}_{P_{\text{synth}}}\ell \leq L_\ell \epsilon$.

\textbf{2. Bounding the Gap between Synthetic and Empirical Synthetic Expectations via Uniform Convergence:}
The difference between the expected loss under $P_{\text{synth}}$ and the empirical loss under $\hat{P}_n$ can be bounded using Rademacher complexity. For a class of functions $\mathcal{F} = \{\ell(h, \cdot) : h \in \mathcal{H}\}$, assuming $\ell$ is bounded by $M$, a standard uniform convergence bound \cite{bartlett2002rademacher} states that, with probability at least $1-\delta$ over the sample $\{z_i\} \sim P_{\text{synth}}$:
\begin{align}
\mathbb{E}_{P_{\text{synth}}}\ell - \mathbb{E}_{\hat{P}_n}\ell \leq 2\mathfrak{R}_n(\mathcal{F}) + M\sqrt{\frac{\log(1/\delta)}{2n}}.
\end{align}
Note that $\mathfrak{R}_n(\mathcal{F}) = \mathfrak{R}_n(\ell \circ \mathcal{H})$.

\textbf{3. Combining Bounds to relate Real to Empirical Synthetic Expectation:}
By combining the bounds from Step 1 and Step 2, we relate the expected loss on real data to the empirical loss on synthetic data. Specifically, adding the inequality from Step 2 ($\mathbb{E}_{P_{\text{synth}}}\ell \leq \mathbb{E}_{\hat{P}_n}\ell + 2\mathfrak{R}_n(\mathcal{F}) + M\sqrt{\dots}$) to the upper bound from Step 1 ($\mathbb{E}_{P_{\text{real}}}\ell \leq \mathbb{E}_{P_{\text{synth}}}\ell + L_\ell \epsilon$), we obtain:
\begin{align}
\mathbb{E}_{P_{\text{real}}}\ell &\leq (\mathbb{E}_{\hat{P}_n}\ell + 2\mathfrak{R}_n(\mathcal{F}) + M\sqrt{\frac{\log(1/\delta)}{2n}}) + L_\ell \epsilon \\
\mathbb{E}_{P_{\text{real}}}\ell - \mathbb{E}_{\hat{P}_n}\ell &\leq L_\ell \epsilon + 2\mathfrak{R}_n(\mathcal{F}) + M\sqrt{\frac{\log(1/\delta)}{2n}}.
\end{align}
This inequality holds with probability at least $1-\delta$ over the sample $\{z_i\}$ drawn from $P_{\text{synth}}$. This completes the proof sketch, demonstrating how the bound depends on the Wasserstein distance between real and synthetic distributions ($\epsilon$) and the complexity of the hypothesis class.
\end{proof}

\subsection{High-dimensional Linear Regression}\label{High-dimensional Linear Regression}
\begin{figure}[htbp]
    \centering
    \includegraphics[width=0.25\textwidth]{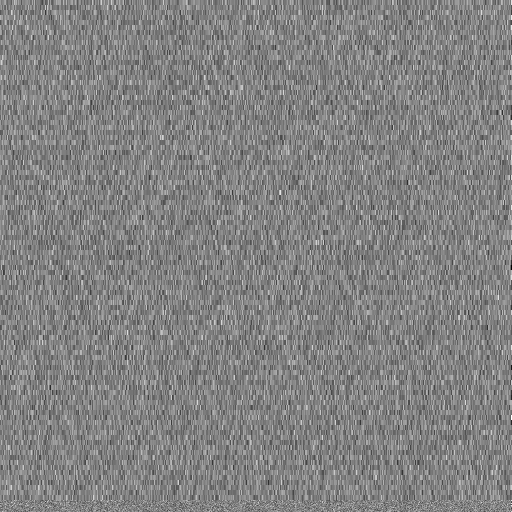}
    \caption{A grayscale representation $\mathcal{F}_1$ of $V_1$ under high dimentional settings, $\mathcal{M}_i(v)=e^{0.05v}$}
    \label{high_dimention_grayscale_representation}
\end{figure}

We similarly extract a subset \( V_1\) of size \( n = 100 \) to generate the grayscale representation \( \mathcal{F} \), as depicted in Figure \ref{high_dimention_grayscale_representation}. The remaining 100 samples are denoted as \( V_2 \). 

The generation process employs the \texttt{Stable\allowbreak\ 
 DiffusionImg2Img Pipeline} with the \textit{stable-diffusion-xl-refiner-1.0} model. We specify the following prompt to guide the image generation:

\begin{quote}
\textit{"Highly detailed grayscale noise matrix, 512$\times$512 pixels, each row represents an independent data sample. The last column is the response variable. High dimensional data distribution. Emphasizing row-wise independence, technical dataset representation with no artistic effects. Pure numerical matrix. Sharp detail. Vertical data patterns."}
\end{quote}

\subsection{High-dimensional Generalized Linear Regression}\label{highglm}
The generation process employs the \texttt{StableDiffusionImg2ImgPipeline} with the \textit{stable-diffusion-xl-refiner-1.0} model. We set the strength from 0.01 to 1 in steps of 0.01 and \texttt{guidance\_scale} to 15.0. The following prompt is specified to guide the image generation:

\begin{quote}
\textit{"Highly detailed grayscale noise matrix, 512$\times$512 pixels, each row represents an independent data sample. The last column is the response variable. High dimensional data distribution. Emphasizing row-wise independence, technical dataset representation with no artistic effects. Pure numerical matrix. Sharp detail. Vertical data patterns."}
\end{quote}

\begin{figure}[htbp]
    \centering
    \includegraphics[width=0.25\textwidth]{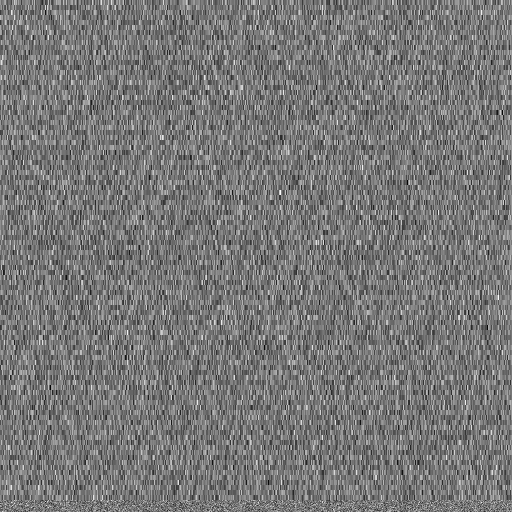}
    \caption{A grayscale representation $\mathcal{F}_1$ of $V_1$ under generalized high dimentional settings, $\mathcal{M}_i(v)=e^{0.05v}$}
    \label{generalized_high_dimention_grayscale_representation}
\end{figure}

In the image reconstruction process, we adopt a thresholding strategy for the response variable \( y \). Specifically, the pixel value in the last column of the generated grayscale matrix is compared to a threshold of 0.5. If the pixel value exceeds 0.5, we classify \( y = 1 \); otherwise, \( y = 0 \). This binary classification is consistent with the logistic regression framework, where the model predicts the probability of \( y = 1 \).

To successfully detect and obtain a precise estimate of \( \beta \), we employ the \texttt{glmtrans} method to identify the transferrable data and utilize the same method to compute the estimated parameter \( \hat{\beta} \). For each iteration, we randomly select subsets of data to construct source datasets, each containing 300 samples. This process is repeated 100 times, and the results are averaged across all iterations. 

\subsection{Boston House Price Dataset} \label{boston}
The Boston Housing Dataset is a widely studied benchmark in regression analysis and statistical modeling. Initially compiled by Harris and Tobin in 1978 at Harvard University, the dataset was designed to investigate the relationship between housing prices and various socioeconomic and environmental factors in the Boston metropolitan area.

The dataset comprises 506 observations, each representing a distinct neighborhood in Boston. It includes 13 predictor variables capturing a diverse range of attributes and a single response variable, the median value of owner-occupied homes (\texttt{MEDV}), measured in thousands of dollars.The predictor variables are detailed as follows:
\begin{itemize}
    \item \texttt{CRIM}: Per capita crime rate by town.
    \item \texttt{ZN}: Proportion of residential land zoned for lots larger than 25,000 square feet.
    \item \texttt{INDUS}: Proportion of non-retail business acres per town.
    \item \texttt{CHAS}: Charles River dummy variable (1 if tract bounds river; 0 otherwise).
    \item \texttt{NOX}: Nitric oxide concentration (parts per 10 million).
    \item \texttt{RM}: Average number of rooms per dwelling.
    \item \texttt{AGE}: Proportion of owner-occupied units built prior to 1940.
    \item \texttt{DIS}: Weighted distances to five Boston employment centers.
    \item \texttt{RAD}: Index of accessibility to radial highways.
    \item \texttt{TAX}: Full-value property-tax rate per \$10,000.
    \item \texttt{PTRATIO}: Pupil-teacher ratio by town.
    \item \texttt{B}: $1000(Bk - 0.63)^2$, where $Bk$ is the proportion of Black residents by town.
    \item \texttt{LSTAT}: Percentage of lower socioeconomic status population.
\end{itemize}
The target variable is:
\begin{itemize}
    \item \texttt{MEDV}: The median value of homes, which serves as the primary focus for regression analysis.
\end{itemize}

For our study, we adopt a low-dimensional linear model using \texttt{MEDV} as the response variable. To evaluate the model performance, the dataset is divided into three parts: 
\begin{itemize}
    \item \textbf{Training and Validation Sets (80\%)}: The data is evenly split into two sets, denoted as $V_1$ and $V_2$. These sets are treated symmetrically to explore reversible transformations and their impact on model performance.
    \item \textbf{Test Set (20\%)}: A held-out set is used exclusively for evaluating the final model performance.
\end{itemize}

\begin{figure}[htbp]
    \centering
    \includegraphics[width=0.25\textwidth]{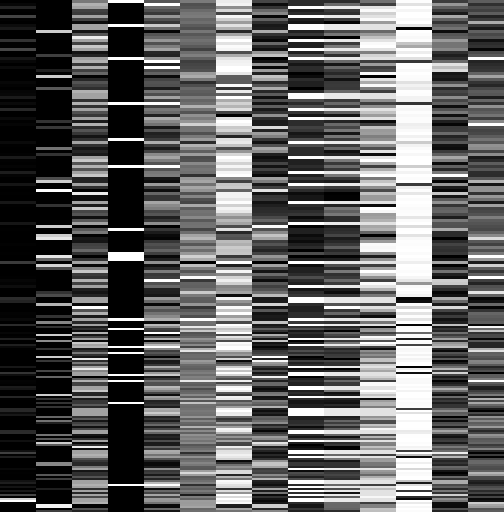}
    \caption{A grayscale representation $\mathcal{F}_1$ of $V_1$ of Boston House Price dataset, $\mathcal{M}$ is a column-wise min-max normalization}
    \label{boston_grayscale_representation}
\end{figure}

The proposed reversible mapping framework, $\mathcal{M}$, ensures consistent augmentation between $V_1$ and $V_2$, preserving the symmetry of their roles. To address the varying scales and units of the predictor variables, we apply a column-wise min-max normalization $\mathcal{M}$ prior to constructing the mapping as in \ref{boston_grayscale_representation}. This preprocessing step ensures that each variable contributes comparably to the reversible transformation, thereby enhancing the interpretability and robustness of the augmentation process. This experimental setup facilitates rigorous evaluation of the data augmentation method while adhering to statistical principles and reproducibility.

The generation process employs the \texttt{StableDiffusionImg2ImgPipeline} with the \textit{stable-diffusion-xl-refiner-1.0} model. We set the strength from 0.01 to 1 in steps of 0.01 and \texttt{guidance\_scale} to 7.5. The following prompt is specified to guide the image generation:

\begin{quote}
\textit{"Create a grayscale matrix image with 512 rows and 14 columns, designed to visually represent high-dimensional data distributions. Ensure the last column is visually distinct to highlight the response variable. The image should feature a smooth gradient from left to right, mimicking statistical patterns."}
\end{quote}

To identify transferable sources, we employ the \texttt{glmtrans} method with a parameter setting of $C_0 = 2$. For each iteration, we randomly select subsets of data to construct source datasets, each containing 100 samples. This process is repeated 100 times, and the results are averaged across all iterations. The final outcomes are presented in Figure \ref{boston_result}.

\subsection{German Credit Dataset} \label{german}
The German Credit Dataset is a widely used benchmark for credit risk assessment, commonly applied in both machine learning and statistical analysis. The dataset contains detailed information on 1,000 loan applicants, and it is primarily used to analyze the creditworthiness of individuals. By examining these data, researchers and practitioners can identify key factors influencing credit risk and assist financial institutions in making more informed lending decisions.

The dataset was originally collected by Professor Hans Hofmann at the University of Hamburg and has been made publicly available in the UCI Machine Learning Repository.  

The primary objective of the dataset is to assess the credit risk of loan applicants. Each applicant is classified as either a "good credit risk" (1) or a "bad credit risk" (0), making it suitable for binary classification tasks. By analyzing these classifications, financial institutions can develop more effective risk management strategies, potentially reducing the likelihood of defaults.

The dataset includes 20 predictor variables and a binary target variable, as detailed below:

\begin{itemize}
    \item \texttt{Status of existing checking account}: The status of the applicant's current checking account.
    \item \texttt{Duration in month}: The duration of the loan in months.
    \item \texttt{Credit history}: The applicant's credit history.
    \item \texttt{Purpose}: The purpose of the loan.
    \item \texttt{Credit amount}: The amount of credit requested.
    \item \texttt{Savings account/bonds}: The status of the applicant's savings account or bonds.
    \item \texttt{Present employment since}: The duration of the applicant's current employment.
    \item \texttt{Installment rate in percentage of disposable income}: The proportion of disposable income allocated for loan repayments.
    \item \texttt{Personal status and sex}: The applicant's personal status and gender.
    \item \texttt{Other debtors / guarantors}: Information on other debtors or guarantors.
    \item \texttt{Residence since}: The duration of the applicant's residence at the current address.
    \item \texttt{Property}: The applicant's property status.
    \item \texttt{Age in years}: The applicant's age.
    \item \texttt{Other plans}: Other financial plans or commitments.
    \item \texttt{Housing}: The applicant's housing situation.
    \item \texttt{Number of existing credits at this bank}: The number of current credits with the bank.
    \item \texttt{Job}: The applicant's job type.
    \item \texttt{Dependents}: The number of dependents the applicant has.
    \item \texttt{Telephone}: Whether the applicant has a telephone.
    \item \texttt{Foreign worker}: The applicant's status as a foreign worker.
\end{itemize}

The target variable is:

\begin{itemize}
    \item \texttt{Class}: The credit risk classification (0 = good credit risk, 1 = bad credit risk).
\end{itemize}

For our study, we adopt a high-dimensional logistic model using \texttt{Class} as the response variable. To evaluate the model performance, the dataset is divided into three parts: 
\begin{itemize}
    \item \textbf{Training and Validation Sets (80\%)}: The data is evenly split into two sets, denoted as $V_1$ and $V_2$. These sets are treated symmetrically to explore reversible transformations and their impact on model performance.
    \item \textbf{Test Set (20\%)}: A held-out set is used exclusively for evaluating the final model performance.
\end{itemize}
 
\begin{figure}[htbp]
    \centering
    \includegraphics[width=0.25\textwidth]{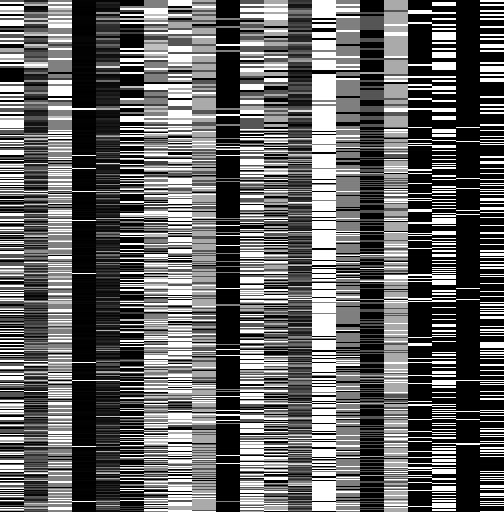}
    \caption{A grayscale representation $\mathcal{F}_1$ of $V_1$ of German Credit Dataset, $\mathcal{M}$ is a column-wise min-max normalization}
    \label{german_grayscale_representation}
\end{figure}

 To address the varying scales and units of the predictor variables, we apply a column-wise min-max normalization $\mathcal{M}$ prior to constructing the mapping as in \ref{german_grayscale_representation}. This preprocessing step ensures that each variable contributes comparably to the reversible transformation, thereby enhancing the interpretability and robustness of the augmentation process. This experimental setup facilitates rigorous evaluation of the data augmentation method while adhering to statistical principles and reproducibility.

The generation process employs the \texttt{StableDiffusionImg2ImgPipeline} with the \textit{stable-diffusion-xl-refiner-1.0} model. We set the strength from 0.01 to 1 in steps of 0.001 and \texttt{guidance\_scale} to 7.5. The following prompt is specified to guide the image generation:

\begin{quote}
\textit{"Highly detailed grayscale matrix, 512rows and 21 columns, each row represents an independent data sample. The last column is the response variable. High dimensional data distribution. Emphasizing row-wise independence, technical dataset representation with no artistic effects. Pure numerical matrix. Sharp detail. Vertical data patterns."}
\end{quote}

To identify transferable sources, we employ the \texttt{glmtrans} method with a parameter setting of $C_0 = 2$. For each iteration, we randomly select subsets of data to construct source datasets, each containing 100 samples. This process is repeated 100 times, and the results are averaged across all iterations. The final outcomes are presented in Figure \ref{german_result}.

\subsection{GTex Data} \label{gtex}
The Genotype-Tissue Expression (GTEx) dataset is a comprehensive resource widely utilized in biomedical research, particularly for studying the relationship between genetic variation and gene expression across human tissues. Initiated in 2010, the GTEx project provides a rich dataset containing gene expression measurements across 49 tissue types from 838 human donors, offering valuable insights into the genetic mechanisms underlying complex diseases.

In this study, we focus on the brain-related subset of the GTEx dataset, particularly examining genes implicated in the pathogenesis of Alzheimer’s disease (AD). Specifically, we analyze 13 brain tissues and a curated set of 119 genes, derived from the Human Molecular Signatures Database, which are downregulated in AD patients \cite{blalock2004incipient}. The target tissue in our analysis is the Hippocampus , a brain region critically associated with memory and affected early in AD. The remaining brain tissues serve as source tissues for cross-tissue analysis.

We investigate the association between the APOE gene, a major genetic risk factor for AD, as the response variable, and the remaining genes in the curated set as predictors. APOE encodes apolipoprotein E, which plays a key role in lipid transport and neuronal repair in the brain. Its allelic variants are strongly associated with an increased risk of developing AD.

For robust evaluation, we employ a repeated random sampling approach. In each experiment, the samples from the target tissue are randomly split into a training set (80\% of the data) and a validation set (20\% of the data). The training set is used to estimate the coefficients (\(\beta_0\)) of the target model, while the validation set is used to compute prediction errors. This procedure is repeated 100 times, and the average prediction errors are used to assess model performance and stability across experiments.

\begin{figure}[htbp]
    \centering
    \includegraphics[width=0.25\textwidth]{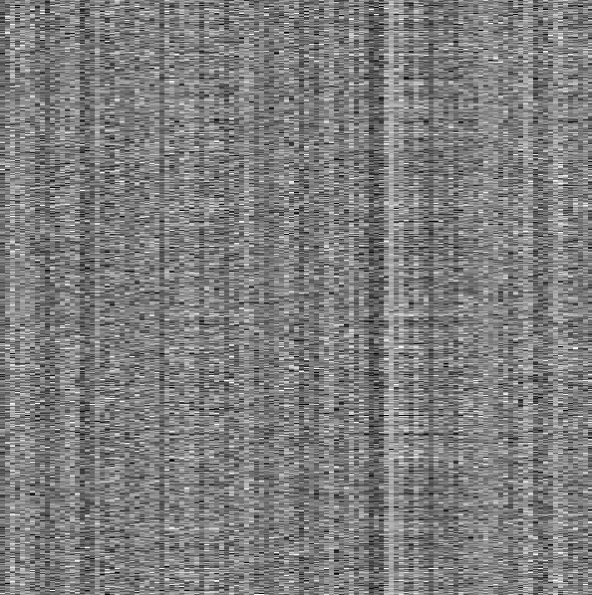}
    \caption{A grayscale representation $\mathcal{F}_1$ of $V_1$ of GTex data set, $\mathcal{M}$ is a column-wise min-max normalization}
    \label{apoe_grayscale_representation}
\end{figure}

To address the varying scales and units of the predictor variables, we apply a column-wise min-max normalization $\mathcal{M}$ prior to constructing the mapping as in \ref{apoe_grayscale_representation}. This preprocessing step ensures that each variable contributes comparably to the reversible transformation, thereby enhancing the interpretability and robustness of the augmentation process. This experimental setup facilitates rigorous evaluation of the data augmentation method while adhering to statistical principles and reproducibility.

The generation process employs the \texttt{StableDiffusionImg2ImgPipeline} with the \textit{stable-diffusion-xl-refiner-1.0} model. We set the strength from 0.01 to 1 in steps of 0.001 and \texttt{guidance\_scale} to 7.5. The following prompt is specified to guide the image generation:

\begin{quote}
\textit{"Highly detailed grayscale matrix, 598 rows and 119 columns, each row represents an independent data sample. The last column is the response variable. High dimensional data distribution. Emphasizing row-wise independence, technical dataset representation with no artistic effects. Pure numerical matrix. Sharp detail. Vertical data patterns."}
\end{quote}

To detect transferable sources, we employ the \texttt{hdtrd} method, for calculating the transferred $\hat{\beta}$, we choose glmtrans. For each iteration, we randomly select subsets of data to construct source datasets, each containing 100 samples. This process is repeated 100 times, and the results are averaged across all iterations. The final outcomes are presented in Figure \ref{apoe_result}.

\subsection{MNIST Dataset} \label{mnitst}
The MNIST dataset consists of 60,000 training and 10,000 test samples of handwritten digits. For our experiments, we randomly selected a subset of 600 samples from the training set to ensure a balanced distribution across all classes. Each grayscale image was converted to a 3-channel RGB format to match the input requirements of the Stable Diffusion model. The pixel values were normalized to the range $[-1, 1]$ using the transformation $\mathbf{x} = 2 \cdot \mathbf{x}_{\text{original}} - 1$, where $\mathbf{x}_{\text{original}} \in [0, 1]^{3 \times 28 \times 28}$ represents the original image tensor. No additional data augmentation techniques were applied to the dataset. 

Image generation was performed using the Stable Diffusion XL Refiner 1.0 model, which was conditioned on the prompt "a black and white handwritten digit." For each selected MNIST image, we generated synthetic images using 50 inference steps and a guidance scale of 7.5. The generated images were resized to $28 \times 28$ pixels to match the original MNIST resolution. To ensure quality, we used a pre-trained AutoencoderKL model to encode both the original and generated images into a latent space $\mathbf{z} \in \mathbb{R}^{4 \times 64 \times 64}$. The Wasserstein distance was computed between the latent representations of the original and generated images, and only the top 80\% of samples with the smallest distances were retained for training.

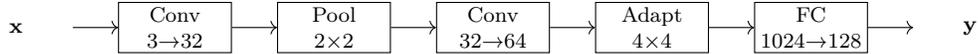
\begin{figure}[htbp]
\centering
\begin{tikzpicture}[
    node distance=0.6cm, 
    auto,
    block/.style={
        draw, 
        rectangle, 
        minimum width=1.5cm, 
        minimum height=0.6cm, 
        align=center,
        font=\scriptsize,
        inner sep=2pt 
    },
    inputnode/.style={
        minimum width=1.5cm,
        minimum height=0.6cm, 
        align=center,
        font=\scriptsize
    }
]
    \node[inputnode] (input) {$\mathbf{x}$};
    \node[block, right=of input] (conv1) {Conv \\ 3→32};
    \node[block, right=of conv1] (pool1) {Pool \\ 2×2};
    \node[block, right=of pool1] (conv2) {Conv \\ 32→64};
    \node[block, right=of conv2] (pool2) {Adapt \\ 4×4};
    \node[block, right=of pool2] (fc1) {FC \\ 1024→128};
    \node[inputnode, right=of fc1] (output) {$\mathbf{y}$};
    
    \path[->, line width=0.4pt] 
        (input) edge (conv1)
        (conv1) edge (pool1)
        (pool1) edge (conv2)
        (conv2) edge (pool2)
        (pool2) edge (fc1)
        (fc1) edge (output);
\end{tikzpicture}
\caption{Network architecture of SimpleCNN. Batch normalization and dropout layers are omitted for clarity.}
\end{figure}

The model was trained using the Adam optimizer with a fixed learning rate of $0.001$ and batch size of 64. Cross-entropy loss was used to optimize the network parameters over 10 epochs. Mixed-precision training (FP16) was employed to accelerate computation and reduce memory usage. The training process was conducted on an NVIDIA V100 GPU with 32GB of memory, requiring approximately 0.5 hours per task. The final model was evaluated on the full MNIST test set to measure classification accuracy.

\begin{figure}[htbp]
    \centering
    \includegraphics[width=0.475\textwidth]{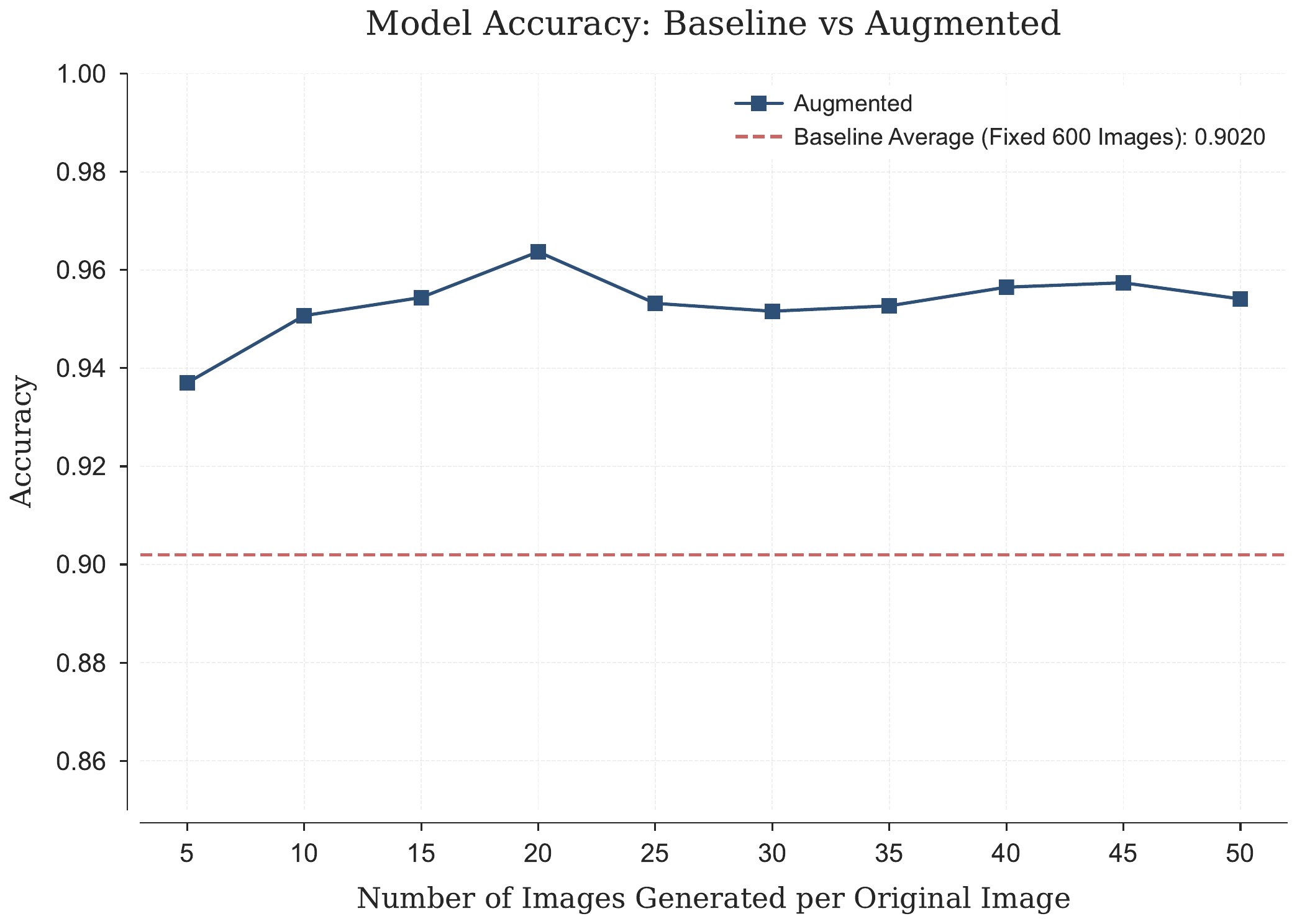}
    \caption{Comparison of test accuracy between the baseline CNN model and the augmented dataset approach.}
    \label{minist_result}
\end{figure}

\subsection{CIFAR-10 Dataset} \label{tab:cifar10}

In our experiments, we employed ResNet-20 as the baseline architecture, trained on a fixed subset of 1,000 samples (100 per class). We applied Stable Diffusion XL (\texttt{strength=0.3}) to each training image, selecting the top 60\% of generated images based on Wasserstein distance in latent space relative to the original set. These filtered images were combined with an additional 2,500 samples and used to train the same ResNet-20 model. This approach yielded an accuracy of approximately 73\%, demonstrating improved performance over the baseline. Performance metrics (\%) on CIFAR-10 with varying numbers of generated images (Gen) indicate that augmentation enhances accuracy, with Wasserstein, Maximum Mean Discrepancy, and Total Variation filtering methods exhibiting comparable efficacy.

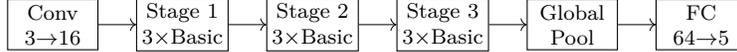
\begin{figure}[htbp]
\centering
\begin{tikzpicture}[
    node distance=0.5cm,
    block/.style={
        draw,
        rectangle,
        minimum width=1.2cm,
        minimum height=0.5cm,
        align=center,
        font=\scriptsize,
        inner sep=2pt
    },
    arr/.style={->, line width=0.3pt}
]
    \node[block] (conv1) {Conv\\3$\rightarrow$16};
    \node[block, right=of conv1] (stage1) {Stage 1\\3$\times$Basic};
    \node[block, right=of stage1] (stage2) {Stage 2\\3$\times$Basic};
    \node[block, right=of stage2] (stage3) {Stage 3\\3$\times$Basic};
    \node[block, right=of stage3] (pool) {Global\\Pool};
    \node[block, right=of pool] (fc) {FC\\64$\rightarrow$5};
    
    \draw[arr] (conv1) -- (stage1);
    \draw[arr] (stage1) -- (stage2);
    \draw[arr] (stage2) -- (stage3);
    \draw[arr] (stage3) -- (pool);
    \draw[arr] (pool) -- (fc);
\end{tikzpicture}
\caption{ResNet-20 architecture for 5-class CIFAR-10 classification. Basic blocks contain two 3$\times$3 convolutions with batch normalization and residual connections.}
\end{figure}

Training employs the Adam optimizer ($\beta_1=0.9$, $\beta_2=0.999$) with initial learning rate $\eta=0.001$, reduced by 50\% every 20 epochs. Models train for 30 epochs using mixed-precision (FP16) on NVIDIA v100 GPUs about 4 hours, with batch size 64 and cross-entropy loss. Baseline models use only original training samples, while augmented models combine original data with the top 50\% Wasserstein-filtered synthetic images.

\begin{table}[htbp]
\centering

\caption{Comparative Performance of Data Filtering Methods on CIFAR10. Data augmentation uses Wass (Wasserstein), MMD (Maximum Mean Discrepancy), and TV (Total Variation) metrics, retaining the top 60\% of images. Baseline: original samples (averaged across Gen); Augmented: unfiltered generated data; Wass, MMD, TV: filtered by respective metrics. Metrics: Acc (Accuracy), Prec (Precision), Rec (Recall), F1 (F1-Score).} 
\footnotesize
\begin{tabular}{@{}p{0.5\textwidth}@{}p{0.5\textwidth}@{}}
\begin{tabular}{clrrrr}
\toprule
\textbf{Gen} & \textbf{Model} & \textbf{Acc} & \textbf{Prec} & \textbf{Rec} & \textbf{F1} \\
\midrule
\multirow{4}{*}{\centering\textbf{2}} & Baseline & 38.55 & 38.48 & 38.55 & 38.41 \\
& Wass & \textbf{42.34} & \textbf{42.47} & \textbf{42.34} & \textbf{42.07} \\
& MMD & 42.14 & 41.71 & 42.14 & 41.63 \\
& TV & 39.93 & 40.20 & 39.93 & 39.85 \\
\midrule
\multirow{4}{*}{\centering\textbf{4}} & Baseline & 39.21 & 39.31 & 39.21 & 38.95 \\
& Wass & 42.37 & 42.44 & 42.37 & \textbf{42.24} \\
& MMD & \textbf{42.90} & \textbf{42.86} & \textbf{42.90} & 42.67 \\
& TV & 42.27 & 42.26 & 42.27 & 42.09 \\
\midrule
\multirow{4}{*}{\centering\textbf{6}} & Baseline & 38.86 & 38.96 & 38.86 & 38.76 \\
& Wass & 41.97 & 42.12 & 41.97 & 41.79 \\
& MMD & \textbf{44.65} & \textbf{44.52} & \textbf{44.65} & \textbf{44.32} \\
& TV & 42.75 & 43.29 & 42.75 & 42.73 \\
\midrule
\multirow{4}{*}{\centering\textbf{8}} & Baseline & 39.12 & 38.87 & 39.12 & 38.59 \\
& Wass & 43.82 & 43.63 & 43.82 & 43.42 \\
& MMD & \textbf{44.42} & \textbf{44.24} & \textbf{44.42} & \textbf{44.22} \\
& TV & 42.84 & 42.66 & 42.84 & 42.35 \\
\midrule
\multirow{4}{*}{\centering\textbf{10}} & Baseline & 38.14 & 38.00 & 38.14 & 37.90 \\
& Wass & 43.93 & 43.79 & 43.93 & 43.63 \\
& MMD & 43.79 & \textbf{44.00} & 43.79 & \textbf{43.67} \\
& TV & 42.89 & 42.53 & \textbf{42.89} & 42.53 \\
\bottomrule
\end{tabular}
&
\begin{tabular}{clrrrr}
\toprule
\textbf{Gen} & \textbf{Model} & \textbf{Acc} & \textbf{Prec} & \textbf{Rec} & \textbf{F1} \\
\midrule
\multirow{4}{*}{\centering\textbf{12}} & Baseline & 38.18 & 38.44 & 38.18 & 38.02 \\
& Wass & \textbf{43.88} & \textbf{43.67} & \textbf{43.88} & \textbf{43.55} \\
& MMD & 43.43 & 42.95 & 43.43 & 43.02 \\
& TV & 42.70 & 42.24 & 42.70 & 42.27 \\
\midrule
\multirow{4}{*}{\centering\textbf{14}} & Baseline & 37.85 & 37.72 & 37.85 & 37.50 \\
& Wass & 42.28 & 42.34 & 42.28 & 42.01 \\
& MMD & 42.74 & 42.31 & 42.74 & 42.33 \\
& TV & \textbf{44.10} & \textbf{43.69} & \textbf{44.10} & \textbf{43.76} \\
\midrule
\multirow{4}{*}{\centering\textbf{16}} & Baseline & 39.65 & 40.11 & 39.65 & 39.45 \\
& Wass & \textbf{44.51} & \textbf{44.28} & \textbf{44.51} & \textbf{44.27} \\
& MMD & 43.80 & 43.27 & 43.80 & 43.29 \\
& TV & 43.74 & 43.16 & 43.74 & 43.27 \\
\midrule
\multirow{4}{*}{\centering\textbf{18}} & Baseline & 38.85 & 39.03 & 38.85 & 38.85 \\
& Wass & 43.79 & 43.53 & 43.79 & \textbf{43.56} \\
& MMD & 43.06 & 42.99 & 43.06 & 42.79 \\
& TV & \textbf{44.22} & \textbf{43.93} & \textbf{44.22} & 43.74 \\
\midrule
\multirow{4}{*}{\centering\textbf{20}} & Baseline & 38.92 & 38.93 & 38.92 & 38.67 \\
& Wass & 43.52 & 43.15 & 43.52 & 43.15 \\
& MMD & \textbf{43.97} & \textbf{43.45} & \textbf{43.97} & \textbf{43.27} \\
& TV & 42.59 & 42.41 & 42.59 & 42.30 \\
\bottomrule
\end{tabular}
\end{tabular}
\end{table}

\subsection{CIFAR-100 Dataset}\label{cifar100}

Performance metrics, reported as percentages, were evaluated on a 20-class subset of the CIFAR-100 dataset across varying training set sizes. For each original image, ten augmented images were generated using Stable Diffusion XL, with five images at a strength of 0.15 and five at 0.8. The models employed a pretrained ResNet-18 architecture with ImageNet weights, where the \texttt{conv1} and \texttt{layer1} to \texttt{layer3} modules were frozen, and only \texttt{layer4} and the classifier were fine-tuned. Training utilized the Adam optimizer with a learning rate of $10^{-4}$ for the baseline and $5 \times 10^{-5}$ for augmented models, a batch size of 32, and a dropout rate of 0.5. Data filtering methods (Wasserstein, Total Variation, and Maximum Mean Discrepancy) yielded performance comparable to unfiltered augmentation, with negligible differences. This similarity is attributed to CIFAR-100's strong representation in Stable Diffusion's pretraining, which minimizes generation anomalies. Nonetheless, for fine-grained classification tasks, we recommend applying filtering to enhance model robustness.

\begin{table}[htbp] 
\centering
\caption{Evaluation of Filtered Data Augmentation on CIFAR-100 (20 Classes). Baseline: original samples; None: mean of Wasserstein, TV, MMD at 100\% tolerance; Wass, TV, MMD: filtered data at 40\%, 60\%, 80\% tolerance.} 
\footnotesize
\begin{tabular}{@{}p{0.5\textwidth}@{}p{0.5\textwidth}@{}}
\begin{tabular}{clrrrr}
\toprule
\textbf{Size} & \textbf{Model} & \textbf{Acc} & \textbf{Prec} & \textbf{Rec} & \textbf{F1} \\
\midrule
\multirow{2}{*}{\centering\textbf{500}} 
  & Baseline & 0.692 & 0.695 & 0.692 & 0.692 \\
  & None & \textbf{0.840} & 0.840 & 0.837 & 0.837 \\
\cmidrule(lr){2-6}
  & Wass-40 & \textbf{0.813} & 0.814 & 0.810 & 0.810 \\
  & Wass-60 & \textbf{0.825} & 0.824 & 0.821 & 0.821 \\
  & Wass-80 & \textbf{0.840} & 0.837 & 0.834 & 0.834 \\
\midrule[0.5pt]
  & TV-40 & \textbf{0.818} & 0.813 & 0.809 & 0.808 \\
  & TV-60 & \textbf{0.820} & 0.817 & 0.815 & 0.814 \\
  & TV-80 & \textbf{0.834} & 0.829 & 0.821 & 0.821 \\
\midrule[0.5pt]
  & MMD-40 & \textbf{0.815} & 0.812 & 0.808 & 0.807 \\
  & MMD-60 & \textbf{0.824} & 0.820 & 0.816 & 0.816 \\
  & MMD-80 & \textbf{0.832} & 0.829 & 0.825 & 0.824 \\
\midrule
\multirow{2}{*}{\centering\textbf{1000}} 
  & Baseline & 0.763 & 0.766 & 0.763 & 0.762 \\
  & None & \textbf{0.874} & 0.877 & 0.872 & 0.872 \\
\cmidrule(lr){2-6}
  & Wass-40 & \textbf{0.851} & 0.851 & 0.848 & 0.847 \\
  & Wass-60 & \textbf{0.863} & 0.865 & 0.860 & 0.861 \\
  & Wass-80 & \textbf{0.866} & 0.855 & 0.851 & 0.851 \\
\midrule[0.5pt]
  & TV-40 & \textbf{0.848} & 0.843 & 0.836 & 0.836 \\
  & TV-60 & \textbf{0.860} & 0.858 & 0.856 & 0.855 \\
  & TV-80 & \textbf{0.872} & 0.873 & 0.870 & 0.869 \\
\midrule[0.5pt]
  & MMD-40 & \textbf{0.856} & 0.8542 & 0.8391 & 0.837823 \\
  & MMD-60 & \textbf{0.863} & 0.8609 & 0.860 & 0.855 \\
  & MMD-80 & \textbf{0.865} & 0.855 & 0.871 & 0.872 \\
\bottomrule
\end{tabular}
&
\begin{tabular}{clrrrr}
\toprule
\textbf{Size} & \textbf{Model} & \textbf{Acc} & \textbf{Prec} & \textbf{Rec} & \textbf{F1} \\
\midrule
\multirow{2}{*}{\centering\textbf{1500}} 
  & Baseline & 0.792 & 0.789 & 0.789 & 0.788 \\
  & None & \textbf{0.882} & 0.877 & 0.874 & 0.874 \\
\cmidrule(lr){2-6}
  & Wass-40 & \textbf{0.867} & 0.866 & 0.864 & 0.864 \\
  & Wass-60 & \textbf{0.872} & 0.868 & 0.866 & 0.865 \\
  & Wass-80 & \textbf{0.875} & 0.872 & 0.870 & 0.869 \\
\midrule[0.5pt]
  & TV-40 & \textbf{0.861} & 0.858 & 0.856 & 0.854 \\
  & TV-60 & \textbf{0.872} & 0.869 & 0.867 & 0.866 \\
  & TV-80 & \textbf{0.873} & 0.875 & 0.873 & 0.873 \\
\midrule[0.5pt]
  & MMD-40 & \textbf{0.867} & 0.866 & 0.864 & 0.864 \\
  & MMD-60 & \textbf{0.872} & 0.868 & 0.866 & 0.865 \\
  & MMD-80 & \textbf{0.875} & 0.872 & 0.870 & 0.869 \\
\midrule
\multirow{2}{*}{\centering\textbf{2000}} 
  & Baseline & 0.815 & 0.815 & 0.813 & 0.812 \\
  & None & \textbf{0.888} & 0.886 & 0.884 & 0.884 \\
\cmidrule(lr){2-6}
  & Wass-40 & \textbf{0.874} & 0.872 & 0.870 & 0.870 \\
  & Wass-60 & \textbf{0.876} & 0.877 & 0.876 & 0.876 \\
  & Wass-80 & \textbf{0.883} & 0.882 & 0.880 & 0.880 \\
\midrule[0.5pt]
  & TV-40 & \textbf{0.869} & 0.869 & 0.866 & 0.865 \\
  & TV-60 & \textbf{0.884} & 0.876 & 0.874 & 0.874 \\
  & TV-80 & \textbf{0.884} & 0.877 & 0.875 & 0.874 \\
\midrule[0.5pt]
  & MMD-40 & \textbf{0.873} & 0.871 & 0.870 & 0.869 \\
  & MMD-60 & \textbf{0.878} & 0.876 & 0.875 & 0.874 \\
  & MMD-80 & \textbf{0.882} & 0.880 & 0.879 & 0.879 \\
\bottomrule
\end{tabular}
\end{tabular}

\end{table}

\subsection{ISIC Dataset} \label{ISIC}
The ISIC 2018 dataset comprising 7,015 dermoscopic images across seven diagnostic categories was utilized, with balanced subsets created through stratified sampling (1,000 training and 200 test images). Each 450$\times$450 RGB image underwent channel-wise normalization using ImageNet statistics prior to processing. For diffusion-based augmentation, we employed Stable Diffusion XL Refiner 1.0, generating five variants per original image through 30-step inference at guidance scale 7.5. The VAE latent space ($\mathbb{R}^{4\times128\times128}$) embeddings were computed for both original and generated images, with Wasserstein distance thresholds determined per-class at the 50th percentile of pairwise distances.

The ResNet-20 architecture was implemented with batch normalization and dropout (0.5) after global average pooling. Training proceeded for 50 epochs using Adam optimizer (initial learning rate 5$\times$10$^{-5}$, $\beta_1=0.9$, $\beta_2=0.999$) with cosine learning rate decay. Mini-batches of 32 samples combined 70\% original and 30\% augmented images, applying random horizontal/vertical flips for regularization. Evaluation metrics were computed over three independent runs, reporting mean macro-F1 scores with 95\% confidence intervals derived through bootstrap resampling (1,000 iterations). All experiments utilized mixed-precision training on NVIDIA A100 GPUs, completing in under 2.5 hours per configuration.

Performance metrics (\%) on a 7-class skin cancer image dataset (ISIC 2018) with 1,257 original training samples and varying numbers of generated images (Gen) from SD-XL, mixed at \texttt{strength=0.15} for fidelity and \texttt{strength=0.85} for diverisity. Higher strength increases diversity but introduces suboptimal samples, requiring Wass filtering. Wass consistently enhances performance over the baseline.

\begin{table}[htbp]
\centering
\caption{Evaluation of Wasserstein-Filtered Data Augmentation on ISIC 2018 Dataset. Baseline: original samples (averaged across tasks); None: unfiltered generated data; Wass: Wasserstein-filtered data, retaining the top 60\% of images.}
\footnotesize
\begin{tabular}{@{}p{0.5\textwidth}@{}p{0.5\textwidth}@{}}
\begin{tabular}{clrrrr}
\toprule
\textbf{Gen} & \textbf{Model} & \textbf{Acc} & \textbf{Prec} & \textbf{Rec} & \textbf{F1} \\
\midrule
\multirow{2}{*}{\centering\textbf{3}} & None & 60.00 & 58.73 & 60.00 & 57.79 \\
                                      & Wass      & \textbf{62.86} & \textbf{63.48} & \textbf{62.86} & \textbf{61.07} \\
\midrule[0.5pt]
\multirow{2}{*}{\centering\textbf{6}} & None & 45.71 & 45.69 & 45.71 & 44.48 \\
                                      & Wass      & \textbf{57.14} & \textbf{63.81} & \textbf{57.14} & \textbf{56.76} \\
\midrule[0.5pt]
\multirow{2}{*}{\centering\textbf{9}} & None & 50.00 & 52.90 & 50.00 & 50.10 \\
                                      & Wass      & \textbf{55.71} & \textbf{59.18} & \textbf{55.71} & \textbf{53.73} \\
\midrule[0.5pt]
\multirow{2}{*}{\centering\textbf{12}} & None & 52.86 & 54.32 & 52.86 & 51.94 \\
                                       & Wass      & \textbf{57.14} & \textbf{60.25} & \textbf{57.14} & \textbf{56.28} \\
\bottomrule
\end{tabular}
&
\begin{tabular}{clrrrr}
\toprule
\textbf{Gen} & \textbf{Model} & \textbf{Acc} & \textbf{Prec} & \textbf{Rec} & \textbf{F1} \\
\midrule
\multirow{2}{*}{\centering\textbf{15}} & None & 51.43 & 53.88 & 51.43 & 50.67 \\
                                       & Wass      & \textbf{56.29} & \textbf{58.94} & \textbf{56.29} & \textbf{55.82} \\
\midrule[0.5pt]
\multirow{2}{*}{\centering\textbf{18}} & None & 48.57 & 61.77 & 48.57 & 47.52 \\
                                       & Wass      & \textbf{58.57} & \textbf{57.51} & \textbf{58.57} & \textbf{57.38} \\
\midrule[0.5pt]
\multirow{2}{*}{\centering\textbf{21}} & None & 47.14 & 51.16 & 47.14 & 47.73 \\
                                       & Wass      & \textbf{64.29} & \textbf{65.19} & \textbf{64.29} & \textbf{63.63} \\
\midrule[0.5pt]
\multirow{2}{*}{\centering\textbf{24}} & None & 55.71 & 52.91 & 55.71 & 51.67 \\
                                       & Wass      & \textbf{64.29} & \textbf{65.37} & \textbf{64.29} & \textbf{63.91} \\
\bottomrule
\end{tabular}
\end{tabular}

\vspace{0.5em}
\begin{tabular}{@{}l@{}}
\toprule
\textbf{Baseline (Avg.):} \quad 
\textbf{Acc}: 52.32 \quad 
\textbf{Prec}: 56.64 \quad 
\textbf{Rec}: 52.32 \quad 
\textbf{F1}: 51.88 \\
\bottomrule
\end{tabular}
\end{table}

\subsection{Cassava Lead Disease Dateset}
Performance metrics (\%) on a 5-class cassava leaf disease dataset with varying training sizes (Size). For each original image, 10 images are generated using Stable Diffusion XL, with \texttt{strength=0.2} (5 images) and \texttt{strength=0.6} (5 images). Models use pretrained EfficientNet-B0 (ImageNet weights), with feature extraction layers frozen and the classifier trained using Adam optimizer (learning rate 1e-4, batch size 32, dropout 0.5). Results show that filtering is effective, with small differences among metrics.

\label{Cassava}
\begin{table}[htbp]
\centering
\caption{Evaluation of Filtered Data Augmentation on Cassava Leaf Disease Dataset. Baseline: original samples,from 125 to 500; None: unfiltered augmentation (mean of 100\% tolerance); Wass, TV, MMD: filtered data at 20\%, 60\%, 80\% tolerance.} 
\footnotesize
\begin{tabular}{@{}p{0.5\textwidth}@{}p{0.5\textwidth}@{}}
\begin{tabular}{clrrrr}
\toprule
\textbf{Size} & \textbf{Model} & \textbf{Acc} & \textbf{Prec} & \textbf{Rec} & \textbf{F1} \\
\midrule
\multirow{2}{*}{\centering\textbf{125}} 
  & Baseline & 0.266 & 0.271 & 0.266 & 0.246 \\
  & None & \textbf{0.316} & 0.359 & 0.316 & 0.290 \\
\cmidrule(lr){2-6}
  & Wass-20 & \textbf{0.318} & 0.353 & 0.318 & 0.273 \\
  & Wass-60 & \textbf{0.352} & 0.391 & 0.352 & 0.331 \\
  & Wass-80 & \textbf{0.364} & 0.397 & 0.364 & 0.353 \\
\midrule[0.5pt]
  & TV-20 & 0.288 & 0.312 & 0.288 & 0.249 \\
  & TV-60 & \textbf{0.332} & 0.365 & 0.332 & 0.319 \\
  & TV-80 & \textbf{0.362} & 0.404 & 0.362 & 0.348 \\
\midrule[0.5pt]
  & MMD-20 & \textbf{0.334} & 0.392 & 0.334 & 0.316 \\
  & MMD-60 & \textbf{0.338} & 0.355 & 0.338 & 0.327 \\
  & MMD-80 & \textbf{0.372} & 0.392 & 0.372 & 0.362 \\
\midrule
\multirow{2}{*}{\centering\textbf{250}} 
  & Baseline & 0.350 & 0.351 & 0.350 & 0.337 \\
  & None & \textbf{0.387} & 0.398 & 0.387 & 0.367 \\
\cmidrule(lr){2-6}
  & Wass-20 & \textbf{0.396} & 0.392 & 0.396 & 0.384 \\
  & Wass-60 & \textbf{0.384} & 0.396 & 0.384 & 0.364 \\
  & Wass-80 & \textbf{0.388} & 0.396 & 0.388 & 0.381 \\
\midrule[0.5pt]
  & TV-20 & \textbf{0.418} & 0.422 & 0.418 & 0.411 \\
  & TV-60 & \textbf{0.396} & 0.406 & 0.396 & 0.381 \\
  & TV-80 & \textbf{0.442} & 0.442 & 0.442 & 0.433 \\
\midrule[0.5pt]
  & MMD-20 & \textbf{0.422} & 0.439 & 0.422 & 0.406 \\
  & MMD-60 & \textbf{0.422} & 0.432 & 0.422 & 0.415 \\
  & MMD-80 & \textbf{0.418} & 0.441 & 0.418 & 0.398 \\
\bottomrule
\end{tabular}
&
\begin{tabular}{clrrrr}
\toprule
\textbf{Size} & \textbf{Model} & \textbf{Acc} & \textbf{Prec} & \textbf{Rec} & \textbf{F1} \\
\midrule
\multirow{2}{*}{\centering\textbf{375}} 
  & Baseline & 0.398 & 0.399 & 0.398 & 0.390 \\
  & None & \textbf{0.430} & 0.420 & 0.430 & 0.413 \\
\cmidrule(lr){2-6}
  & Wass-20 & \textbf{0.434} & 0.429 & 0.434 & 0.424 \\
  & Wass-60 & \textbf{0.432} & 0.443 & 0.432 & 0.419 \\
  & Wass-80 & \textbf{0.430} & 0.415 & 0.430 & 0.402 \\
\midrule[0.5pt]
  & TV-20 & \textbf{0.472} & 0.464 & 0.472 & 0.460 \\
  & TV-60 & \textbf{0.426} & 0.419 & 0.426 & 0.414 \\
  & TV-80 & \textbf{0.458} & 0.451 & 0.458 & 0.445 \\
\midrule[0.5pt]
  & MMD-20 & \textbf{0.434} & 0.423 & 0.434 & 0.415 \\
  & MMD-60 & \textbf{0.420} & 0.413 & 0.420 & 0.399 \\
  & MMD-80 & \textbf{0.436} & 0.428 & 0.436 & 0.420 \\
\midrule
\multirow{2}{*}{\centering\textbf{500}} 
  & Baseline & 0.414 & 0.415 & 0.414 & 0.411 \\
  & None & \textbf{0.465} & 0.465 & 0.465 & 0.460 \\
\cmidrule(lr){2-6}
  & Wass-20 & \textbf{0.476} & 0.477 & 0.476 & 0.472 \\
  & Wass-60 & 0.452 & 0.461 & 0.452 & 0.436 \\
  & Wass-80 & 0.440 & 0.439 & 0.440 & 0.428 \\
\midrule[0.5pt]
  & TV-20 & \textbf{0.476} & 0.480 & 0.476 & 0.474 \\
  & TV-60 & 0.462 & 0.460 & 0.462 & 0.459 \\
  & TV-80 & \textbf{0.466} & 0.464 & 0.466 & 0.462 \\
\midrule[0.5pt]
  & MMD-20 & 0.452 & 0.452 & 0.452 & 0.446 \\
  & MMD-60 & \textbf{0.474} & 0.475 & 0.474 & 0.465 \\
  & MMD-80 & \textbf{0.474} & 0.471 & 0.474 & 0.468 \\
\bottomrule
\end{tabular}
\end{tabular}

\end{table}


\begin{thebibliography}{175}
\providecommand{\natexlab}[1]{#1}
\providecommand{\url}[1]{\texttt{#1}}
\expandafter\ifx\csname urlstyle\endcsname\relax
  \providecommand{\doi}[1]{doi: #1}\else
  \providecommand{\doi}{doi: \begingroup \urlstyle{rm}\Url}\fi

\bibitem[Alaa et~al.(2022)Alaa, Van~Breugel, Saveliev, and van~der Schaar]{alaa2022faithful}
Ahmed Alaa, Boris Van~Breugel, Evgeny~S Saveliev, and Mihaela van~der Schaar.
\newblock How faithful is your synthetic data? sample-level metrics for evaluating and auditing generative models.
\newblock In \emph{International Conference on Machine Learning}, pages 290--306. PMLR, 2022.

\bibitem[Albahri et~al.(2023)Albahri, Duhaim, Fadhel, Alnoor, Baqer, Alzubaidi, Albahri, Alamoodi, Bai, Salhi, et~al.]{albahri2023systematic}
Ahmed~Shihab Albahri, Ali~M Duhaim, Mohammed~A Fadhel, Alhamzah Alnoor, Noor~S Baqer, Laith Alzubaidi, Osamah~Shihab Albahri, Abdullah~Hussein Alamoodi, Jinshuai Bai, Asma Salhi, et~al.
\newblock A systematic review of trustworthy and explainable artificial intelligence in healthcare: Assessment of quality, bias risk, and data fusion.
\newblock \emph{Information Fusion}, 96:\penalty0 156--191, 2023.

\bibitem[Ambrosio et~al.(2008)Ambrosio, Gigli, and Savar{\'e}]{ambrosio2008gradient}
Luigi Ambrosio, Nicola Gigli, and Giuseppe Savar{\'e}.
\newblock \emph{Gradient Flows: in Metric Spaces and in the Space of Probability Measures}.
\newblock Springer Science \& Business Media, 2008.

\bibitem[Andrieu et~al.(2003)Andrieu, de~Freitas, Doucet, and Jordan]{andrieu2003introduction}
Christophe Andrieu, Nando de~Freitas, Arnaud Doucet, and Michael~I. Jordan.
\newblock An introduction to {MCMC} for machine learning.
\newblock \emph{Machine Learning}, 50\penalty0 (1):\penalty0 5--43, 2003.

\bibitem[Anthony and Bartlett(2009)]{anthony2009}
Martin Anthony and Peter~L. Bartlett.
\newblock \emph{Neural Network Learning: Theoretical Foundations}.
\newblock Cambridge University Press, USA, 1st edition, 2009.
\newblock ISBN 052111862X.

\bibitem[Arjovsky and Bottou(2017)]{Arjovsky2017Towards}
Martin Arjovsky and Léon Bottou.
\newblock Towards principled methods for training generative adversarial networks.
\newblock In \emph{International Conference on Learning Representations}, 2017.

\bibitem[Arjovsky et~al.(2017)Arjovsky, Chintala, and Bottou]{arjovsky2017wasserstein}
Martin Arjovsky, Soumith Chintala, and L{\'e}on Bottou.
\newblock Wasserstein generative adversarial networks.
\newblock In \emph{International conference on machine learning}, pages 214--223. PMLR, 2017.

\bibitem[Arnold(2012)]{arnold2012geometrical}
Vladimir~Igorevich Arnold.
\newblock \emph{Geometrical methods in the theory of ordinary differential equations}, volume 250.
\newblock Springer Science \& Business Media, 2012.

\bibitem[Assefa et~al.(2020)Assefa, Dervovic, Mahfouz, Tillman, Reddy, and Veloso]{assefa2020generating}
Samuel~A Assefa, Danial Dervovic, Mahmoud Mahfouz, Robert~E Tillman, Prashant Reddy, and Manuela Veloso.
\newblock Generating synthetic data in finance: opportunities, challenges and pitfalls.
\newblock In \emph{Proceedings of the First ACM International Conference on AI in Finance}, pages 1--8, 2020.

\bibitem[Baktashmotlagh et~al.(2013)Baktashmotlagh, Harandi, Lovell, and Salzmann]{baktashmotlagh2013unsupervised}
Mahsa Baktashmotlagh, Mehrtash~T Harandi, Brian~C Lovell, and Mathieu Salzmann.
\newblock Unsupervised domain adaptation by domain invariant projection.
\newblock In \emph{Proceedings of the IEEE international conference on computer vision}, pages 769--776, 2013.

\bibitem[Bao et~al.(2017)Bao, Chen, Wen, Li, and Hua]{Bao_2017_ICCV}
Jianmin Bao, Dong Chen, Fang Wen, Houqiang Li, and Gang Hua.
\newblock Cvae-gan: Fine-grained image generation through asymmetric training.
\newblock In \emph{Proceedings of the IEEE International Conference on Computer Vision (ICCV)}, Oct 2017.

\bibitem[Bartlett and Mendelson(2002)]{bartlett2002rademacher}
Peter~L Bartlett and Shahar Mendelson.
\newblock Rademacher and gaussian complexities: Risk bounds and structural results.
\newblock \emph{Journal of Machine Learning Research}, 3\penalty0 (Nov):\penalty0 463--482, 2002.

\bibitem[Bartlett et~al.(2019)Bartlett, Harvey, Liaw, and Mehrabian]{bartlett2019}
Peter~L. Bartlett, Nick Harvey, Christopher Liaw, and Abbas Mehrabian.
\newblock Nearly-tight vc-dimension and pseudodimension bounds for piecewise linear neural networks.
\newblock \emph{Journal of Machine Learning Research}, 20\penalty0 (63):\penalty0 1--17, 2019.
\newblock URL \url{http://jmlr.org/papers/v20/17-612.html}.

\bibitem[Beal(2003)]{beal2003variational}
Matthew~J Beal.
\newblock \emph{Variational Algorithms for Approximate Bayesian Inference}.
\newblock PhD thesis, University College London, 2003.

\bibitem[Becker and Schmidt-Thieme(2018)]{becker2018generating}
Jens Becker and Lars Schmidt-Thieme.
\newblock Generating synthetic data for machine learning.
\newblock \emph{Data Mining and Knowledge Discovery}, 32\penalty0 (5):\penalty0 1350--1376, 2018.

\bibitem[Ben-David et~al.(2010)Ben-David, Blitzer, Crammer, Kulesza, Pereira, and Vaughan]{ben2010theory}
Shai Ben-David, John Blitzer, Koby Crammer, Alex Kulesza, Fernando Pereira, and Jennifer~Wortman Vaughan.
\newblock A theory of learning from different domains.
\newblock \emph{Machine learning}, 79:\penalty0 151--175, 2010.

\bibitem[Betzalel et~al.(2024)Betzalel, Avidan, Kass, Meiri, and Neufeld]{Betzalel2024EvaluationMetrics}
Eyal Betzalel, Dvir Avidan, Adam Kass, Avihai Meiri, and Roy J.~J. Neufeld.
\newblock Evaluation metrics for generative models: An empirical study.
\newblock \emph{Machine Learning and Knowledge Extraction}, 6\penalty0 (3):\penalty0 1531--1544, 2024.

\bibitem[Bhattacharya et~al.(2023)Bhattacharya, Fan, and Mukherjee]{bhattacharya2023deep}
Sohom Bhattacharya, Jianqing Fan, and Debarghya Mukherjee.
\newblock Deep neural networks for nonparametric interaction models with diverging dimension.
\newblock \emph{arXiv e-prints}, pages arXiv--2302, 2023.

\bibitem[Bickel and Doksum(2015)]{bickel2015mathematical}
Peter~J Bickel and Kjell~A Doksum.
\newblock \emph{Mathematical statistics: basic ideas and selected topics, volumes I-II package}.
\newblock Chapman and Hall/CRC, 2015.

\bibitem[Bierkens et~al.(2019)Bierkens, Fearnhead, Roberts, et~al.]{bierkens2019zig}
Joris Bierkens, Paul Fearnhead, Gareth Roberts, et~al.
\newblock The zig-zag process and super-efficient sampling for bayesian analysis of big data.
\newblock \emph{The Annals of Statistics}, 47\penalty0 (3):\penalty0 1288--1320, 2019.

\bibitem[Billingsley(2017)]{billingsley2017probability}
Patrick Billingsley.
\newblock \emph{Probability and measure}.
\newblock John Wiley \& Sons, 2017.

\bibitem[Blalock et~al.(2004)Blalock, Geddes, Chen, Porter, Markesbery, and Landfield]{blalock2004incipient}
Eric~M Blalock, James~W Geddes, Kuey~Chu Chen, Nada~M Porter, William~R Markesbery, and Philip~W Landfield.
\newblock Incipient alzheimer's disease: microarray correlation analyses reveal major transcriptional and tumor suppressor responses.
\newblock \emph{Proceedings of the National Academy of Sciences}, 101\penalty0 (7):\penalty0 2173--2178, 2004.

\bibitem[Blei et~al.(2017)Blei, Kucukelbir, and McAuliffe]{blei2017variational}
David~M. Blei, Alp Kucukelbir, and Jon~D. McAuliffe.
\newblock Variational inference: A review for statisticians.
\newblock \emph{Journal of the American Statistical Association}, 112\penalty0 (518):\penalty0 859--877, 2017.

\bibitem[Bodenham and Kawahara(2023)]{bodenham2023eummd}
Dean~A Bodenham and Yoshinobu Kawahara.
\newblock eummd: efficiently computing the mmd two-sample test statistic for univariate data.
\newblock \emph{Statistics and Computing}, 33\penalty0 (5):\penalty0 110, 2023.

\bibitem[Bol{\'o}n-Canedo et~al.(2013)Bol{\'o}n-Canedo, S{\'a}nchez-Maro{\~n}o, and Alonso-Betanzos]{bolon2013review}
Ver{\'o}nica Bol{\'o}n-Canedo, Noelia S{\'a}nchez-Maro{\~n}o, and Amparo Alonso-Betanzos.
\newblock A review of feature selection methods on synthetic data.
\newblock \emph{Knowledge and information systems}, 34:\penalty0 483--519, 2013.

\bibitem[Bouchard-Cote et~al.(2018)Bouchard-Cote, Vollmer, and Doucet]{bouchard2018bouncy}
Alexandre Bouchard-Cote, Sebastian~J Vollmer, and Arnaud Doucet.
\newblock The bouncy particle sampler: A nonreversible rejection-free markov chain monte carlo method.
\newblock \emph{Journal of the American Statistical Association}, 113\penalty0 (522):\penalty0 855--867, 2018.

\bibitem[Brinker et~al.(2018)Brinker, Hekler, Utikal, Grabe, Schadendorf, Klode, Berking, Steeb, Enk, and Von~Kalle]{brinker2018skin}
Titus~Josef Brinker, Achim Hekler, Jochen~Sven Utikal, Niels Grabe, Dirk Schadendorf, Joachim Klode, Carola Berking, Theresa Steeb, Alexander~H Enk, and Christof Von~Kalle.
\newblock Skin cancer classification using convolutional neural networks: systematic review.
\newblock \emph{Journal of medical Internet research}, 20\penalty0 (10):\penalty0 e11936, 2018.

\bibitem[Brooks et~al.(2011)Brooks, Gelman, Jones, and Meng]{brooks2011handbook}
Steve Brooks, Andrew Gelman, Galin~L. Jones, and Xiao-Li Meng.
\newblock \emph{{Handbook of Markov Chain Monte Carlo}}.
\newblock CRC Press, 2011.

\bibitem[Carithers and Moore(2015)]{carithers2015genotype}
Latarsha~J Carithers and Helen~M Moore.
\newblock The genotype-tissue expression (gtex) project.
\newblock \emph{Biopreservation and biobanking}, 13\penalty0 (5):\penalty0 307, 2015.

\bibitem[Chawla et~al.(2002)Chawla, Bowyer, Hall, and Kegelmeyer]{Chawla2002SMOTE}
Nitesh~V. Chawla, Kevin~W. Bowyer, Lawrence~O. Hall, and Wang Kegelmeyer.
\newblock {SMOTE}: Synthetic minority over-sampling technique.
\newblock \emph{Journal of Artificial Intelligence Research}, 16:\penalty0 321--357, 2002.
\newblock URL \url{https://www.jair.org/index.php/jair/article/view/10301}.

\bibitem[Chen et~al.(2018)Chen, Zhang, Wang, Li, and Chen]{chen2018unified}
Changyou Chen, Ruiyi Zhang, Wenlin Wang, Bai Li, and Liqun Chen.
\newblock A unified particle-optimization framework for scalable bayesian sampling.
\newblock In \emph{UAI}, 2018.

\bibitem[Chen et~al.(2014)Chen, Fox, and Guestrin]{chen2014stochastic}
Tianqi Chen, Emily Fox, and Carlos Guestrin.
\newblock Stochastic gradient {H}amiltonian {M}onte {C}arlo.
\newblock In Eric~P. Xing and Tony Jebara, editors, \emph{Proceedings of the 31st International Conference on Machine Learning}, volume~32 of \emph{Proceedings of Machine Learning Research}, pages 1683--1691. PMLR, 22--24 Jun 2014.

\bibitem[Cinquini et~al.(2021)Cinquini, Giannotti, and Guidotti]{CinquiniBoostingSyntheticCogMI2021}
Martina Cinquini, Fosca Giannotti, and Riccardo Guidotti.
\newblock Boosting synthetic data generation with effective nonlinear causal discovery.
\newblock In \emph{2021 IEEE Third International Conference on Cognitive Machine Intelligence (CogMI)}, 2021.

\bibitem[Dawid(2007)]{dawid2007geometry}
A~Philip Dawid.
\newblock The geometry of proper scoring rules.
\newblock \emph{Annals of the Institute of Statistical Mathematics}, 59\penalty0 (1):\penalty0 77--93, 2007.

\bibitem[Dikkala et~al.(2020)Dikkala, Lewis, Mackey, and Syrgkanis]{dikkala2020minimax}
Nishanth Dikkala, Greg Lewis, Lester Mackey, and Vasilis Syrgkanis.
\newblock Minimax estimation of conditional moment models.
\newblock \emph{Advances in Neural Information Processing Systems}, 33:\penalty0 12248--12262, 2020.

\bibitem[Dinh et~al.(2015)Dinh, Krueger, and Bengio]{dinh2014nice}
Laurent Dinh, David Krueger, and Yoshua Bengio.
\newblock {NICE}: Non-linear independent components estimation.
\newblock In \emph{International Conference on Learning Representations}, 2015.

\bibitem[Dinh et~al.(2017)Dinh, Sohl-Dickstein, and Bengio]{dinh2016density}
Laurent Dinh, Jascha Sohl-Dickstein, and Samy Bengio.
\newblock Density estimation using {R}eal {NVP}.
\newblock In \emph{International Conference on Learning Representations}, 2017.

\bibitem[Dong et~al.(2022)Dong, Guo, Li, Ting, Liu, and Kung]{dong2022neural}
Xin Dong, Junfeng Guo, Ang Li, Wei-Te Ting, Cong Liu, and HT~Kung.
\newblock Neural mean discrepancy for efficient out-of-distribution detection.
\newblock In \emph{Proceedings of the IEEE/CVF Conference on Computer Vision and Pattern Recognition}, pages 19217--19227, 2022.

\bibitem[Donoho et~al.(2000)]{donoho2000high}
David~L Donoho et~al.
\newblock High-dimensional data analysis: The curses and blessings of dimensionality.
\newblock \emph{AMS math challenges lecture}, 1\penalty0 (2000):\penalty0 32, 2000.

\bibitem[Dowson and Landau(1982)]{DowsonLandau1982Frechet}
D.~C. Dowson and B.~V. Landau.
\newblock The fr\'echet distance between multivariate normal distributions.
\newblock \emph{Journal of Mathematical Analysis and Applications}, 12\penalty0 (3):\penalty0 450--455, 1982.

\bibitem[Duane et~al.(1987)Duane, Kennedy, Pendleton, and Roweth]{duane87hybrid}
Simon Duane, A.D. Kennedy, Brian~J. Pendleton, and Duncan Roweth.
\newblock Hybrid {M}onte {C}arlo.
\newblock \emph{Physics Letters B}, 195\penalty0 (2):\penalty0 216--222, 1987.

\bibitem[Duncan et~al.(2019)Duncan, N{\"u}sken, and Szpruch]{duncan2019geometry}
Andrew Duncan, Nikolas N{\"u}sken, and Lukasz Szpruch.
\newblock On the geometry of {S}tein variational gradient descent.
\newblock \emph{arXiv preprint arXiv:1912.00894}, 2019.

\bibitem[Dunson and Johndrow(2019)]{dunson2019hastings}
D~B Dunson and J~E Johndrow.
\newblock {The Hastings algorithm at fifty}.
\newblock \emph{Biometrika}, 107\penalty0 (1):\penalty0 1--23, 2019.

\bibitem[Dziugaite et~al.(2015)Dziugaite, Roy, and Ghahramani]{dziugaite2015training}
Gintare~Karolina Dziugaite, Daniel~M Roy, and Zoubin Ghahramani.
\newblock Training generative neural networks via maximum mean discrepancy optimization.
\newblock \emph{arXiv preprint arXiv:1505.03906}, 2015.

\bibitem[Efron and Morris(1977)]{efron1977stein}
Bradley Efron and Carl Morris.
\newblock Stein's paradox in statistics.
\newblock \emph{Scientific American}, 236\penalty0 (5):\penalty0 119--127, 1977.

\bibitem[Efron and Tibshirani(1994)]{efron1994introduction}
Bradley Efron and Robert~J Tibshirani.
\newblock \emph{An introduction to the bootstrap}.
\newblock CRC press, 1994.

\bibitem[Fan and Gu(2023)]{fan2023factor}
Jianqing Fan and Yihong Gu.
\newblock Factor augmented sparse throughput deep relu neural networks for high dimensional regression.
\newblock \emph{Journal of the American Statistical Association}, \penalty0 (just-accepted):\penalty0 1--28, 2023.

\bibitem[Fan et~al.(2021)Fan, Ma, and Zhong]{fan2021selective}
Jianqing Fan, Cong Ma, and Yiqiao Zhong.
\newblock A selective overview of deep learning.
\newblock \emph{Statistical Science}, 36\penalty0 (2):\penalty0 264--290, 2021.

\bibitem[Fan et~al.(2022)Fan, Gu, and Zhou]{fan2022noise}
Jianqing Fan, Yihong Gu, and Wen-Xin Zhou.
\newblock How do noise tails impact on deep relu networks?
\newblock \emph{arXiv preprint arXiv:2203.10418}, 2022.

\bibitem[Farrell et~al.(2021)Farrell, Liang, and Misra]{farrell2021deep}
Max~H Farrell, Tengyuan Liang, and Sanjog Misra.
\newblock Deep neural networks for estimation and inference.
\newblock \emph{Econometrica}, 89\penalty0 (1):\penalty0 181--213, 2021.

\bibitem[Fisher(1922)]{fisher1922mathematical}
Ronald~A Fisher.
\newblock On the mathematical foundations of theoretical statistics.
\newblock \emph{Philosophical transactions of the Royal Society of London. Series A, containing papers of a mathematical or physical character}, 222\penalty0 (594-604):\penalty0 309--368, 1922.

\bibitem[Frigyik et~al.(2008)Frigyik, Srivastava, and Gupta]{Frigyik2008Functional}
B.A. Frigyik, S.~Srivastava, and M.~R. Gupta.
\newblock Functional bregman divergence and bayesian estimation of distributions.
\newblock \emph{IEEE Transactions on Information Theory}, 54\penalty0 (11):\penalty0 5130--5139, 2008.

\bibitem[Gal et~al.(2022)Gal, Alaluf, Yifrach, Patashnik, Simhon, Chechik, and Cohen-Or]{Gal2022TextualInversion}
Rinon Gal, Yuval Alaluf, Yuval Yifrach, Or~Patashnik, Yotam Simhon, Gal Chechik, and Daniel Cohen-Or.
\newblock An image is worth one word: Personalizing text-to-image generation using textual inversion.
\newblock \emph{arXiv preprint arXiv:2208.01618}, 2022.
\newblock URL \url{https://arxiv.org/abs/2208.01618}.

\bibitem[Gao et~al.(2021{\natexlab{a}})Gao, Liu, Zhang, Han, Liu, Niu, and Sugiyama]{gao2021maximum}
Ruize Gao, Feng Liu, Jingfeng Zhang, Bo~Han, Tongliang Liu, Gang Niu, and Masashi Sugiyama.
\newblock Maximum mean discrepancy test is aware of adversarial attacks.
\newblock In \emph{International Conference on Machine Learning}, pages 3564--3575. PMLR, 2021{\natexlab{a}}.

\bibitem[Gao et~al.(2019)Gao, Jiao, Wang, Wang, Yang, and Zhang]{gao2019deep}
Yuan Gao, Yuling Jiao, Yang Wang, Yao Wang, Can Yang, and Shunkang Zhang.
\newblock Deep generative learning via variational gradient flow.
\newblock In Kamalika Chaudhuri and Ruslan Salakhutdinov, editors, \emph{Proceedings of the 36th International Conference on Machine Learning}, volume~97 of \emph{Proceedings of Machine Learning Research}, pages 2093--2101. PMLR, 09--15 Jun 2019.

\bibitem[Gao et~al.(2021{\natexlab{b}})Gao, Huang, Jiao, Liu, Lu, and Yang]{gao2021deep}
Yuan Gao, Jian Huang, Yuling Jiao, Jin Liu, Xiliang Lu, and Zhijian Yang.
\newblock Deep generative learning with {E}uler particle transport.
\newblock In \emph{Proceedings of Machine Learning Research vol 145:1-33, 2021 2nd Annual Conference on Mathematical and Scientific Machine Learning}, 2021{\natexlab{b}}.

\bibitem[Gerber et~al.(2023)Gerber, Han, and Polyanskiy]{gerber23a}
Patrik~R. Gerber, Yanjun Han, and Yury Polyanskiy.
\newblock Minimax optimal testing by classification.
\newblock In Gergely Neu and Lorenzo Rosasco, editors, \emph{Proceedings of Thirty Sixth Conference on Learning Theory}, volume 195 of \emph{Proceedings of Machine Learning Research}, pages 5395--5432. PMLR, 12--15 Jul 2023.
\newblock URL \url{https://proceedings.mlr.press/v195/gerber23a.html}.

\bibitem[Gershman et~al.(2012)Gershman, Hoffman, and Blei]{2012Nonparametric}
S.~Gershman, M.~Hoffman, and D.~Blei.
\newblock Nonparametric variational inference.
\newblock \emph{ICML}, 2012.

\bibitem[Giné and Nickl(2021)]{gine_nickl2021}
Evarist Giné and Richard Nickl.
\newblock \emph{Mathematical Foundations of Infinite-Dimensional Statistical Models}.
\newblock Cambridge Series in Statistical and Probabilistic Mathematics. Cambridge University Press, 2021.
\newblock \doi{10.1017/9781009022811}.

\bibitem[Girolami and Calderhead(2011)]{girolami2011riemann}
Mark Girolami and Ben Calderhead.
\newblock {Riemann manifold Langevin and Hamiltonian Monte Carlo methods}.
\newblock \emph{Journal of the Royal Statistical Society: Series B (Statistical Methodology)}, 73\penalty0 (2):\penalty0 123--214, 2011.

\bibitem[Gneiting and Raftery(2007)]{gneiting2007strictly}
Tilmann Gneiting and Adrian~E Raftery.
\newblock Strictly proper scoring rules, prediction, and estimation.
\newblock \emph{Journal of the American statistical Association}, 102\penalty0 (477):\penalty0 359--378, 2007.

\bibitem[Goodfellow et~al.(2014)Goodfellow, Pouget-Abadie, Mirza, Xu, Warde-Farley, Ozair, Courville, and Bengio]{goodfellow2014generative}
Ian Goodfellow, Jean Pouget-Abadie, Mehdi Mirza, Bing Xu, David Warde-Farley, Sherjil Ozair, Aaron Courville, and Yoshua Bengio.
\newblock Generative adversarial nets.
\newblock In \emph{Advances in Neural Information Processing Systems 27}, pages 2672--2680. Curran Associates, Inc., 2014.

\bibitem[Grenander and Miller(1994)]{grenander1994representations}
Ulf Grenander and Michael~I Miller.
\newblock Representations of knowledge in complex systems.
\newblock \emph{Journal of the Royal Statistical Society: Series B (Methodological)}, 56\penalty0 (4):\penalty0 549--581, 1994.

\bibitem[Gretton et~al.(2006)Gretton, Borgwardt, Rasch, Sch{\"o}lkopf, and Smola]{gretton2006kernel}
Arthur Gretton, Karsten Borgwardt, Malte Rasch, Bernhard Sch{\"o}lkopf, and Alex Smola.
\newblock A kernel method for the two-sample-problem.
\newblock \emph{Advances in neural information processing systems}, 19, 2006.

\bibitem[Gretton et~al.(2012)Gretton, Borgwardt, Rasch, Sch{\"o}lkopf, and Smola]{gretton2012kernel}
Arthur Gretton, Karsten~M Borgwardt, Malte~J Rasch, Bernhard Sch{\"o}lkopf, and Alexander Smola.
\newblock A kernel two-sample test.
\newblock \emph{The Journal of Machine Learning Research}, 13\penalty0 (1):\penalty0 723--773, 2012.

\bibitem[Grosse et~al.(2017)Grosse, Manoharan, Papernot, Backes, and McDaniel]{grosse2017statistical}
Kathrin Grosse, Praveen Manoharan, Nicolas Papernot, Michael Backes, and Patrick McDaniel.
\newblock On the (statistical) detection of adversarial examples.
\newblock \emph{arXiv preprint arXiv:1702.06280}, 2017.

\bibitem[Gulrajani et~al.(2017)Gulrajani, Ahmed, Arjovsky, Dumoulin, and Courville]{gulrajani2017improved}
Ishaan Gulrajani, Faruk Ahmed, Martin Arjovsky, Vincent Dumoulin, and Aaron~C Courville.
\newblock Improved training of wasserstein gans.
\newblock \emph{Advances in neural information processing systems}, 30, 2017.

\bibitem[Gutmann and Hirayama(2011)]{2011Bregman}
M.~Gutmann and J.~I. Hirayama.
\newblock Bregman divergence as general framework to estimate unnormalized statistical models.
\newblock In \emph{Conference on Uai}, 2011.

\bibitem[Gutmann and Hyvärinen(2010)]{gutmann2010Noise}
M.~Gutmann and A~Hyvärinen.
\newblock Noise-contrastive estimation: A new estimation principle for unnormalized statistical models.
\newblock \emph{Journal of Machine Learning Research}, 9:\penalty0 297--304, 2010.

\bibitem[Han et~al.(2021)Han, Zhang, Ding, Gu, Liu, Huo, Qiu, Yao, Zhang, Zhang, Han, Huang, Jin, Lan, Liu, Liu, Lu, Qiu, Song, Tang, Wen, Yuan, Zhao, and Zhu]{HAN2021225}
Xu~Han, Zhengyan Zhang, Ning Ding, Yuxian Gu, Xiao Liu, Yuqi Huo, Jiezhong Qiu, Yuan Yao, Ao~Zhang, Liang Zhang, Wentao Han, Minlie Huang, Qin Jin, Yanyan Lan, Yang Liu, Zhiyuan Liu, Zhiwu Lu, Xipeng Qiu, Ruihua Song, Jie Tang, Ji-Rong Wen, Jinhui Yuan, Wayne~Xin Zhao, and Jun Zhu.
\newblock Pre-trained models: Past, present and future.
\newblock \emph{AI Open}, 2:\penalty0 225--250, 2021.
\newblock ISSN 2666-6510.
\newblock \doi{https://doi.org/10.1016/j.aiopen.2021.08.002}.
\newblock URL \url{https://www.sciencedirect.com/science/article/pii/S2666651021000231}.

\bibitem[Hastings(1970)]{hastings1970monte}
W.~Keith Hastings.
\newblock {M}onte {C}arlo sampling methods using {M}arkov chains and their applications.
\newblock \emph{Biometrika}, 57\penalty0 (1):\penalty0 97--109, 1970.

\bibitem[He et~al.(2016)He, Zhang, Ren, and Sun]{he2016deep}
Kaiming He, Xiangyu Zhang, Shaoqing Ren, and Jian Sun.
\newblock Deep residual learning for image recognition.
\newblock In \emph{Proceedings of the IEEE conference on computer vision and pattern recognition}, pages 770--778, 2016.

\bibitem[Hemmat et~al.(2023)Hemmat, Gholami, and Azimifar]{Hemmat2023FeedbackGuided}
Reyhane~Askari Hemmat, Behnam Gholami, and Zohreh Azimifar.
\newblock Feedback-guided data synthesis for imbalanced classification.
\newblock \emph{arXiv preprint arXiv:2310.00158}, 2023.
\newblock URL \url{https://arxiv.org/abs/2310.00158}.

\bibitem[Heusel et~al.(2017)Heusel, Ramsauer, Unterthiner, Nessler, Handl, Widmann, Konrád, Potamitis, Herdin, Hua, Fischer, Takeda, Kreil, H{\"o}glinger, Klambauer, Mayr, and Hochreiter]{Heusel2017GANs}
Martin Heusel, Hubert Ramsauer, Thomas Unterthiner, Bernhard Nessler, Georg~Friedrich Handl, Michael Widmann, Jakub Konrád, Ilias Potamitis, Heinrich Herdin, Xia Hua, Asja Fischer, Seiji Takeda, Thomas Kreil, Michael H{\"o}glinger, G{\"u}nter Klambauer, Andreas Mayr, and Sepp Hochreiter.
\newblock Gans trained by a two time-scale update rule converge to a local nash equilibrium.
\newblock In \emph{Advances in Neural Information Processing Systems 30}, 2017.

\bibitem[Hines et~al.(2025)Hines, Freeman, Kumar, and Zhang]{Hines2025ScalingLaws}
Greg Hines, C.~Daniel Freeman, Ananya Kumar, and Yi~Zhang.
\newblock Improving the scaling laws of synthetic data with deliberate practice.
\newblock \emph{arXiv preprint arXiv:2502.15588}, 2025.
\newblock URL \url{https://arxiv.org/abs/2502.15588}.

\bibitem[Hoffman and Gelman(2014)]{hoffman2014nuts}
Matthew~D. Hoffman and Andrew Gelman.
\newblock The {N}o-{U}-{T}urn {S}ampler: Adaptively setting path lengths in {H}amiltonian {M}onte {C}arlo.
\newblock \emph{Journal of Machine Learning Research}, 15\penalty0 (47):\penalty0 1593--1623, 2014.

\bibitem[Hoffman et~al.(2013)Hoffman, Blei, Wang, and Paisley]{hoffman2013stochastic}
Matthew~D Hoffman, David~M Blei, Chong Wang, and John Paisley.
\newblock Stochastic variational inference.
\newblock \emph{Journal of Machine Learning Research}, 2013.

\bibitem[Huang et~al.(2024)Huang, Li, and Huang]{huang2024bayesian}
Ding Huang, Ting Li, and Jian Huang.
\newblock Bayesian power steering: An effective approach for domain adaptation of diffusion models.
\newblock \emph{arXiv preprint arXiv:2406.03683}, 2024.

\bibitem[Huang et~al.(2022{\natexlab{a}})Huang, Jiao, Li, Liu, Wang, and Yang]{huang2022error}
Jian Huang, Yuling Jiao, Zhen Li, Shiao Liu, Yang Wang, and Yunfei Yang.
\newblock An error analysis of generative adversarial networks for learning distributions.
\newblock \emph{The Journal of Machine Learning Research}, 23\penalty0 (1):\penalty0 5047--5089, 2022{\natexlab{a}}.

\bibitem[Huang et~al.(2022{\natexlab{b}})Huang, Lam, and Zhang]{huang2022evaluating}
Ziyi Huang, Henry Lam, and Haofeng Zhang.
\newblock Evaluating aleatoric uncertainty via conditional generative models.
\newblock \emph{arXiv preprint arXiv:2206.04287}, 2022{\natexlab{b}}.

\bibitem[Ioffe(2015)]{ioffe2015batch}
Sergey Ioffe.
\newblock Batch normalization: Accelerating deep network training by reducing internal covariate shift.
\newblock \emph{arXiv preprint arXiv:1502.03167}, 2015.

\bibitem[Jia et~al.(2021)Jia, Nezhadarya, Wu, and Ba]{jia2021efficient}
Sheng Jia, Ehsan Nezhadarya, Yuhuai Wu, and Jimmy Ba.
\newblock Efficient statistical tests: A neural tangent kernel approach.
\newblock In \emph{International Conference on Machine Learning}, pages 4893--4903. PMLR, 2021.

\bibitem[Jiao et~al.(2023)Jiao, Shen, Lin, and Huang]{jiao2023deep}
Yuling Jiao, Guohao Shen, Yuanyuan Lin, and Jian Huang.
\newblock Deep nonparametric regression on approximate manifolds: Nonasymptotic error bounds with polynomial prefactors.
\newblock \emph{The Annals of Statistics}, 51\penalty0 (2):\penalty0 691--716, 2023.

\bibitem[Jordan et~al.(1998)Jordan, Kinderlehrer, and Otto]{jordan1998variational}
Richard Jordan, David Kinderlehrer, and Felix Otto.
\newblock The variational formulation of the {F}okker--{P}lanck equation.
\newblock \emph{SIAM Journal on Mathematical Analysis}, 29\penalty0 (1):\penalty0 1--17, 1998.

\bibitem[Kallenberg and Kallenberg(1997)]{kallenberg1997foundations}
Olav Kallenberg and Olav Kallenberg.
\newblock \emph{Foundations of modern probability}, volume~2.
\newblock Springer, 1997.

\bibitem[Kanamori and Sugiyama(2014)]{kanamori2014statistical}
Takafumi Kanamori and Masashi Sugiyama.
\newblock Statistical analysis of distance estimators with density differences and density ratios.
\newblock \emph{Entropy}, 16\penalty0 (2):\penalty0 921--942, 2014.

\bibitem[Khurana et~al.(2023)Khurana, Joshi, Kumar, Varma, and Chaudhari]{Khurana2023Fillup}
Utkarsh Khurana, Amogh Joshi, Abhinav Kumar, Manik Varma, and Pratik Chaudhari.
\newblock Fill-up: Balancing long-tailed data with generative models.
\newblock \emph{arXiv preprint arXiv:2306.07200}, 2023.
\newblock URL \url{https://arxiv.org/abs/2306.07200}.

\bibitem[Kim et~al.(2019)Kim, Lee, and Lei]{kim2019global}
Ilmun Kim, Ann~B Lee, and Jing Lei.
\newblock Global and local two-sample tests via regression.
\newblock 2019.

\bibitem[Kim et~al.(2021)Kim, Ramdas, Singh, and Wasserman]{kim2021classification}
Ilmun Kim, Aaditya Ramdas, Aarti Singh, and Larry Wasserman.
\newblock Classification accuracy as a proxy for two-sample testing.
\newblock 2021.

\bibitem[Kingma(2013)]{kingma2013auto}
Diederik~P Kingma.
\newblock Auto-encoding variational bayes.
\newblock \emph{arXiv preprint arXiv:1312.6114}, 2013.

\bibitem[Kingma and Welling(2014)]{kingma2014auto}
Diederik~P Kingma and Max Welling.
\newblock Auto-encoding variational bayes.
\newblock In \emph{ICLR}, 2014.

\bibitem[Korba et~al.(2020)Korba, Salim, Arbel, Luise, and Gretton]{korba2020non}
Anna Korba, Adil Salim, Michael Arbel, Giulia Luise, and Arthur Gretton.
\newblock A non-asymptotic analysis for stein variational gradient descent.
\newblock \emph{Advances in Neural Information Processing Systems}, 33, 2020.

\bibitem[Korba et~al.(2021)Korba, Aubin-Frankowski, Majewski, and Ablin]{korba2021kernel}
Anna Korba, Pierre-Cyril Aubin-Frankowski, Szymon Majewski, and Pierre Ablin.
\newblock Kernel stein discrepancy descent.
\newblock In \emph{International Conference on Machine Learning}, pages 5719--5730. PMLR, 2021.

\bibitem[Krizhevsky et~al.(2012)Krizhevsky, Sutskever, and Hinton]{krizhevsky2012imagenet}
Alex Krizhevsky, Ilya Sutskever, and Geoffrey~E Hinton.
\newblock Imagenet classification with deep convolutional neural networks.
\newblock \emph{Advances in neural information processing systems}, 25, 2012.

\bibitem[K{\"u}bler et~al.(2022)K{\"u}bler, Stimper, Buchholz, Muandet, and Sch{\"o}lkopf]{kubler2022automl}
Jonas~M K{\"u}bler, Vincent Stimper, Simon Buchholz, Krikamol Muandet, and Bernhard Sch{\"o}lkopf.
\newblock Automl two-sample test.
\newblock \emph{Advances in Neural Information Processing Systems}, 35:\penalty0 15929--15941, 2022.

\bibitem[Kullback and Leibler(1951)]{kullback1951information}
Solomon Kullback and Richard~A Leibler.
\newblock On information and sufficiency.
\newblock \emph{The annals of mathematical statistics}, 22\penalty0 (1):\penalty0 79--86, 1951.

\bibitem[LeCun et~al.(1998)LeCun, Bottou, Bengio, and Haffner]{lecun1998gradient}
Yann LeCun, L{\'e}on Bottou, Yoshua Bengio, and Patrick Haffner.
\newblock Gradient-based learning applied to document recognition.
\newblock \emph{Proceedings of the IEEE}, 86\penalty0 (11):\penalty0 2278--2324, 1998.

\bibitem[LeVeque(2007)]{leveque2007finite}
Randall~J LeVeque.
\newblock \emph{Finite Difference Methods for Ordinary and Partial Differential Equations: Steady-state and Time-dependent Problems}, volume~98.
\newblock SIAM, 2007.

\bibitem[Li et~al.(2017)Li, Chang, Cheng, Yang, and P{\'o}czos]{li2017mmd}
Chun-Liang Li, Wei-Cheng Chang, Yu~Cheng, Yiming Yang, and Barnab{\'a}s P{\'o}czos.
\newblock Mmd gan: Towards deeper understanding of moment matching network.
\newblock \emph{Advances in neural information processing systems}, 30, 2017.

\bibitem[Li et~al.(2019)Li, Bai, Li, Wang, Chen, and Carin]{li2019adversarial}
Chunyuan Li, Ke~Bai, Jianqiao Li, Guoyin Wang, Changyou Chen, and Lawrence Carin.
\newblock Adversarial learning of a sampler based on an unnormalized distribution.
\newblock In \emph{The 22nd International Conference on Artificial Intelligence and Statistics}, pages 3302--3311, 2019.

\bibitem[Lin et~al.(2018)Lin, Khetan, Fanti, and Oh]{lin2018pacgan}
Zinan Lin, Ashish Khetan, Giulia Fanti, and Sewoong Oh.
\newblock Pacgan: The power of two samples in generative adversarial networks.
\newblock \emph{Advances in neural information processing systems}, 31, 2018.

\bibitem[Liu et~al.(2019{\natexlab{a}})Liu, Zhuo, Cheng, Zhang, and Zhu]{liu2019understanding}
Chang Liu, Jingwei Zhuo, Pengyu Cheng, Ruiyi Zhang, and Jun Zhu.
\newblock Understanding and accelerating particle-based variational inference.
\newblock In Kamalika Chaudhuri and Ruslan Salakhutdinov, editors, \emph{Proceedings of the 36th International Conference on Machine Learning}, volume~97 of \emph{Proceedings of Machine Learning Research}, pages 4082--4092. PMLR, 09--15 Jun 2019{\natexlab{a}}.

\bibitem[Liu et~al.(2019{\natexlab{b}})Liu, Zhuo, and Zhu]{liu2019understandingmcmc}
Chang Liu, Jingwei Zhuo, and Jun Zhu.
\newblock Understanding {MCMC} dynamics as flows on the {W}asserstein space.
\newblock In Kamalika Chaudhuri and Ruslan Salakhutdinov, editors, \emph{Proceedings of the 36th International Conference on Machine Learning}, volume~97 of \emph{Proceedings of Machine Learning Research}, pages 4093--4103. PMLR, 09--15 Jun 2019{\natexlab{b}}.

\bibitem[Liu et~al.(2020)Liu, Xu, Lu, Zhang, Gretton, and Sutherland]{liu2020learning}
Feng Liu, Wenkai Xu, Jie Lu, Guangquan Zhang, Arthur Gretton, and Danica~J Sutherland.
\newblock Learning deep kernels for non-parametric two-sample tests.
\newblock In \emph{International conference on machine learning}, pages 6316--6326. PMLR, 2020.

\bibitem[Liu(2017)]{liu2017stein}
Qiang Liu.
\newblock Stein variational gradient descent as gradient flow.
\newblock In I.~Guyon, U.~V. Luxburg, S.~Bengio, H.~Wallach, R.~Fergus, S.~Vishwanathan, and R.~Garnett, editors, \emph{Advances in Neural Information Processing Systems}, volume~30. Curran Associates, Inc., 2017.

\bibitem[Liu and Wang(2016)]{liu2016stein}
Qiang Liu and Dilin Wang.
\newblock Stein variational gradient descent: A general purpose bayesian inference algorithm.
\newblock In D.~Lee, M.~Sugiyama, U.~Luxburg, I.~Guyon, and R.~Garnett, editors, \emph{Advances in Neural Information Processing Systems}, volume~29. Curran Associates, Inc., 2016.

\bibitem[Liu et~al.(2021)Liu, Zhou, Jiao, and Huang]{liu2021wasserstein}
Shiao Liu, Xingyu Zhou, Yuling Jiao, and Jian Huang.
\newblock Wasserstein generative learning of conditional distribution.
\newblock \emph{arXiv preprint arXiv:2112.10039}, 2021.

\bibitem[Liutkus et~al.(2019)Liutkus, Simsekli, Majewski, Durmus, and St{\"o}ter]{liutkus2019sliced}
Antoine Liutkus, Umut Simsekli, Szymon Majewski, Alain Durmus, and Fabian-Robert St{\"o}ter.
\newblock Sliced-{W}asserstein flows: Nonparametric generative modeling via optimal transport and diffusions.
\newblock In Kamalika Chaudhuri and Ruslan Salakhutdinov, editors, \emph{Proceedings of the 36th International Conference on Machine Learning}, volume~97 of \emph{Proceedings of Machine Learning Research}, pages 4104--4113. PMLR, 09--15 Jun 2019.

\bibitem[Lopez-Paz and Oquab(2016)]{lopez2016revisiting}
David Lopez-Paz and Maxime Oquab.
\newblock Revisiting classifier two-sample tests.
\newblock \emph{arXiv preprint arXiv:1610.06545}, 2016.

\bibitem[Lu et~al.(2019)Lu, Lu, and Nolen]{lu2019scaling}
Jianfeng Lu, Yulong Lu, and James Nolen.
\newblock Scaling limit of the {S}tein variational gradient descent: The mean field regime.
\newblock \emph{SIAM Journal on Mathematical Analysis}, 51\penalty0 (2):\penalty0 648--671, 2019.

\bibitem[Mammen(1993)]{mammen1993bootstrap}
Enno Mammen.
\newblock Bootstrap and wild bootstrap for high dimensional linear models.
\newblock \emph{The annals of statistics}, 21\penalty0 (1):\penalty0 255--285, 1993.

\bibitem[Meehan et~al.(2020)Meehan, Chaudhuri, and Dasgupta]{meehan2020non}
Casey Meehan, Kamalika Chaudhuri, and Sanjoy Dasgupta.
\newblock A non-parametric test to detect data-copying in generative models.
\newblock In \emph{International Conference on Artificial Intelligence and Statistics}, 2020.

\bibitem[Metropolis et~al.(1953)Metropolis, Rosenbluth, Rosenbluth, Teller, and Teller]{metropolis1953equation}
Nicholas Metropolis, Arianna~W Rosenbluth, Marshall~N Rosenbluth, Augusta~H Teller, and Edward Teller.
\newblock Equation of state calculations by fast computing machines.
\newblock \emph{The journal of Chemical Physics}, 21\penalty0 (6):\penalty0 1087--1092, 1953.

\bibitem[Mohamed and Lakshminarayanan(2016)]{mohamed2016learning}
Shakir Mohamed and Balaji Lakshminarayanan.
\newblock Learning in implicit generative models.
\newblock \emph{arXiv preprint arXiv:1610.03483}, 2016.

\bibitem[Mroueh et~al.(2019)Mroueh, Sercu, and Raj]{mroueh2019sobolve}
Youssef Mroueh, Tom Sercu, and Anant Raj.
\newblock Sobolev descent.
\newblock In Kamalika Chaudhuri and Masashi Sugiyama, editors, \emph{Proceedings of the Twenty-Second International Conference on Artificial Intelligence and Statistics}, volume~89 of \emph{Proceedings of Machine Learning Research}, pages 2976--2985. PMLR, 16--18 Apr 2019.

\bibitem[M{\"u}ller(1997)]{muller1997integral}
Alfred M{\"u}ller.
\newblock Integral probability metrics and their generating classes of functions.
\newblock \emph{Advances in applied probability}, 29\penalty0 (2):\penalty0 429--443, 1997.

\bibitem[Naeem et~al.(2020)Naeem, Oh, Uh, Choi, and Yoo]{naeem2020reliable}
Muhammad~Ferjad Naeem, Seong~Joon Oh, Youngjung Uh, Yunjey Choi, and Jaejun Yoo.
\newblock Reliable fidelity and diversity metrics for generative models.
\newblock In \emph{International Conference on Machine Learning}, pages 7176--7185. PMLR, 2020.

\bibitem[Nakada and Imaizumi(2020)]{nakada2020adaptive}
Ryumei Nakada and Masaaki Imaizumi.
\newblock Adaptive approximation and generalization of deep neural network with intrinsic dimensionality.
\newblock \emph{The Journal of Machine Learning Research}, 21\penalty0 (1):\penalty0 7018--7055, 2020.

\bibitem[Neal(2011)]{neal2011mcmc}
Radford~M. Neal.
\newblock \emph{MCMC Using Hamiltonian Dynamics}, chapter~5.
\newblock CRC Press, 2011.

\bibitem[Newey and McFadden(1994)]{newey1994large}
Whitney~K Newey and Daniel McFadden.
\newblock Large sample estimation and hypothesis testing.
\newblock \emph{Handbook of econometrics}, 4:\penalty0 2111--2245, 1994.

\bibitem[Nguyen et~al.(2024)Nguyen, Vu, Tran, and Nguyen]{nguyen2024dataset}
Quang Nguyen, Truong Vu, Anh Tran, and Khoi Nguyen.
\newblock Dataset diffusion: Diffusion-based synthetic data generation for pixel-level semantic segmentation.
\newblock \emph{Advances in Neural Information Processing Systems}, 36, 2024.

\bibitem[Nguyen et~al.(2010)Nguyen, Wainwright, and Jordan]{nguyen2010estimating}
XuanLong Nguyen, Martin~J Wainwright, and Michael~I Jordan.
\newblock Estimating divergence functionals and the likelihood ratio by convex risk minimization.
\newblock \emph{IEEE Transactions on Information Theory}, 56\penalty0 (11):\penalty0 5847--5861, 2010.

\bibitem[Nowozin et~al.(2016)Nowozin, Cseke, and Tomioka]{nowozin2016f-gan}
Sebastian Nowozin, Botond Cseke, and Ryota Tomioka.
\newblock f-gan: Training generative neural samplers using variational divergence minimization.
\newblock In \emph{Advances in Neural Information Processing Systems 29 (NIPS 2016)}, pages 271--279. Curran Associates, Inc., October 2016.

\bibitem[Pan and Yang(2009)]{pan2009survey}
Sinno~Jialin Pan and Qiang Yang.
\newblock A survey on transfer learning.
\newblock \emph{IEEE Transactions on knowledge and data engineering}, 22\penalty0 (10):\penalty0 1345--1359, 2009.

\bibitem[Podell et~al.(2023)Podell, English, Lacey, Blattmann, Dockhorn, M{\"u}ller, Penna, and Rombach]{podell2023sdxl}
Dustin Podell, Zion English, Kyle Lacey, Andreas Blattmann, Tim Dockhorn, Jonas M{\"u}ller, Joe Penna, and Robin Rombach.
\newblock Sdxl: Improving latent diffusion models for high-resolution image synthesis.
\newblock \emph{arXiv preprint arXiv:2307.01952}, 2023.

\bibitem[Preston(2009)]{preston2009note}
Chris Preston.
\newblock A note on standard borel and related spaces.
\newblock \emph{Journal of Contemporary Mathematical Analysis}, 44:\penalty0 63--71, 2009.

\bibitem[Rao(2008)]{rao2008linear}
C~Radhakrishna Rao.
\newblock Linear models and generalizations, 2008.

\bibitem[Ren et~al.(2016)Ren, Zhu, Li, and Luo]{ren2016conditional}
Yong Ren, Jun Zhu, Jialian Li, and Yucen Luo.
\newblock Conditional generative moment-matching networks.
\newblock \emph{Advances in Neural Information Processing Systems}, 29, 2016.

\bibitem[Repasky et~al.(2023)Repasky, Cheng, and Xie]{repasky2023neural}
Matthew Repasky, Xiuyuan Cheng, and Yao Xie.
\newblock Neural stein critics with staged l 2-regularization.
\newblock \emph{IEEE Transactions on Information Theory}, 2023.

\bibitem[Rezende and Mohamed(2015)]{rezende2015variational}
Danilo Rezende and Shakir Mohamed.
\newblock Variational inference with normalizing flows.
\newblock In Francis Bach and David Blei, editors, \emph{Proceedings of the 32nd International Conference on Machine Learning}, volume~37 of \emph{Proceedings of Machine Learning Research}, pages 1530--1538, Lille, France, 07--09 Jul 2015. PMLR.

\bibitem[Roberts and Stramer(2002)]{roberts2002langevin}
Gareth~O Roberts and Osnat Stramer.
\newblock {Langevin diffusions and Metropolis-Hastings algorithms}.
\newblock \emph{Methodology and computing in applied probability}, 4\penalty0 (4):\penalty0 337--357, 2002.

\bibitem[Roberts and Tweedie(1996)]{roberts1996tweedie}
Gareth~O. Roberts and Richard~L. Tweedie.
\newblock {Exponential convergence of Langevin distributions and their discrete approximations}.
\newblock \emph{Bernoulli}, 2\penalty0 (4):\penalty0 341 -- 363, 1996.

\bibitem[Rombach et~al.(2022)Rombach, Blattmann, Lorenz, Esser, and Ommer]{rombach2022high}
Robin Rombach, Andreas Blattmann, Dominik Lorenz, Patrick Esser, and Bj{\"o}rn Ommer.
\newblock High-resolution image synthesis with latent diffusion models.
\newblock In \emph{Proceedings of the IEEE/CVF conference on computer vision and pattern recognition}, pages 10684--10695, 2022.

\bibitem[Roth et~al.(2017)Roth, Lucchi, Nowozin, and Hofmann]{roth2017stabilizing}
Kevin Roth, Aurelien Lucchi, Sebastian Nowozin, and Thomas Hofmann.
\newblock Stabilizing training of generative adversarial networks through regularization.
\newblock In \emph{Advances in neural information processing systems}, pages 2018--2028, 2017.

\bibitem[Rubin(1987)]{rubin1987multiple}
Donald~B. Rubin.
\newblock \emph{Multiple Imputation for Nonresponse in Surveys}.
\newblock John Wiley \& Sons, 1987.

\bibitem[Sajjadi et~al.(2018)Sajjadi, Bachem, Lucic, Bousquet, and Braun]{AssessingGenerativeModelsPRRecall2018}
Mehdi S.~M. Sajjadi, Olivier Bachem, Mario Lucic, Olivier Bousquet, and Sylvain Braun.
\newblock Assessing generative models via precision and recall.
\newblock In \emph{Advances in Neural Information Processing Systems 31 (NeurIPS 2018)}, 2018.
\newblock URL \url{https://proceedings.neurips.cc/paper/2018/hash/80a2b5baa4cd17bee929356f2b96ad42-Abstract.html}.

\bibitem[Salakhutdinov(2015)]{salakhutdinov2015learning}
Ruslan Salakhutdinov.
\newblock Learning deep generative models.
\newblock \emph{Annual Review of Statistics and Its Application}, 2\penalty0 (1):\penalty0 361--385, 2015.

\bibitem[Salim et~al.(2020)Salim, Korba, and Luise]{salim2020wasserstein}
Adil Salim, Anna Korba, and Giulia Luise.
\newblock The wasserstein proximal gradient algorithm.
\newblock \emph{arXiv preprint arXiv:2002.03035}, 2020.

\bibitem[Salim et~al.(2021)Salim, Sun, and Richt{\'a}rik]{salim2021complexity}
Adil Salim, Lukang Sun, and Peter Richt{\'a}rik.
\newblock Complexity analysis of stein variational gradient descent under talagrand's inequality t1.
\newblock \emph{arXiv preprint arXiv:2106.03076}, 2021.

\bibitem[Salimans et~al.(2016)Salimans, Goodfellow, Zaremba, Cheung, Radford, and Chen]{salimans2016improved}
Tim Salimans, Ian Goodfellow, Wojciech Zaremba, Vicki Cheung, Alec Radford, and Xi~Chen.
\newblock Improved techniques for training gans.
\newblock \emph{Advances in neural information processing systems}, 29, 2016.

\bibitem[Santambrogio(2015)]{santambrogio2015optimal}
Filippo Santambrogio.
\newblock \emph{Optimal transport for applied mathematicians}.
\newblock Springer, 2015.

\bibitem[Schrab et~al.(2021)Schrab, Kim, Albert, Laurent, Guedj, and Gretton]{schrab2021mmd}
Antonin Schrab, Ilmun Kim, M{\'e}lisande Albert, B{\'e}atrice Laurent, Benjamin Guedj, and Arthur Gretton.
\newblock Mmd aggregated two-sample test.
\newblock \emph{arXiv preprint arXiv:2110.15073}, 2021.

\bibitem[Schrab et~al.(2022)Schrab, Guedj, and Gretton]{schrab2022ksd}
Antonin Schrab, Benjamin Guedj, and Arthur Gretton.
\newblock Ksd aggregated goodness-of-fit test.
\newblock \emph{Advances in Neural Information Processing Systems}, 35:\penalty0 32624--32638, 2022.

\bibitem[Shen et~al.(2021)Shen, Liu, and Shen]{ShenBoostingDataAnalytics}
Xiaotong Shen, Yifei Liu, and Rex Shen.
\newblock Boosting data analytics with synthetic volume expansion.
\newblock In \emph{Proceedings of the 27th ACM SIGKDD Conference on Knowledge Discovery and Data Mining (KDD '21)}, 2021.

\bibitem[Shen et~al.(2019)Shen, Yang, and Zhang]{shen2019deep}
Zuowei Shen, Haizhao Yang, and Shijun Zhang.
\newblock Deep network approximation characterized by number of neurons.
\newblock \emph{arXiv preprint arXiv:1906.05497}, 2019.

\bibitem[Snderby et~al.(2017)Snderby, Caballero, Theis, Shi, and Huszár]{Snderby2017Amortised}
Casper~Kaae Snderby, Jose Caballero, Lucas Theis, Wenzhe Shi, and Ferenc Huszár.
\newblock Amortised map inference for image super-resolution.
\newblock In \emph{International Conference on Learning Representations}, 2017.

\bibitem[Song et~al.(2023)Song, Wang, Shen, Lin, and Huang]{song2023wasserstein}
Shanshan Song, Tong Wang, Guohao Shen, Yuanyuan Lin, and Jian Huang.
\newblock Wasserstein generative regression.
\newblock \emph{arXiv preprint arXiv:2306.15163}, 2023.

\bibitem[Song and Ermon(2019)]{song2019generative}
Yang Song and Stefano Ermon.
\newblock Generative modeling by estimating gradients of the data distribution.
\newblock In \emph{Advances in Neural Information Processing Systems}, volume~32. Curran Associates, Inc., 2019.

\bibitem[Song et~al.(2021)Song, Sohl-Dickstein, Kingma, Kumar, Ermon, and Poole]{song2021scorebased}
Yang Song, Jascha Sohl-Dickstein, Diederik~P Kingma, Abhishek Kumar, Stefano Ermon, and Ben Poole.
\newblock Score-based generative modeling through stochastic differential equations.
\newblock In \emph{International Conference on Learning Representations}, 2021.

\bibitem[Srivastava et~al.(2014)Srivastava, Hinton, Krizhevsky, Sutskever, and Salakhutdinov]{srivastava2014dropout}
Nitish Srivastava, Geoffrey Hinton, Alex Krizhevsky, Ilya Sutskever, and Ruslan Salakhutdinov.
\newblock Dropout: a simple way to prevent neural networks from overfitting.
\newblock \emph{The journal of machine learning research}, 15\penalty0 (1):\penalty0 1929--1958, 2014.

\bibitem[Sugiyama et~al.(2012{\natexlab{a}})Sugiyama, Suzuki, and Kanamori]{sugiyama2012density}
Masashi Sugiyama, Taiji Suzuki, and Takafumi Kanamori.
\newblock \emph{Density ratio estimation in machine learning}.
\newblock Cambridge University Press, 2012{\natexlab{a}}.

\bibitem[Sugiyama et~al.(2012{\natexlab{b}})Sugiyama, Suzuki, and Kanamori]{sugiyama2012density3}
Masashi Sugiyama, Taiji Suzuki, and Takafumi Kanamori.
\newblock Density-ratio matching under the bregman divergence: a unified framework of density-ratio estimation.
\newblock \emph{Annals of the Institute of Statistical Mathematics}, 64\penalty0 (5):\penalty0 1009--1044, 2012{\natexlab{b}}.

\bibitem[Sutherland et~al.(2016)Sutherland, Tung, Strathmann, De, Ramdas, Smola, and Gretton]{sutherland2016generative}
Danica~J Sutherland, Hsiao-Yu Tung, Heiko Strathmann, Soumyajit De, Aaditya Ramdas, Alex Smola, and Arthur Gretton.
\newblock Generative models and model criticism via optimized maximum mean discrepancy.
\newblock \emph{arXiv preprint arXiv:1611.04488}, 2016.

\bibitem[Tian and Feng(2023)]{tian2023transfer}
Ye~Tian and Yang Feng.
\newblock Transfer learning under high-dimensional generalized linear models.
\newblock \emph{Journal of the American Statistical Association}, 118\penalty0 (544):\penalty0 2684--2697, 2023.

\bibitem[Tierney(1994)]{tierney1994}
Luke Tierney.
\newblock \text{Markov Chains} for exploring posterior distributions.
\newblock \emph{The Annals of Statistics}, 22\penalty0 (4):\penalty0 1701--1728, 1994.

\bibitem[Trabucco et~al.(2024)Trabucco, Doherty, Gurinas, and Salakhutdinov]{Trabucco2024DAFusion}
Brian Trabucco, Kyle Doherty, Maria Gurinas, and Ruslan Salakhutdinov.
\newblock Effective data augmentation with diffusion models.
\newblock In \emph{The Twelfth International Conference on Learning Representations (ICLR 2024)}, 2024.
\newblock URL \url{https://openreview.net/forum?id=V2W9b3mY_l}.

\bibitem[Uehara et~al.(2016)Uehara, Sato, Suzuki, Nakayama, and Matsuo]{uehara2016generative}
Masatoshi Uehara, Issei Sato, Masahiro Suzuki, Kotaro Nakayama, and Yutaka Matsuo.
\newblock Generative adversarial nets from a density ratio estimation perspective, 2016.

\bibitem[van~der Vaart(1998)]{Vaart1998}
A.~W. van~der Vaart.
\newblock \emph{Aymptotic statistics}.
\newblock Cambridge University Press, 1998.

\bibitem[Van~der Vaart(2000)]{van2000asymptotic}
Aad~W Van~der Vaart.
\newblock \emph{Asymptotic statistics}, volume~3.
\newblock Cambridge university press, 2000.

\bibitem[van~der Vaart and Wellner(1996)]{Vaart1996}
Aad~W. van~der Vaart and Jon~A. Wellner.
\newblock \emph{Weak Convergence and Empirical Processes}.
\newblock Springer New York, 1996.
\newblock \doi{10.1007/978-1-4757-2545-2}.

\bibitem[Villani(2008)]{villani2008optimal}
C{\'e}dric Villani.
\newblock \emph{Optimal Transport: Old and New}, volume 338.
\newblock Springer Science \& Business Media, 2008.

\bibitem[Villani and Villani(2009)]{villani2009wasserstein}
C{\'e}dric Villani and C{\'e}dric Villani.
\newblock The wasserstein distances.
\newblock \emph{Optimal transport: old and new}, pages 93--111, 2009.

\bibitem[Villani et~al.(2009)]{villani2009optimal}
C{\'e}dric Villani et~al.
\newblock \emph{Optimal transport: old and new}, volume 338.
\newblock Springer, 2009.

\bibitem[Wainwright and Jordan(2008)]{wainwright2008graphical}
Martin~J. Wainwright and Michael~I. Jordan.
\newblock Graphical models, exponential families, and variational inference.
\newblock \emph{Foundations and Trends in Machine Learning}, 1\penalty0 (1):\penalty0 1--305, 2008.

\bibitem[Wang et~al.(2021)Wang, Gao, and Xie]{wang2021two}
Jie Wang, Rui Gao, and Yao Xie.
\newblock Two-sample test with kernel projected wasserstein distance.
\newblock \emph{arXiv preprint arXiv:2102.06449}, 2021.

\bibitem[Wasserman(2004)]{wasserman2004all}
Larry Wasserman.
\newblock \emph{All of Statistics: A Concise Course in Statistical Inference}.
\newblock Springer, 2004.

\bibitem[Welling and Teh(2011)]{welling2011bayesian}
Max Welling and Yee~Whye Teh.
\newblock Bayesian learning via stochastic gradient {L}angevin dynamics.
\newblock In \emph{Proceedings of the 28th international conference on machine learning}, ICML'11, pages 681--688. ACM, 2011.

\bibitem[Wu(1986)]{wu1986jackknife}
Chien-Fu~Jeff Wu.
\newblock Jackknife, bootstrap and other resampling methods in regression analysis.
\newblock \emph{the Annals of Statistics}, 14\penalty0 (4):\penalty0 1261--1295, 1986.

\bibitem[Xu et~al.(2019)Xu, Skoularidou, Antonoglou, and van~der Schaar]{Xu2019CTGAN}
Lei Xu, Maria Skoularidou, Aris Antonoglou, and Mihaela van~der Schaar.
\newblock {CTGAN}: Effective training of conditional {GAN} for tabular data.
\newblock In \emph{Advances in Neural Information Processing Systems 32 (NeurIPS 2019)}, 2019.
\newblock URL \url{https://proceedings.neurips.cc/paper/2019/hash/2f4fe3f97dd210394cba7198ee88bb07-Abstract.html}.

\bibitem[Xu et~al.(2018)Xu, Huang, Yuan, Guo, Sun, Wu, and Weinberger]{xu2018empirical}
Qiantong Xu, Gao Huang, Yang Yuan, Chuan Guo, Yu~Sun, Felix Wu, and Kilian Weinberger.
\newblock An empirical study on evaluation metrics of generative adversarial networks.
\newblock \emph{arXiv preprint arXiv:1806.07755}, 2018.

\bibitem[Zhang and Yang(2021)]{zhang2021survey}
Yu~Zhang and Qiang Yang.
\newblock A survey on multi-task learning.
\newblock \emph{IEEE transactions on knowledge and data engineering}, 34\penalty0 (12):\penalty0 5586--5609, 2021.

\bibitem[Zhao et~al.(2021)Zhao, Sinha, He, Perreault, Song, and Ermon]{zhao2021comparing}
Shengjia Zhao, Abhishek Sinha, Yutong He, Aidan Perreault, Jiaming Song, and Stefano Ermon.
\newblock Comparing distributions by measuring differences that affect decision making.
\newblock In \emph{International Conference on Learning Representations}, 2021.

\bibitem[Zhou et~al.(2023)Zhou, Jiao, Liu, and Huang]{zhou2023}
Xingyu Zhou, Yuling Jiao, Jin Liu, and Jian Huang.
\newblock A deep generative approach to conditional sampling.
\newblock \emph{Journal of the American Statistical Association}, 118\penalty0 (543):\penalty0 1837--1848, 2023.

\bibitem[Zhu et~al.(2020)Zhu, Liu, and Zhu]{zhu2020variance}
Michael Zhu, Chang Liu, and Jun Zhu.
\newblock Variance reduction and quasi-{N}ewton for particle-based variational inference.
\newblock In Hal~Daumé III and Aarti Singh, editors, \emph{Proceedings of the 37th International Conference on Machine Learning}, volume 119 of \emph{Proceedings of Machine Learning Research}, pages 11576--11587. PMLR, 13--18 Jul 2020.

\bibitem[Zhuo et~al.(2018)Zhuo, Liu, Shi, Zhu, Chen, and Zhang]{zhuo2018message}
Jingwei Zhuo, Chang Liu, Jiaxin Shi, Jun Zhu, Ning Chen, and Bo~Zhang.
\newblock Message passing {S}tein variational gradient descent.
\newblock In Jennifer Dy and Andreas Krause, editors, \emph{Proceedings of the 35th International Conference on Machine Learning}, volume~80 of \emph{Proceedings of Machine Learning Research}, pages 6018--6027. PMLR, 10--15 Jul 2018.

\end{thebibliography}
\end{document}